\documentclass{article}
\usepackage{microtype}
\usepackage{graphicx}
\usepackage{booktabs} 
\usepackage[shortlabels]{enumitem}
\usepackage{titletoc}

\DeclareUnicodeCharacter{2161}{.}

\usepackage{hyperref}

\usepackage[skins,theorems]{tcolorbox}
\tcbset{highlight math style={enhanced,
colframe=black,colback=white,arc=0pt,boxrule=1.0pt}}

\newcommand{\proposed}{{MF-CDMs}}

\usepackage{hyperref}
 \hypersetup{
     colorlinks=true,
     linkcolor=red,
     filecolor=blue,
     citecolor=blue,      
     urlcolor=cyan,
     }
\usepackage{url}            
\usepackage{booktabs}       
\usepackage{physics}

\usepackage{amsfonts}       
\usepackage{nicefrac}       
\usepackage{microtype}      
\usepackage{graphicx}
\usepackage{booktabs} 
\usepackage{amssymb}
\usepackage{amsmath}
\usepackage{amsthm}
\usepackage{mathtools}
\usepackage{tikz}
\usepackage{tikz-cd}
\usepackage{centernot}
\usepackage{bbm}
\usepackage{romannum}
\usepackage{calc}
\usetikzlibrary{shapes, snakes}
\usepackage[font={small}]{caption}
\usepackage{lipsum}  
\usepackage{etoolbox}
\usepackage{wrapfig, blindtext}
\usepackage{multirow}
\usepackage[outline]{contour}
\usepackage[framemethod=tikz]{mdframed}
\usepackage{tikz-cd}

\usepackage[capitalize,noabbrev]{cleveref}

\theoremstyle{plain}
\newtheorem{theorem}{Theorem}[section]
\newtheorem{proposition}[theorem]{Proposition}
\newtheorem{lemma}[theorem]{Lemma}
\newtheorem{corollary}[theorem]{Corollary}
\newtheorem{definition}[theorem]{Definition}

\newtheorem*{remark}{Remark}

\newtheorem{manualtheoreminner}{Theorem}
\newenvironment{manualtheorem}[1]{%
  \renewcommand\themanualtheoreminner{#1}%
  \manualtheoreminner
}{\endmanualtheoreminner}

\newtheorem{manualpropositioninner}{Proposition}
\newenvironment{manualproposition}[1]{%
  \renewcommand\themanualpropositioninner{#1}%
  \manualpropositioninner
}{\endmanualpropositioninner}

\newtheorem{manualcorollaryinner}{Corollary}
\newenvironment{manualcorollary}[1]{%
  \renewcommand\themanualcorollaryinner{#1}%
  \manualcorollaryinner
}{\endmanualcorollaryinner}

\usepackage{pifont}
\usepackage{wrapfig, blindtext}
\usepackage{adjustbox}
\usepackage{subcaption}
\usepackage{multirow}
\usepackage{esvect}
\usepackage{enumitem}
\usepackage{hyperref}
\usepackage{url}
\usepackage{mathtools}
\usepackage{cancel}

\newcommand{\cmark}{\ding{51}}%
\newcommand{\xmark}{\ding{55}}%

\usepackage[font=small,labelfont=bf]{caption}

\usepackage{etoolbox}\AtBeginEnvironment{algorithmic}{\small} 

\newcommand{\ie}{\textit{i}.\textit{e}., }
\newcommand{\eg}{\textit{e}.\textit{g}., }

\usepackage{multibib}
\newcites{AP}{Appendix References}
\nociteAP{*}

\newlist{steps}{enumerate}{1}
\setlist[steps]{
  label=\textbf{Step} \Roman*,
  leftmargin=*,
  align=left
}
\tcbuselibrary{skins}
\tcolorboxenvironment{steps}{
  colframe=red!75!black
}

\usepackage[accepted]{icml2024}

\usepackage{amsmath}
\usepackage{amssymb}
\usepackage{mathtools}
\usepackage{amsthm}

\usepackage[capitalize,noabbrev]{cleveref}

\usepackage[textsize=tiny]{todonotes}

\icmltitlerunning{Mean-field Chaos Diffusion Models}

\begin{document}

\twocolumn[
\icmltitle{Mean-field Chaos Diffusion Models}




\begin{icmlauthorlist}
\icmlauthor{Sungwoo Park}{Berkeley}
\icmlauthor{Dongjun Kim}{Stanford}
\icmlauthor{Ahmed M. Alaa}{Berkeley,UCSF}
\end{icmlauthorlist}

\icmlaffiliation{Berkeley}{Department of Electrical Engineering and Computer Sciences, UC Berkeley}
\icmlaffiliation{Stanford}{Department of Computer Science, Stanford}
\icmlaffiliation{UCSF}{UCSF}

\icmlcorrespondingauthor{Ahmed M. Alaa}{amalaa@berkeley.edu}

\icmlkeywords{Machine Learning, ICML}

\vskip 0.3in
]



\printAffiliationsAndNotice{}  

\pagenumbering{arabic}

\begin{abstract}
In this paper, we introduce a new class of score-based generative models (SGMs) designed to handle high-cardinality data distributions by leveraging concepts from mean-field theory. We present mean-field chaos diffusion models (MF-CDMs), which address the curse of dimensionality inherent in high-cardinality data by utilizing the propagation of chaos property of interacting particles. By treating high-cardinality data as a large stochastic system of interacting particles, we develop a novel score-matching method for infinite-dimensional chaotic particle systems and propose an approximation scheme that employs a subdivision strategy for efficient training. Our theoretical and empirical results demonstrate the scalability and effectiveness of MF-CDMs for managing large high-cardinality data structures, such as 3D point clouds.    
\end{abstract}

\vspace{-8mm}
\section{Introduction}

Generative models serve as a fundamental focus in machine learning, aiming to learn a high-dimensional probability density function. Among the contenders such as Normalizing flows~\cite{rezende2015variational} and energy-based models~\cite{zhao2016energy}, Score-based Generative Models (\textbf{SGMs}), especially have gained widespread recognition of their capabilities on various domains, such as images~\cite{song2021maximum}, time-series~\cite{tashiro2021csdi, time-series_mean}, graphs~\cite{jo2022score} and point-clouds~\cite{zeng2022lion}. The key idea of SGMs is to conceptualize a combination of forward and reverse diffusion processes as generative models. In forward dynamics, the data density is progressively corrupted by following a Markov probability trajectory, eventually transformed into Gaussian density. Consequently, denoising score networks sequentially remove noises in the reverse dynamics, aiming to restore the original state. 

Despite the remarkable empirical successes, recent theoretical studies~\cite{de2022convergence, chen2023improved} have highlighted the limitations on the scalability of SGMs when applied to high-dimensional and high-cardinality data structures. To tackle the challenge, a series of research~\cite{lim2023scorebased,  kerrigan2023diffusion, dutordoir2023neural, hagemann2023multilevel} broadens the scope of diffusion models, introducing new methods for data representation in an infinite-dimensional function space. These macroscopic approaches fully mitigate dimensionality issues in diffusion modeling; however, they make strong assumptions on the function-valued representations of the input data, which limits their applicability to practical settings such as modeling 3D point clouds.

\begin{wrapfigure}{r}{0.163\textwidth}
\small
\vspace{-5mm}
  \begin{center}
    \includegraphics[width=0.16\textwidth]{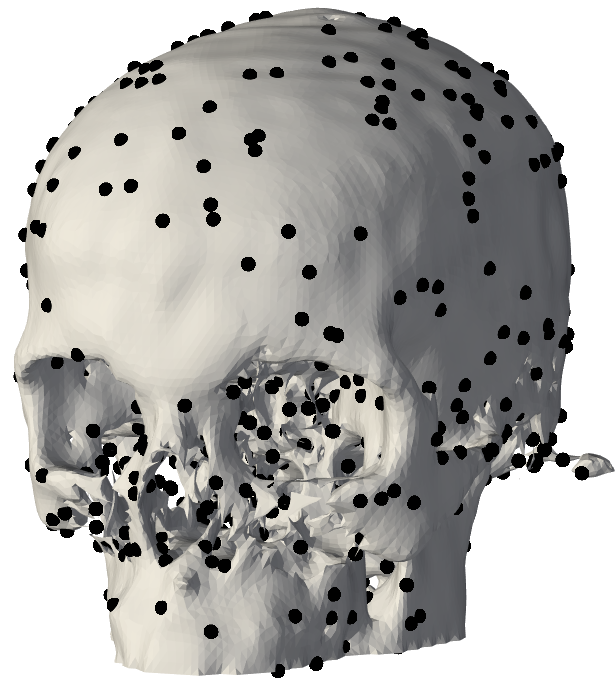}
  \end{center}
  \vspace{-5mm}
  \caption{3D representations of $(\nu^N, \mu)$.}
  \vspace{-4mm}
  \label{fig:example}
\end{wrapfigure}

This paper introduces another strategy to manage high cardinality data through the lens of \textit{mean-field theory} (\textbf{MFT}) and restructure existing SGMs. MFT has long been recognized as a powerful analytical tool for large-scale particle systems in multiple disciplines, such as statistical physics~\cite{kadanoff2009more}, biology~\cite{koehl1994application}, and macroeconomics~\cite{lachapelle2010computation}. Among the diverse concepts developed in MFT, our interest specifically focuses on the property called \textit{propagation of chaos} \textbf{(PoC)}~\cite{sznitman1991topics, gottlieb1998markov}, which describes statistical independency and symmetry in proximity to the mean-field limit of large-particle system. While the direct integration of PoC into conventional SGMs poses a considerable challenge due to the infinite dimensionality, our systematic approach begins by defining denoising models with interacting $N$-particle diffusion dynamics (\ie, $\nu^N$, \textbf{block dots}, Fig.~\ref{fig:example}). We then explore ways to approximate its mean-field limit ($\ie N \rightarrow \infty, \mu$, \textbf{organ surface}), which can possess extensive representational capabilities. This work is centered on two key contributions to achieve this:

\vspace{-3mm}
\begin{itemize}[leftmargin=2.5mm]
    \item \textbf{Mean-field Score Matching.} We introduce a variational framework on Wasserstein space by applying the Itô-Wentzell-Lions formula and derive a \textit{mean-field score matching} (\textbf{MF-SM}) to generalize conventional SGMs for mean-field particle system. We provide mean-field analysis on the asymptotic behavior of the proposed novel framework to elucidate the effectiveness in learning large cardinality data distribution.
    \item \textbf{Subdivision for Efficiency.} For the ease of computational complexity, we introduce a subdivision of chaotic entropy, which establishes piece-wise discontinuous gradient flows and efficiently approximates the true discrepancy in a divide-and-conquer manner.  
\end{itemize}

\vspace{-6mm}
\section{Mean-field Chaos Diffusion Models}\label{sec:methods}

\subsection{Score-based Generative Models}\label{sec:revisit}
Before presenting our proposed method, we provide a brief background on SGMs. For notations not discussed, refer to the detailed descriptions in Appendix. 

Let us consider a probabilistic space $(\mathcal{Y}, \mathcal{F}_t, \mathbb{P})$ and two respective diffusion paths for variables $t$ and $u \coloneqq T - t$.  
\begin{align}
& d\mathbf{X}_u = f_u(\mathbf{X}_u) du + \sigma_u dB_u, \quad \mathbf{X}_u, \mathbf{X}_t \in \mathcal{Y},
\label{eq:sgm_forward_reverse_1} \\ &d\mathbf{X}_t = \left[ f_t(\mathbf{X}_t) - \sigma_t^2 \nabla \log \zeta_t(\mathbf{X}_t) \right]dt + \sigma_t dB_t.
\label{eq:sgm_forward_reverse_2}
\end{align}
A pair of Markovian probability measures $(\zeta_s, \nu_t)$ corresponding to the system of the above SDEs, called forward-reverse SDEs (\ie FR-SDEs), illustrates noising and denoising processes, respectively. A primitive form of the standard objective of SGMs is to minimize the discrepancy (\eg relative entropy, $\tilde{\mathcal{H}}$) between data generative model $\nu_T$ and target data $\zeta_0$ at the terminal state of reverse dynamics, $t = T$: 
\begin{equation}\label{eq:original_relative_entropy}
\mathbf{(P0)} \quad  \min_{\nu_{[0, T]}} \tilde{\mathcal{H}}[\nu_T | \zeta_0],
\end{equation}
where $\nu_{[0, T]}$ denotes a path measure on the interval $[0, T]$. As the direct calculation is intractable, \citet{song2021maximum} have shown that the optimization of an alternative tractable formulation, known as \textit{score matching objective}, can minimize the discrepancy between $\nu_T$ and $\zeta_0$. The goal of SGMs is then to train a score network $\mathbf{s}_{\theta}$ to approximate a \textit{score function} $(\ie \nabla \log \zeta_t)$:
\begin{equation}\label{eq:score_matching}
    \mathcal{J}_{SM}(\theta) \propto \mathbb{E}_{t, \mathbf{X}_t}\left[\norm{\mathbf{s}_{\theta}(t , \mathbf{X}_t) - \nabla \log \zeta_t(\mathbf{X}_t)}^2\right].
\end{equation}
Given the basic machinery defined above, one question naturally arises considering the goal outlined in the introduction: 
\begin{center}
\textbf{Q1.~}\textit{How can we restructure existing diffusion models to preserve \underbar{robust performance} when $\mathbf{dim}(\mathcal{Y}) \rightarrow \infty$?}
\end{center}
Throughout the paper, we address this fundamental question using principles of MFT. As a first step, we begin with dissecting a decomposition of generic FR-SDEs defined on $\mathcal{Y}~(\eg \mathbb{R}^{Nd})$ into the mean-field interacting $N$-particle system on the space $\mathcal{X}~(\eg \mathbb{R}^{d})$.

\subsection{Mean-field Stochastic Differential Equations}\label{sec:mean-field_denoising}
Our new definition of SDEs called \textit{mean-field stochastic differential equations} (\textbf{MF-SDEs}) takes microscopic perspective to model diffusion processes:
\begin{definition}\label{def:denoising_mean_field}(Mean-field SDEs). For the atomless Polish space $\mathcal{X}$, let $\{ B_t^{i, N} \}_{i \leq N}$ be a set of independent Wiener processes on probability space $(\mathcal{X}, \mathcal{F}_t, \mathbb{P})$. Then, we define the $N$-particle system as follows:
\begin{align}
    & d \mathbf{X}_u^{i, N} = 
        f_s(\mathbf{X}_u^{i, N})du + \sigma_u dB_u^{i, N}, \quad \mathbf{X}_u, \mathbf{X}_t \in \mathcal{X}, \label{eq:nerual_mean_field_diffusion_1}\\
    & d \mathbf{X}_t^{i, N} = 
        [f_t(\mathbf{X}_t^{i, N}) -\sigma_t^2 \nabla \log \zeta_t(\mathbf{X}_t^{i, N})]dt
        + \sigma_t dB_t^{i, N}, \label{eq:nerual_mean_field_diffusion_2}
\end{align}
where the initial states of each dynamics is i.i.d. standard Gaussian random vectors, $\ie \mathbf{X}_0^{i, N} \sim \mathcal{N}[\mathbf{I}_d]$.
\end{definition}

The proposed dynamics explicitly delineate the $N$ individual rules of each particle, modeling detailed inter-associations between particles. Upon the structure of MF-SDEs in Definition~\ref{def:denoising_mean_field}, the $N$-particle system is endowed with weak probabilistic structure $\varrho_t^N$ in the $Nd$-dimensional coordinate system $\mathbf{x}^N = (\mathbf{x}_1, \cdots, \mathbf{x}_N) \in \mathcal{X}^N$ and admits a joint density defined as following:
\begin{align}
\mathbf{X}_t^N \sim \nu_t^N \coloneqq \mathbf{Law}(\mathbf{X}_t^{1, N}, \cdots \mathbf{X}_t^{N, N})  = \varrho^{N}_t d\mathbf{x}^N, \label{eq:probabilistic_constraints}\\
\varrho^{M, N}_t(\mathbf{x}^M) = \int_{\mathcal{X}^{N-M}} \varrho^{N}_t(\mathbf{x}^N)d\mathbf{x}_{M+1}\cdots d\mathbf{x}_N.
\end{align}
Furthermore, a set of $N$ particles in the proposed system is exchangeable, satisfying the following \textit{symmetry} property for any given permutation $\tau \in S_N$:
\begin{equation}\label{eq:exchangeable_density}
    \varrho^N_t(\mathbf{x}_1, \cdots, \mathbf{x}_N) = \varrho^N_t(\mathbf{x}_{\tau(1)}, \cdots, \mathbf{x}_{\tau(N)}).
\end{equation}

\textbf{Empirical Measures as Data.} Compared to the data description $\nu_t$ of the macroscopic approach in FR-SDE, our framework interprets a single instance (\eg point cloud) as an empirical random measure $\nu_t^N$, in which particles (\eg point) are represented as marginal random variables $\mathbf{X}_t^{i, N}$,
\begin{equation}
    \underbrace{\mathcal{P}_{2} (\mathcal{Y}) \ni \nu_T}_{\text{FR-SDEs}} ~\leftrightarrow~ \underbrace{\nu_T^N \coloneqq \varrho^N d\mathbf{x}^N \in \mathcal{P}(\mathcal{P}_{2}(\mathcal{X}^N))}_{\text{MF-SDEs}}.
\end{equation}
It is clear from the context that the term `\textbf{cardinality}' stands for the degree of $N$, and the proposed interpretation features two key points. First, our method simply augments the particle count $N\uparrow$ in handling high-resolution data instances, keeping the dimensionality $d = \mathbf{dim}(\mathcal{X})$. This modeling can explicitly expose the effect of increasing cardinality in the analysis as opposite to FR-SDEs, which adjust the dimensionality of the ambient space $N d = \mathbf{dim}(\mathcal{Y})$ without comprehensive details. Second, data representations $\nu_T^N$ naturally inherit the \textit{permutation invariance} which is essential for efficient learning~\cite{niu2020permutation, kim2021setvae} \textit{unstructured data} (\eg sets, point-clouds) as it postulates the exchangeability between the particles (\eg elements, points) as depicted in Eq.~\ref{eq:exchangeable_density}. Throughout, this paper focuses on unstructured data generation to fully leverage this symmetry property.

\vspace{-1mm}
\subsection{Propagation of Chaos and Chaotic Entropy}\label{Sec:PoC}

While we have established a system of individual particles to provide flexible representations, our next step is to adjust the original problem of entropy estimation in $\mathbf{(P0)}$ for $N$-particle system. To do so, we consider the \textit{$N$-particle relative entropy} as a tool for comparing discrepancy between target and generative representations.
\begin{equation}\label{eq:relative_entropy_N}
   \mathcal{H}(\nu_T^{N} | \zeta_0^{\otimes N}) = \frac{1}{N}\int_{\mathcal{X}^N} \bigg[\log \frac{\varrho_T^{N}}{\zeta_{0}^{\otimes N}} \bigg] \varrho_T^{N} d\mathbf{x}^N.
\end{equation}
As the forward diffusion process is defined as a time-varying Ornstein-Ulenbeck process (\eg VP SDE~\cite{score_based_generative_model}), its density for $N$-particles can be represented as a product of Gaussian measures $\zeta_t^{\otimes N}$ defined as: 
\begin{equation}
    d\zeta_{t}^{\otimes N}(\mathbf{x}^N) \coloneqq \prod_{j = 1}^N \mathcal{N}\left(\mathbf{x}_j ; \mathbf{m}_{\zeta}(t), \sigma_{\zeta}^2(t)\mathbf{I}_d \right)d\mathbf{x}_j,
\end{equation}
where the mean vector $\mathbf{m}_{\zeta}(t)$ and covariance matrix $\sigma_{\zeta}^2(t) \mathbf{I}_{d}$ of forward noising Gaussian process $\zeta_t$ are determined by the selection of the model parameters.

\textbf{Propagation of Chaos.} 
Now, we address the question in Sec~\ref{sec:revisit} by bringing attention to the concept in MFT known as \textit{propagation of chaos} suggested by Kac~\cite{kac1956foundations}.

\begin{definition}\label{def:Kac_chaos}(Kac's Chaos). We say that the sequence of marginal measures $\{ \nu_t^{M, N} \}_{M \leq N}$ is $\mu_t$-chaotic, if the following equality holds a.s $[t]$ for all continuous and bounded test functions $\phi$ in the weak sense:
\begin{equation}\label{eq:kac_chaos}
    \langle \nu_t^{M, N}, \phi \rangle \xrightarrow{N \rightarrow \infty} \langle \mu_t^{\otimes M}, \phi\rangle, \quad \forall 1\leq M \ll N.
\end{equation}
\end{definition}

The $\mu_t$-chaotic measures $\{\nu_t^{M, N}\}_{M \leq N}$ begin to behave as if they are statistically indistinguishable with their mean-field limit $\mu_t$ in weak sense for the infinitely large cardinality  $(\ie N\rightarrow \infty)$. With the fact\footnote{Please, refer to Proposition~\ref{prop:chaos} for details.} that our $N$-particle system already enjoys chaoscity, this work exploits the property presented in Eq.~\ref{eq:kac_chaos} to alleviate analytic and computational complexities in generative modeling with infinitely many particles: A finite number $(\eg M)$ of chaotic SDEs can be utilized for training and sampling high-cardinality data instances $(\eg \mu_T)$ only with marginal errors. We will delve into the detailed theoretical rationale in Sec~\ref{sec:sampling}.

\textbf{Chaotic Entropy.} To formalize the problem by leveraging Kac's chaos, we articulate our objective as minimization of \textit{chaotic entropy}~\cite{jabin2017mean, hauray2014kac}, which entails the convergence property $\mathcal{H}(\nu^{N}_T | \zeta^{\otimes N}_0) \xrightarrow{N \rightarrow \infty} \mathcal{H}(\mu_T | \zeta_0)$. Particularly, we propose a new challenging problem: extrapolating the macroscopic modeling from the problem $\mathbf{(P0)}$ to the microscopic counterpart for infinitely many exchangeable particles.
\begin{align}\label{eq:p1}
    \mathbf{(P1)}\quad \min_{\mu_{[0, T]}} \mathcal{H}(\mu_T | \zeta_0) = \min_{\nu_{[0, T]}}  \lim_{N \rightarrow \infty} \mathcal{H}(\nu_T^N | \zeta_0^{\otimes N}).
\end{align}
The equality holds as the property of PoC guarantees weak convergence $\nu_T^{N} \xrightarrow[]{w} \mu_T$. To highlight our approach in addressing the chaotic entropy minimization problem, we have designated our methodology as \textit{mean-field chaos diffusion models} $\textbf{(MF-CDMs)}$. The latter portion of this paper is dedicated to tackling both theoretical and numerical issues associated with solving problem $\mathbf{(P1)}$, by progressively generalizing the main concepts in SGMs. Table~\ref{tab:evolution} outlines how redefined problems in subsequent sections broaden the application of SGMs under the mean-field assumption, featuring the following two key aspects.

(1) \textit{SGMs with Chaotic Entropy}. Due to the intrinsic symmetry in Eq.~\ref{eq:exchangeable_density}, a straightforward derivation of a score-based objective with chaotic relative entropy is non-trivial. Section~\ref{sec:chaotic_entropy} presents the concept of probability measure flows and proposes the \textit{mean-field score matching} objective $(\ie \mathcal{J}_{MF}^{\infty})$ that offers a tractable evaluation of chaotic entropy. 

(2) \textit{Handling Large Cardinality}. Section~\ref{sec:subdivision} introduces a novel numerical approximation scheme termed \textit{subdivision of entropy}, designed to simplify the complex problem presented in $\mathbf{(P1)}$ into new manageable sub-problems in $\mathbf{(P3)}$, efficiently overcoming computational complexity.

\begin{table}[t]
    \centering
    \footnotesize
    \setlength{\tabcolsep}{3pt}
    \renewcommand{\arraystretch}{1.15}
    \begin{tabular}{c|cccc}
    \toprule
    \toprule
         Key & $\nu_T^{\infty}$ & $\mathcal{J}_{MF}^{\infty}$ & $\mathcal{H}_T(\nu_{T}^{\infty})$ & Appx.$\infty$  \\
        concepts & Sec~\ref{sec:methods} & \multicolumn{2}{c}{Sec~\ref{sec:chaotic_entropy}} & Sec~\ref{sec:subdivision} \\
    \midrule
         VP-SDE, $\mathbf{(P0)}$ & \textcolor{red}{\xmark} & \textcolor{red}{\xmark}  & 
         \textcolor{red}{\xmark} & \textcolor{red}{\xmark} \\
         Ours,~$\mathbf{(P1)}$  & \textcolor{blue}{\cmark} & \textcolor{red}{\xmark} & \textcolor{red}{\xmark} & \textcolor{red}{\xmark}\\ 
         Ours,~$\mathbf{(P2)}$  & \textcolor{blue}{\cmark} & \textcolor{blue}{\cmark} & \textcolor{blue}{\cmark} & \textcolor{red}{\xmark}\\     
         Ours,~$\mathbf{(P3)}$  & \textcolor{blue}{\cmark} & \textcolor{blue}{\cmark} & \textcolor{blue}{\cmark} & \textcolor{blue}{\cmark} \\
    \bottomrule
    \bottomrule
    \end{tabular}
    \vspace{-2mm}
    \caption{\textbf{The List of Key Concepts in SGMs for $N \rightarrow \infty$}.}
    \label{tab:evolution}
    \vspace{-3mm}
\end{table}
\vspace{-1mm}
\section{Training MF-CDMs with Chaotic Entropy}\label{sec:chaotic_entropy}

Analysis based on the coordinate system in Eq.~\ref{eq:probabilistic_constraints} rapidly becomes impractical with varying $N$, owing to the curse of dimensionality. To circumvent the issue, we explore an equivalent representation of the $N$-particle system in the space of probability measures: \textit{Wasserstein space} $\mathcal{P}_2(\mathcal{X})$, a domain in which both $\nu_t^N, \mu_t$ inherently lie.

\vspace{-1mm}
\subsection{Denoising Wasserstein Gradient Flows}\label{sec:wgf}

We denote $\mathcal{P}_{2}$ as Wasserstein space consisting of absolutely continuous measures, each of which is characterized by bounded second moments, $\ie \mathcal{P}_{2}(\mathcal{X}) \coloneqq \{ \nu ; d\nu = \varrho d \mathbf{x}, \mathbb{E}d_{\mathcal{X}}^2(\mathbf{x}, \mathbf{x}_0) d\nu(\mathbf{x})  < \infty \}$ and the metric space $(\mathcal{P}_{2}(\mathcal{X}), \mathcal{W}_2)$ can be~\cite{santambrogio2017euclidean} equipped with $2$-Wasserstein distance, $\ie \mathcal{W}_2$. This geometric realization allows functional flows $\mathcal{E} : \mathcal{P}_{2} \rightarrow \mathbb{R}$ along the gradient direction of energy reduction: $\nabla_{\mathcal{P}_2}\mathcal{E}(\varrho) = -\nabla \cdot \left(\varrho \frac{\partial \mathcal{E}}{\partial \delta}(\varrho)\right)(\mathbf{x})$, where the first variation $\partial \mathcal{E}/\partial \delta(\varrho)$~\cite{santambrogio2015optimal} is defined as $\mathbb{E}[\partial \mathcal{E}/\partial \delta(\varrho)\phi(\mathbf{x})] = \lim_{\varepsilon \rightarrow 0}d/d\varepsilon \mathcal{E}(\varrho + \epsilon \phi)$ for all $\phi \in C_0^{\infty}(\mathcal{X})$ satisfying $\mathbb{E}\phi = 0$. To reformulate MF-SDE in a distributional sense, we adopt the concept of \textit{Wasserstein gradient flows} (WGFs) in Eq.~\ref{eq:path_measures} corresponding to denoising $N$-particle MF-SDEs in Eq.~\ref{eq:nerual_mean_field_diffusion_2}.
\begin{align}
    &\frac{\partial}{\partial t}\nu_t^N = -\nabla_{\mathcal{P}_2} \mathcal{E}[\nu_t^N], \quad t \in [0, T] \label{eq:path_measures}\\   
    &\mathcal{E}[\nu_t^N] = \int V^N(t, \mathbf{x}^N, \nu_t^N) + \frac{\sigma_t^2}{2}\log \varrho_t^N d\nu_t^N.
\end{align}  
We specify the functional $V^N$ by extending the concept of variance-preserving SDE~\cite{score_based_generative_model} to the proposed mean-field system. Notably, we consider potential functions $V^N : [0, T] \times \mathcal{X}^N \rightarrow \mathcal{X}^N$ for $N$-particles configurations, termed \textit{mean-field VP-SDE} \textbf{(MF VP-SDE)},  which can be characterized by
\begin{equation}\label{eq:MF-VP-SDE}
    V^N(t, \mathbf{x}^N) = -f_t^{\otimes N}(\mathbf{x}^N) + \sigma^2_t\log \zeta_{T-t}^{\otimes N}(\mathbf{x}^N),
\end{equation}
where we define a drift function as $f_t^{\otimes N} = \beta_t \norm{\mathbf{x}^N}_E^2 / 4$, and the volatility constant is simply set to $\beta_t = \sigma_t^2$ for the pre-defined hyperparameter $\beta_t$.
\begin{figure}[H]
\begin{equation}\label{eq:equivalence}
\small
\tcbset{myinner/.style={no shadow,shrink tight,extrude by=1mm,colframe=black,
  boxrule=0.15pt,frame style={opacity=0.5},interior style={opacity=0.1}}}
\tcbhighmath[drop fuzzy shadow=black!50!yellow,colback=gray!10!white, left=2pt, right=2pt, top=1pt, bottom=1pt]
{
    \underbrace{\frac{\partial}{\partial t} \varrho_t^N = \mathcal{L}_t^N \varrho_t^N}_{\textbf{MF-SDEs}} ~\overset{\text{Prop}~\ref{prop:chaos}}{\longleftrightarrow}~
     \underbrace{\frac{\partial}{\partial t}\nu_t^N = -\nabla_{\mathcal{P}_2}\mathcal{E}[\nu_t^N]}_{\textbf{dWGFs}}.
}
\end{equation}
\vspace{-5mm}
\end{figure}

\textbf{Denoising WFGs.} Eq.~\ref{eq:equivalence} shows that the Liouville equation associated with MF-SDE on the left-hand side can be identified with the proposed WGF on the right-hand side. This implies that our WGF can substitute MF-SDE as a denoising scheme for generative results. From now on, we utilize denoising WGF \textbf{(dWGF)} as our primary tool and derive variation equations in the next section.

\vspace{-2mm}
\subsection{Mean-field Score Matching}\label{sec:mean_field_score_matching}
This section examines a variational equation associated with chaotic entropy. The core idea is to capture infinitesimal changes in Wasserstein metric by applying Itô-Wentzell-Lions formula~\cite{dos2023ito, guo2023ito} to our dWGFs and derive tractable upper bounds.

\begin{theorem}\label{prop:objective_flows}(Wasserstein Variational Equations) Let $\mathcal{M} \coloneqq \mathcal{M}(\zeta_0) < \infty$ be a squared second moment of target data instance $\zeta_0$. We shall refer to the $N$-particle relative entropy as follows:
\begin{equation}
     \mathcal{H}_t^N(\nu_t^N) \coloneqq \mathcal{H}(\nu_t^N | \zeta_{T - t}^{\otimes N}).
\end{equation}
Then, for arbitrary temporal variables $0 \leq s < t \leq T$, and some numerical constants $\mathrm{C}_{0} \lesssim
 \mathcal{O}(\sqrt{d} + \mathcal{M}^2)$, $\mathrm{C}_{1} \lesssim
 \mathcal{O}(T)$, we have variational equations satisfying
\begin{multline}\label{eq:flow_measures_primitive_main}
    \mathcal{H}^N_t(\nu_t^{N}) \lesssim
 \mathcal{H}^N_s(\nu_s^{N}) + \mathrm{C}_{0} \int_s^t \mathcal{O}\left(\mathbb{E}\norm{\nabla_{\mathcal{P}_2}\mathcal{H}^N_r}^{2}_E\right)dr \\ + \mathrm{C}_{1} \int_s^t \mathcal{O}\left( \mathbb{E}\norm{\nabla_x\nabla_{\mathcal{P}_2}\mathcal{H}^N_r}_F^{2}\right) dr.
\end{multline} 
\end{theorem}
As shown in Theorem~\ref{prop:objective_flows}, the geometric deviation in the Wasserstein space affects the norm of the gradient $\nabla_{\mathcal{P}_2}\mathcal{H}^N_r$ in the right-hand side. This indicates that our variation equation exploits geometric information around the law of particles induced by the \textbf{Wasserstein gradient} ($\ie \nabla_{\mathcal{P}_2}\mathcal{H}_t$). This approach is opposed to conventional methodologies~\cite{song2021maximum, CLD} that employ the variational equation concerning \textbf{temporal derivative} ($\ie \partial_t \mathcal{H}_t$). Section~\ref{sec:comparison_VEs} provides an in-depth discussion of the dissimilarity between these two approaches. 

As a comprehensive restatement, we refine the right-hand side in Eq.~\ref{eq:flow_measures_primitive_main} as the Sobolev norm of score functions. 
\begin{corollary}\label{corollary:sobolev_upper_bound} Let $\norm{\cdot}_{W}$ be a norm defined on Sobolev space $W^{1, 2}(\mathcal{X}^N, \nu_t^N)$. Let us define $\mathcal{G}_t = \nabla \log \varrho_{t}^N - \nabla \log \zeta_{T - t}^{\otimes N}$. Then, the $N$-particle entropy can be upper-bounded as follows.
\begin{equation}\label{eq:sobolev_upper}
\mathcal{H}_T^N(\nu_t^N) \precsim \frac{\mathcal{M}}{\sqrt{Nd}}\int_0^T \norm{\mathcal{G}_t}^2_W dt.
\end{equation}
\end{corollary}

Recall that the Sobolev norm of vector-valued function $h \in W^{1,2}$ is defined as $\norm{h}^2_W = \mathbb{E}[\norm{h}_E^2 + \norm{\nabla h}_F^2]$. Corollary~\ref{corollary:sobolev_upper_bound} asserts that the minimization of the $N$-particle relative entropy is achievable when the Sobolev norm on the right-hand side tends to be zero. Motivated by recent studies~\cite{CLD, song2021maximum}, we leverage the inequality in Eq.~\ref{eq:sobolev_upper} to derive our \textit{mean-field score matching} (\textbf{MF-SM}) objective by substituting the score function $\nabla \log \varrho_t^{N}$ with score networks $\mathbf{s}_{\theta}$.  

\begin{definition}\label{definition:MF-SM}(Mean-field Score-Matching) Let us define score networks, denoted as $\mathbf{s}_{\theta} : \Theta \times [0, T] \times \mathcal{X}^N \times \mathcal{P}_2  \rightarrow \mathcal{X}^N$, that satisfies mild regularity conditions. Then, we propose a score-matching objective as
\begin{multline}\label{eq:mean_field_score_matching}
    \mathcal{J}_{MF}^N( \theta, \nu_{[0, T]}^N) \coloneqq \\ \mathbb{E}_{t \sim p(t)}\norm{\mathbf{s}_{\theta}(t, \mathbf{X}_t^N, \nu_t^N) - \nabla \log \zeta_{T - t}^{\otimes N}(\mathbf{X}_t^N)}^2_{W},
\end{multline}
where $p(t)$ is the uniform density on $[0, T]$ and we specify the denoising score networks $\mathbf{s}_{\theta}$ as follows:
\begin{equation}
    \mathbf{s}_{\theta}(t, \mathbf{x}^N, \nu_t^N) = \mathrm{A}_{\theta}(t, \mathbf{x}^N) + \mathrm{B}_{\theta}[\nu_t^{N}](\mathbf{x}^N).
\end{equation}
\end{definition}

\textbf{Design of Mean-field Interaction.} In constructing $\mathbf{s}_{\theta}$, we incorporate  \textit{mean-field interactions} to encapsulate the information of external forces affected by their neighboring particles. To be more specific, we propose a local convolution-based interaction model inspired by \textit{grouping} operations~\cite{qi2017pointnet, qi2017pointnet++, wang2019dynamic} in architectures for 3D point-clouds. 
\begin{equation}
    \mathrm{B}_{\theta}[\nu_t^N](\mathbf{x}^N) \coloneqq [\mathrm{B}_{\theta} *_{\mathbb{B}} \nu_t^{N}](\mathbf{x}^N).
\end{equation}
Here, $*_{\mathbb{B}}$ denotes a truncated convolution operation with respect to the Euclidean ball $\mathbb{B}_R$ of radius $R$. This modeling signifies that interaction with particles outside the convolution domain will be excluded in probability. One may intuitively view this operation as an infinite-dimensional positional encoding, which encapsulates information about geometrically proximate particles. Section~\ref{sec:appendix_ECR} elaborates details on the design of two functions $\mathrm{A}_{\theta}, \mathrm{B}_{\theta}[\nu_t^N]$. 

\textbf{Variation Equation for $\mu_T$.} From the result obtained in Corollary~\ref{corollary:sobolev_upper_bound}, we extend a concept of variation equation for the mean-field limit $\mu_T$ in the subsequent result:

\begin{proposition}\label{prop:upper_bound_H_T} There exist numerical constants $\mathrm{C}_2, \mathrm{C}_3, \mathrm{C}_4 > 0$ such that the $N$-particle relative entropy for an infinity cardinality $N \rightarrow \infty$ can be bounded:
\begin{multline}\label{eq:wasserstein_ve_1}
\underbrace{\mathcal{H}_T^{\infty}(\mu_{T})}_{\mathbf{(P1)}} \precsim 
 \lim_{N \rightarrow \infty}\frac{\mathcal{M}}{\sqrt{Nd}}\mathcal{J}_{MF}^N( \theta, \nu_{[0, T]}^N) \\ + \sigma^{-2}_{\zeta}(T)\underbrace{\mathcal{O}\left(\frac{\mathrm{C}_2}{N}  + \frac{\mathrm{C}_3}{N^{1/2}} + \frac{\mathrm{C}_4}{N^{3/2}}\right)}_{\text{Cardinality Errors } : ~\mathrm{E}(N)} \xrightarrow{N \rightarrow \infty} 0.
\end{multline}
\end{proposition}
Proposition~\ref{prop:upper_bound_H_T} shows that the minimization problem $\mathbf{(P1)}$ on the left-hand side can be upper-bounded with MF-SM and cardinality errors $\mathrm{E}(N)$ in the right-hand side. It is worth noting that our variational framework enhances the conventional score matching, particularly for the representation of data with high cardinality. The coefficient $1/\sqrt{Nd}$ induces robust score estimations and renders the proposed framework \textit{robust} to large cardinality $N$, a property not present in conventional SGMs. As a consequence of the result, the chaotic entropy minimization problem $\mathbf{(P1)}$ can be restructured to involve MF-SM:
\begin{equation}
\begin{aligned}
    & \mathbf{(P2)} \quad \min_{\theta} \lim_{N \rightarrow \infty} \mathcal{J}_{MF}^N( \theta, \nu_{[0, T]}^N).
\end{aligned}    
\end{equation}
The restructured objective reveals that score networks $\mathbf{s}_{\theta}$ is trained to restore vector fields $f_t^{\otimes \infty} - \beta_t\mathbf{s}_{\theta*} \approx \nabla V^{\infty}$ to reconstruct the target instance $\mu_T$ via sampling dWGFs. Unfortunately, optimizing $\mathbf{(P2)}$ may confronts intractability with large cardinality as our score networks $\mathbf{s}_{\theta}$ takes inputs defined on $Nd$-dimensional space $(\eg \mathbf{X}^N \in \mathcal{X}^N)$. 

\vspace{-2mm}
\section{Subdivision of Chaotic Entropy}\label{sec:subdivision} 

\begin{figure*}[t]
\centering
\footnotesize
\includegraphics[width=0.90\textwidth]{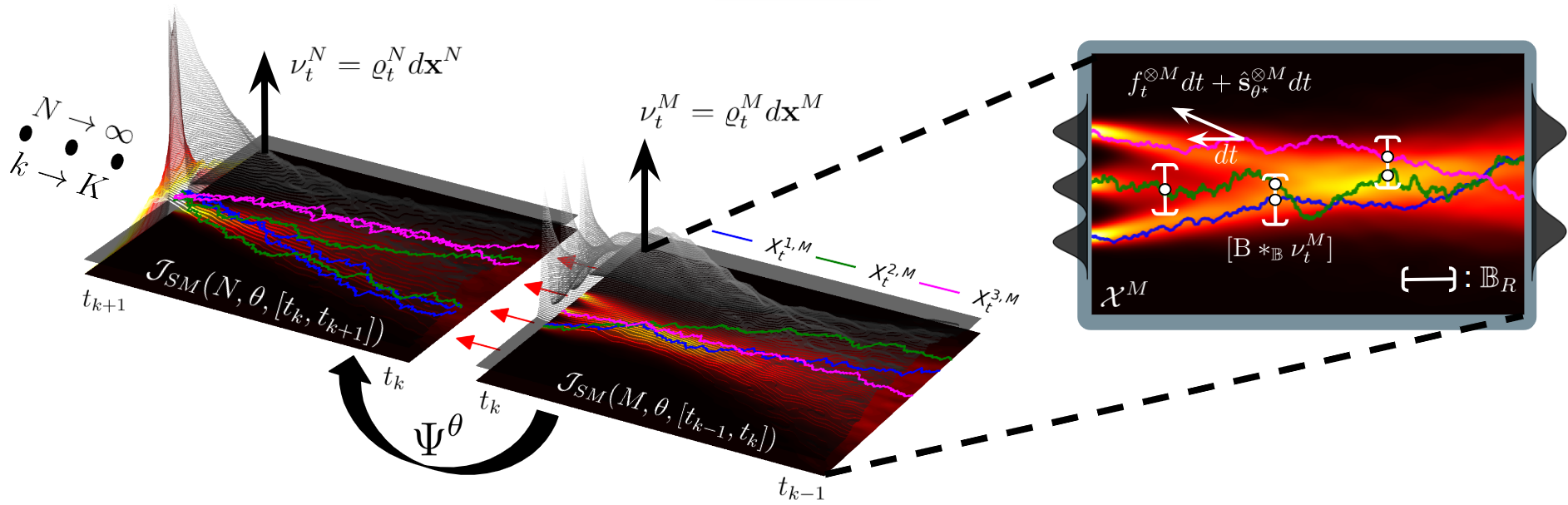}
\vspace{-3mm}
\caption{\textbf{Illustrative Overview of Denoising MF-SDEs/WGFs.} MF-SDEs governing $M$ particles are evolved with respect to vector fields $f_t^{\otimes M} + \mathbf{s}_{\theta}^{\otimes M}$ over the interval $[t_{k-1}, t_k]$, interacting with proximate particles lying in $\mathbb{B}_R$. The illustration depicts the scenario in which the particle branching function $\Psi^{\theta}$ transforms the density of $M=3$ particles into an expanded density for $N=6$ particles $(\eg \text{branching ratio }~\mathfrak{b} = 2)$ following the time interval $t_k$ and result in the joint density $\varrho_t^N$.}
\label{fig:subdivision}
\vspace{-4mm}
\end{figure*}

Our next step is to design an approximation framework that transforms the score-matching objective into computationally tractable variants. Let $\mathbb{N} = \{ N_k ; N_K = N\}$ be a set of non-decreasing cardinality, and $\mathbb{T} = \{ t_k ; t_K = T \}$ be a partition of the interval $[0, T]$, where $k \in \{0, \ldots, K\}$. Then, we subdivide Eq.~\ref{eq:wasserstein_ve_1} into $K$ sub-sequences to obtain alternative and computable upper-bounds:

\begin{proposition}\label{prop:subdivision}(Subdivision) Under the assumption of reducibility\footnote{See Section~\ref{sec:appendix_ECR} for detailed definition and the discussion.} and $\mathfrak{b} > 0, N_{k+1} = \mathfrak{b}N_k$, the chaotic entropy can be split into $K$ sub-problems. 
\begin{multline}\label{eq:subdivision}
\mathcal{H}_T^{\infty}(\mu_{T}) \precsim \lim_{K \rightarrow \infty} \sum_{k=0}^K \bigg[ \sigma_{\zeta}^{-2}(T)\underbrace{\mathrm{E}(N_{k+1})}_{\text{Eq}.~\ref{eq:wasserstein_ve_1}} \\ + \frac{\mathcal{M}}{\sqrt{d}}\underbrace{\bigg( \frac{1}{{{\mathfrak{b}}}\sqrt{N_{k+1}}}\bigg)^{k}\mathcal{J}_{MF}(N_k, \theta, \nu_{[t_k, t_{k+1}]}^{N_k})}_{\text{Subdivision Errors } \geq~\mathcal{J}_{MF}^N( \theta, \nu_{[0, T]}^{N})} \bigg].
\end{multline}
\end{proposition}
We observe that chaotic entropy can be approximated by aggregating $K$ sub-problems of MF-SM, each corresponding to a unique cardinality $N_k$ and a specific interval $[t_k, t_{k+1}]$. This implies that a divide-and-conquer strategy can be effectively employed to address problem $\mathbf{(P2)}$, by treating the sub-problems $\mathcal{J}_{MF}(N_k, \cdot, \cdot)$ individually.

In the decomposed upper-bound in Eq.~\ref{eq:subdivision}, the \textit{particle branching ratio} $\mathfrak{b}$ moderates the impact of sub-problems for large cardinality in the score estimation, leading to improved robustness against $N$. Our final objective function in $\mathbf{(P3)}$ reflects the subdivision of chaotic entropy and the summation is only taken for finite $K$ sub-problems, leveraging the canceling effect gained from the branching ratio. 
\begin{equation}\label{eq:P3}
\begin{aligned}
    & \mathbf{(P3)} \quad \min_{\theta}\sum_{N_k \in \mathbb{N}}^{K = |\mathbb{N}|} \frac{1}{\mathfrak{b}^k}\mathcal{J}_{MF}(N_k, \theta, \nu_{[t_k, t_{k+1}]}^{N_k}).
\end{aligned}
\end{equation}
Section~\ref{sec:appendix_simulation} contains a detailed algorithmic procedures for training score networks $\mathbf{s}_{\theta}$ with the objective $\mathbf{(P3)}$.

\textbf{Particle Branching Function $\Psi^{\theta}$.} The discontinuity of $K$ piece-wise dWGFs $\{ \nu_t^{N_k}, t \in [t_k, t_{k+1}] \}$ associated with individual sub-problems makes the sampling schemes intractable, necessitating the development of gluing pieces together to prevent abrupt changes in distribution. As a remedy, we introduce the \textit{particle branching function} $\Psi_{N_{k+1}}^{\theta}$ to connect the end of previous segment of flows $(\eg \nu_{t_k}^{N_k})$ with the start of next flows $(\eg \nu_{t_k}^{N_{k+1}})$. In a distributional sense, this operation can be represented as a product with a push-forward measure:
\begin{equation}
    (\underbrace{ \mathbf{Id}^{\otimes \mathfrak{b} - 1}}_{(\mathfrak{b} - 1)N_k} \otimes \underbrace{\Psi^{\theta}}_{N_k})_{\#} \nu_{t_k}^{N_{k}} ~\xlongrightarrow[]{}~ \underbrace{\hat{\nu}_{t_k}^{\otimes \mathfrak{b}N_k} = \nu_{t_k}^{\mathfrak{b}N_k}}_{N_{k+1} = \mathfrak{b}N_k}.
\end{equation}
where $(\cdot)_{\#}$ stands for the push-forward operator, and $\mathbf{Id}$ is a identity operator. As a consequence of particle branching, the intermediate flows of probability measure presented as a solution to dWGFs for $N_k$ particles $(\ie \nu_{t_k}^{N_k})$ is augmented with another $(\mathfrak{b} - 1)N_k$ particles, yielding new flows with enhanced cardinality $N_{k+1} = \mathfrak{b}N_k$. Proposition~\ref{prop:optimal_particle_branching} reveals the explicit form of optimal particle branching.

\textbf{Sampling Denoising Dynamics.} After finishing training denoising MF-SDEs/WGFs with the triplet $(\mathbb{N}, N, \mathfrak{b})$, we sample the chaotic dynamics by progressively increasing the cardinality in the middle of the denoising process. The procedure begins by taking initial Gaussian noises distributed as $\zeta_T^{\otimes N_0}$ and propagate particles via Euler scheme with score network $\mathbf{s}_{\theta}$ until reaching the next branching step at $T - t_1$ and each particle branches from $N_0$ to $\mathfrak{b}N_0 = N_1$. By the iteration, we achieve the desired number of chaotic particles. Figure~\ref{fig:subdivision} provides an illustrative overview of the sampling procedure with particle branching along with the denoising WFGs. Section~\ref{sec:appendix_sampling} contains a detailed algorithmic procedure.

\vspace{-1.5mm}
\subsection{Mean-field Analysis of MF-CDMs}\label{sec:sampling}

As this work primarily capitalizes on the mean-field property, this section aims to explore the theoretical implications and benefits of incorporating principles of PoC into the framework of SGMs. The subsequent theoretical findings provide insights to address the question (\ie $\mathbf{Q1}$) posed earlier in Section~\ref{sec:revisit}.

\begin{theorem}\label{theorem:MF-consistency} \textbf{(informal)} Let $\mathfrak{f} \coloneqq \mathfrak{f}(\kappa) > 0$ be a numerical constant dependent on log-Sobolev\footnote{Please refer to Sec~\ref{sec:appendix_convergence_dwgf} for detailed definition.} constant $\kappa$ with respect to proposed dWGFs. Given mild regularity conditions for $\mathbf{s}_{\theta}$, we have short-tailed  concentration probability bound:
\begin{multline}\label{eq:first_implication}
    \mathbb{P}\left[ \mathcal{H}(\nu_t^{M, N} | \mu_t^{\otimes M})  \geq \varepsilon \right] \precsim \qquad (M \ll N \rightarrow \infty) \\ \mathcal{O}(\varepsilon^{-\varepsilon^{-d}}) \cdot 
     \mathcal{O}\left(\exp[-M \mathfrak{f}(\kappa)\varepsilon^2 -M \mathfrak{f}(\kappa)\mathfrak{h}(R)]\right).
    \end{multline}
where For the numerical constant $\mathfrak{h}(R)$ dependent on the radius $R > 0$ for truncation of convolution defined in Section~\ref{sec:appendix_ECR}.
\end{theorem}

\textbf{Concentration of Chaotic Entropy.} The short-tailed concentration of chaotic entropy in Eq.~\ref{eq:first_implication} confirms that a relatively small number of particles $M$ suffices to reconstruct the mean-field surface $\mu_t$ even when the total cardinality diverges to infinite $(N \rightarrow \infty)$. In addition, it demonstrates that infinite cardinality constraints $(\ie \lim_{N \rightarrow \infty})$ specified in $\mathbf{(P2)}$ can be circumvented by subdivision of chaotic entropy in $\mathbf{(P3)}$, as score estimation errors are tolerable in practice with a finite number of sub-problems $|\mathbb{N}| < \infty$ and particle counts $\{N_k\}_{k \leq K}$.

\begin{theorem}\textbf{(informal)} Let us define $F_t \coloneqq \norm{\mathcal{G}_t}_E^2 + \norm{\nabla \mathcal{G}_t}_F^2$, and $\mathbb{E}_{\nu_{[0, T]}^N}\mathbb{E}_{t \sim p(t)} F_t(\mathbf{X}_t^N) = \mathcal{J}_{MF}^N( \theta, \nu_{[0, T]}^N)$, there exist constants $\mathrm{C}_5, \mathrm{C}_6 > 0$, $ \mathbb{N}^{+} \ni q > 4$ such that 
\begin{align}\label{eq:prob_concentration}
    &\mathbb{P}\left( \left| \mathbb{E}_t F(\mathbf{X}_t^{N}) - \mathcal{J}_{MF}(N = 1, \theta, \mu_{[0, T]})\right| \geq \varepsilon \right) \leq \\
    & \exp(-\mathrm{C}_5 \mathfrak{f}(\kappa)^{-2} \left[\varepsilon \sqrt{N} - \mathrm{C}_6 \sqrt{\left(1 + N^{(-q + 4)/2q} \right)}\right]^2) \nonumber.
\end{align}
\end{theorem}

\begin{figure*}[h]
  \centering
  \begin{tabular}[b]{c}
    \includegraphics[width=.47\linewidth]{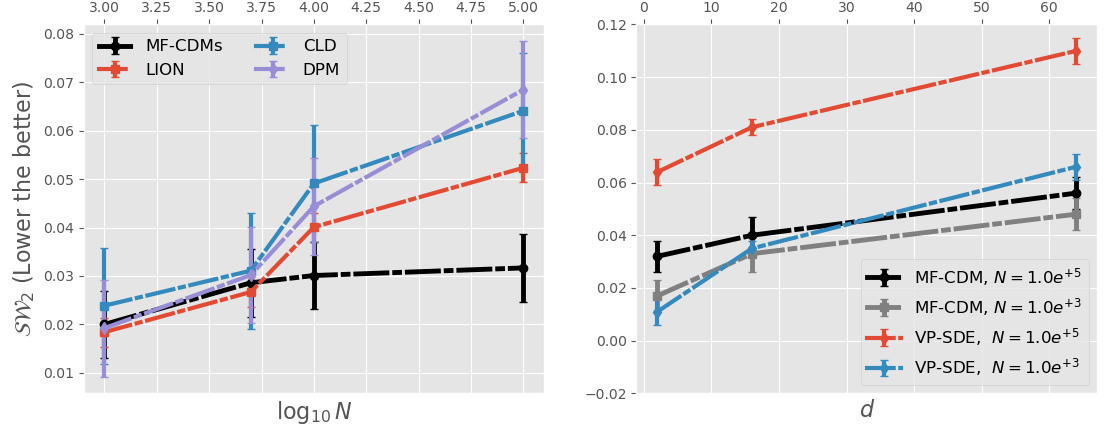}
  \end{tabular}~
  \begin{tabular}[b]{c}
    \includegraphics[width=.47\linewidth]{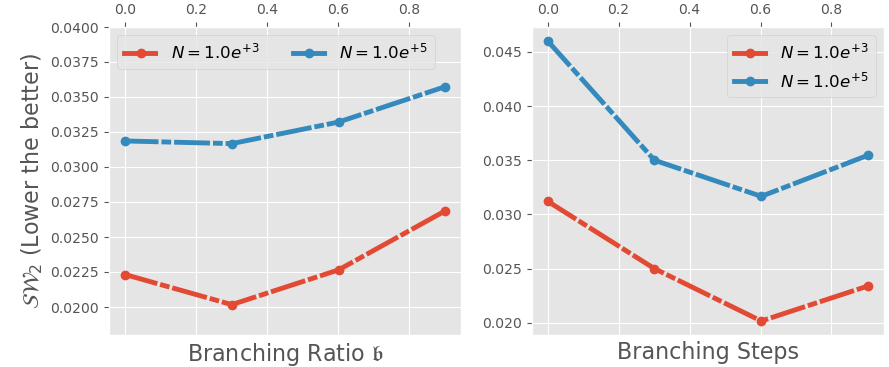}
  \end{tabular}
  \vspace{-8mm}
  \caption{\textit{(Left)} \textbf{Scalability to Data Complexity.} Performance comparisons with varying data dimensionality $d$ and cardinality $N$. \textit{(Right)} \textbf{Ablation Study on Hyperparameters.} Performance variation of MF-CDMs with respect to different hyperparameters; branching ratio $b \in \{1, 2, 4, 8\}$ and number of particle branching $|\mathbb{K}^\prime| \in \{1, 2, 4, 8\}$.}
  \vspace{-4mm}
  \label{fig:ablation_d_N}
\end{figure*}

\textbf{Concentration of MF-SM.} Our second observation in Eq.~\ref{eq:prob_concentration} elucidates that our MF-SM is naturally concentrated on their mean-field limit $\mu_t$ with asymptotically stable probability upper bounds. This shows the remarkable robustness of our objective function when $N \rightarrow \infty$, where conventional score matching objectives $\mathcal{J}_{SM}$ in Eq.~\ref{eq:score_matching} are highly vulnerable to this extreme condition because of the absence of guaranteed stability, as illustrated by Eq.\ref{eq:prob_concentration}.

\section{Related Works}

\textbf{Mean-field Dynamics in Generative Models.} 
Modeling score-based generative models via population dynamics~\cite{koshizuka2023neural, chen2021likelihood, shi2023diffusion} have gained attention recently. Among these, mean-field dynamics through a particle interaction was explored in~\cite{liu2022deep}, where the Schrödinger bridge was integrated to handle mean-field games for the approximation of large population data distributions. \citep{lu2023score} derived score transportation directly from the mean-field Fokker-Planck equation where particle interaction was derived for score-based learning. While these works primarily focus on an analytic perspective and assume an infinite dimensional setting associated with high-dimensional PDEs, our method adopts PoC as a limit algorithm to reduce the potential complexity encountered in dealing with PDEs.

\textbf{Diffusion Models for Unstructured Data.} Recent studies have demonstrated the exceptional performance of diffusion dynamics in point-cloud synthesis~\cite{luo2021diffusion, PVD, zeng2022lion, tyszkiewicz2023gecco}, with a focus on architectural design to impose structural constraints on unstructured data formats. Another stream of research~\cite{hoogeboom2022equivariant, xu2023geometric} considered global geometric constraints to capitalize on \textit{equivariance} property in the modeling of point-clouds. Despite their superior performance, the aforementioned methods face a limitation in the maximum capacity of cardinality owing to rigid structural constraints on localization. In contrast, our method employs a flexible localization using mean-field interaction, requiring only a weak probabilistic structure over the particle set but consistently assures robust performance.

\vspace{-3mm}
\section{Empirical Study}

This section provides a numerical validation of the efficacy of integrating MFT into the SGM framework, particularly in extreme scenarios of large cardinality, where previous works struggle to achieve robust performance.

\textbf{Benchmarks.} We compare our MF-CDMs with well-recognized models in score-based generative models: VP-SDE~\cite{score_based_generative_model}, CLD~\cite{CLD}, and diffusion models for 3D point-cloud: DPM~\cite{luo2021diffusion}, LION~\cite{zeng2022lion}, PVD~\cite{PVD}. For information on the implementation of score networks along with hyperparameters and statistics of datasets with pre-processing, please refer to Sec~\ref{sec:appendix_implementation}.

\vspace{-1.5mm}
\subsection{Synthetic Dataset: Robustness Analysis}\label{sec:experiments_synthetic}

\begin{table}[t]
    \centering
    \small
    \setlength{\tabcolsep}{6pt}
    \renewcommand{\arraystretch}{1.15}
    \begin{tabular}{c|cccc}
    \toprule
    \toprule
         Method& $\small (10^3, 5)$ & $(10^3, 32)$  &  $(10^5, 5)$ &  $(10^5, 32)$  \\
    \midrule
         VP-SDEs &$2.198$ & $2.683$ & $6.943$ & $7.542$\\
         CLD & $2.387$ & $2.826$ & $6.411$ & $7.131$  \\
    \midrule
         DPM & $1.924$ & $2.007$ & $6.847$ & $7.448$  \\ 
         LION  & $\mathbf{1.841}$ & $\mathbf{1.919}$ & $5.234$ & $6.105$  \\  
    \midrule
        \proposed & $2.017$ & $2.413$ & $\mathbf{3.167}$ & $\mathbf{4.059}$ \\
    \bottomrule
    \bottomrule
    \end{tabular}
    \vspace{-2mm}
    \caption{\textbf{Performance Evaluation on the Synthetic data.} We measure performance across different data complexities $(N, d)$ by applying the sliced $2$-Wasserstein distance scaled by a factor of $\times 10^2$. The best results are highlighted in \textbf{bold}.}
    \label{tab:robustness}
    \vspace{-6mm}
\end{table}

The first experiment is designed to evaluate the impact of dimensionality ($\ie d$) and cardinality ($\ie N$) on the robustness of benchmark SGMs when dealing with unstructured data. For this purpose, we generate a synthetic dataset with an equi-weighted Gaussian mixture $\{ \mathbf{Y}_n \}_{n}^N \sim \mathbf{GMM}^{d}(d\mathbf{x}^{d}) \coloneqq (1/8)\sum^{8}_a \mathcal{N}[\mathbf{m}_{a}, \sigma_a \mathbf{I}_{d}]$ where Gaussian parameters $(\mathbf{m}_a, \sigma_{a})$ are randomly selected within unit-cubes $[-1, 1]^d$. The challenge arises as all elements $\{\mathbf{Y}_n \}$ satisfies $p(\mathbf{Y}_m) = p(\mathbf{Y}_n)$ for any $m \neq n \leq N$, and this interchangeability complicates to extract meaningful local associations among the elements, which is essential for efficient learning. To evaluate performance, we employ a tool from optimal transport, \textit{sliced 2-Wasserstein distance} (\ie $\mathcal{SW}_2$)~\cite{kolouri2019generalized}, known for its efficiency in capturing discrepancies between unstructured data instances, especially at high cardinality. 

\begin{figure*}[t]
\centering
\footnotesize
\setlength{\tabcolsep}{6pt}
\renewcommand{\arraystretch}{2.0}
\includegraphics[width=1.0\textwidth]{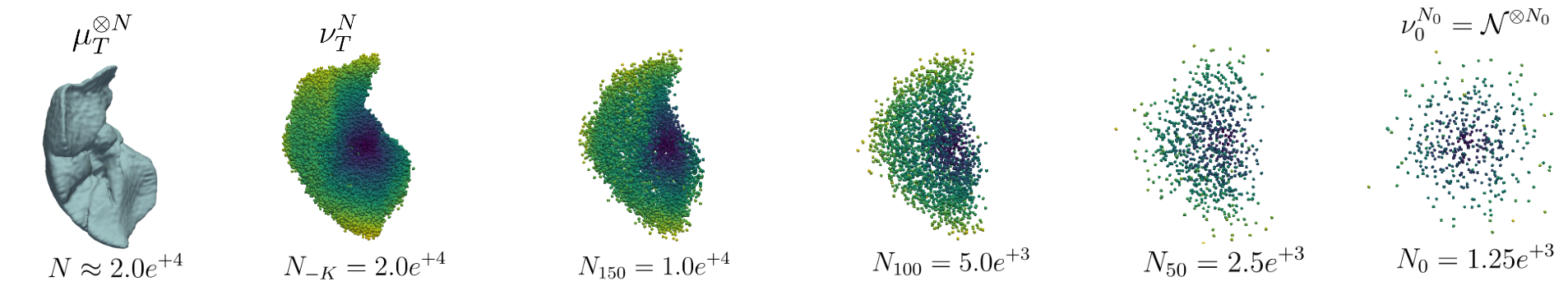}
\vspace{-6mm}
\caption{\textbf{Qualitative Results on MedShapeNet Dataset}. Both $\mu_T^{\otimes N}$ and $\nu_T^{N_{-K}}$ illustrate the target and generated 3D shapes, where displayed \textit{liver} object in MedShapeNet dataset comprises a high-cardinality point-set of nearly $2.0 E^{+4}$ points.}
\label{fig:visualization_liver}
\vspace{-3mm}
\end{figure*}

\textbf{Results.} Fig~\ref{fig:ablation_d_N} and Table~\ref{tab:robustness} present qualitative results when the set cardinality and dimensionality change within the ranges of $N \in \{10^3, 10^5\}$ and $d \in \{5, 32\}$. We note that other methods can easily surpass ours, as the proposed mean-field modeling loses its strength and entails excessive computational complexity with small cardinality (\ie, $N = 10^3$). While existing methods show promising results in low cardinality experiments, their performance significantly deteriorates under conditions of extreme cardinality (\ie, $N = 10^5$). The reason for performance decline is due to their shortcomings in the explainable analysis regarding the curse of dimensionality issue and thus lack of effective modeling of inter-associations among elements. 

In comparison with benchmarks, our method demonstrates robust performance, significantly outperforming all other benchmarks by a large margin in scenarios of $N = 10^5$. Since our methodology has extended VP-SDEs through the integration of PoC in reverse dynamics, the performance gain of MF-CDM over VP-SDEs implies that the chaotic modeling significantly enhances the robustness of conventional SGMs.

\begin{table}[t]
\centering
\footnotesize
\begin{tabular}{c|c|c}
\toprule
\toprule
& \multicolumn{1}{c}{ShapeNet} & \multicolumn{1}{c}{MedShapeNet}\\
~Methods~ & ~~EMD $\downarrow$  ~~/~~ CD $\downarrow$ & EMD $\downarrow$  ~~/~~ CD $\downarrow$~~ \\
\midrule
VP-SDEs  & $4.860 ~~/~~ 4.585$  & $6.387 ~~/~~ 4.616$ \\
CLD  & $4.083 ~~/~~ 5.865$ & $8.647 ~~/~~ 5.632$ \\
\midrule
DPM & $3.058 ~~/~~ 3.269$ & $6.139 ~~/~~ 3.248$ \\
PVD & $3.445 ~~/~~ 3.032$ & $6.386 ~~/~~ 5.902$ \\
LION & $3.248 ~~/~~ 3.248$ & $6.221 ~~/~~ 5.135$ \\
\midrule
\proposed & $\mathbf{2.627 ~~/~~ 1.877}$ & $\mathbf{4.046 ~~/~~ 2.764}$ \\ 
\bottomrule
\bottomrule
\end{tabular}
\label{table:compare_dataset}
\caption{\textbf{Performance Evaluation of 3D point-cloud generation on ShapeNet/MedShapeNet datasets.} The best results are highlighted in \textbf{bold}. Evaluation metrics on EMD and CD are scaled by $10^{2}$ and $10^{2}$, respectively.}
\vspace{-7mm}
\label{table:3d_point_cloud}
\end{table}

\vspace{-1.5mm}
\subsection{Real-world Dataset: 3D Point-cloud Generation}\label{sec:3d_point_cloud}

In the second experiment, we benchmark the empirical performance of MF-CDMs along with existing SGMs for 3D shape diffusion models on two datasets: ShapeNet~\cite{chang2015shapenet} and MedShapeNet~\cite{li2023medshapenet}, with each 3D point-clouds instance consisting of $\mathbf{N = 1.0 E^{+4}}$ and $\mathbf{N = 2.0 E^{+4}}$ points, respectively. The data cardinality in our experiments is up to $10$ times larger than standard setups, which typically focus on scenarios with a relatively limited number of points, ($\eg 2048$). For the fair comparison, we utilized evaluation metrics suggested in~\cite{yang2019pointflow} to compare benchmarks (\ie MMD-EMD, MMD-CD). Owing to the numerical instability of these metrics when applied to high-cardinality objects, we randomly sub-sampled $2048$ points from both the generated $\nu_T^N$ and the target $\mu_T^{\otimes N}$ objects and performed numerical comparisons.

\textbf{Results.} Table~\ref{table:3d_point_cloud} summarizes performance comparisons with benchmarks. Without requiring any strong localization modules, our MF-CDM surpasses all other benchmarks on two datasets, showing its efficiency in real-world settings. It is worth highlighting that task-oriented methods, such as PVD and LION, have achieved state-of-the-art performance on the ShapeNet dataset with $2048$ points. However, they suffer from a drastic performance decline when applied to the MedShapeNet dataset as they depend on fixed localization modules, which are primarily optimized for low cardinality data. We also posit that our superiority stems from the concentration property of large particle systems, as supported by our theoretical findings in Section~\ref{sec:sampling}.

Figure~\ref{fig:visualization_liver} provides a visualization of the intermediate 3D shape during the denoising process with dWGFs. The simulation of dWGFs  starts with $N_0 = 1.25e^{+3}$ particles and the number of particles are doubled $(\eg b = 2)$ at each of the branching steps $k \in \{50, 100, 150, 200\}$, reaching $N \coloneqq N_{-K} = 2.0 e^{+4}$ at the end of the process. The final illustrative result, $\nu_T^{N_{-K}}$, closely resembles the target 3D anatomic structure $\mu_T^{\otimes N}$ $(\ie \textit{liver})$.  

\vspace{-2mm}
\section{Conclusion}

In this study, we propose \textbf{MF-CDMs}, a novel class of SGMs designed for the efficient generation of unstructured data instances with infinite dimensionality. Beginning with the original entropy minimization problem $\mathbf{(P0)}$, we gradually enlarge our discussion to leverage principles of MFT and pose advanced problems $\mathbf{(P1)} \sim \mathbf{(P3)}$ to deal with the curse of dimensionality issues. Our theoretical results reveal that the MF-CDMs naturally inherit chaoticity, ensuring the robust behavior of our model with infinite cardinality. Experimental results on both synthetic and 3D shape datasets empirically validate the superior capability of our framework in generating data instances. In future works, we hope to apply our methodology across diverse tasks in scientific domains, such as 
physical simulation of large-particle dynamical system~\cite{karniadakis2021physics}, large-molecule polymer generation~\cite{chemical_}.

\section*{Impact Statement}

This paper presents work whose goal is to advance the field of Machine Learning. There are many potential societal consequences 
of our work, none which we feel must be specifically highlighted here.

\nocite{langley00}

\bibliography{icml_conference}
\bibliographystyle{icml2024}

\newpage
\appendix
\onecolumn

\section{Appendix}

\subsection{Notations}\label{sec:appendix_backgrounds}

Throughout the paper, we adhere to the following notations:

\colorbox{rgb:red!10,0.5;green!10,0.5;blue!10,0.5}
{
\parbox{0.98\linewidth}
{
\begin{itemize}
    \item Without loss of generality, we employ the same notation for the tensor product across different objects, including functions, and probability measures denoted as $f \otimes g$, $\mu \otimes \nu$. 
    
    \item For any member of continuous bounded and integrable function class $f$, we denote the self $N$-products and its integral as
    \begin{equation}\label{eq:appendix_def_product}
        f^{\otimes N}(\mathbf{x}^N) = [ f(\mathbf{x}_1), \cdots f(\mathbf{x}_N)], \quad \int f^{\otimes N}(\mathbf{x}^N) \mu^{\otimes N}(d\mathbf{x}^N) = \prod_{i \leq N}\int f(\mathbf{x}_i)\mu(\mathbf{x}_i),
    \end{equation}

    \item We denote coordinate system of for $N$-particle system as $\mathbf{x}^N = (\mathbf{x}_1, \cdots \mathbf{x}_N) \in \mathcal{X}^N$ where each component is represented as $\mathbf{x}_i \in \mathcal{X}$, $i \leq N$.
    \item For the probability measure $\nu$ and the integrable test function $f$, we simply denote $\langle \nu, f \rangle \coloneqq \int f d\nu$ as integral.
    \item The law of the $N$-particle joint density, $\nu_t^N$, falls within the 2-Wasserstein space, which specifically contains absolutely continuous measures, represented by $\mathcal{P}_2 \equiv \mathcal{P}_{2, ac}$. We routinely presume the absolute continuity of all probability measures in this context. 
    
    \item The $N$-particle mean-field dynamics is represented as $\mathbf{X}^N \sim \nu^N \in \mathcal{P}_{2, ac}(\mathcal{X}^N)$. Following by the absolutely continuity, we define the density representation with Radon-Nikodym derivative: $d \nu^N = \varrho^N d\mathbf{x}^N$.
    \item The first $M$ component of $N$-particles will be denoted by $\mathbf{X}^{M, N} \sim \nu^{M, N} \in \mathcal{P}_{2, ac}(\mathcal{X}^M)$ with $d \nu^{M, N} = \varrho^{M, N} d\mathbf{x}^M$.
    \item The $N$-product of probability measure $\nu$ will be denoted by $\nu^{\otimes N} \in \mathcal{P}_{2, ac}(\mathcal{X}^N)$ with $d\nu^{\otimes N} = \varrho^{\otimes N}d\mathbf{x}^N$.
    \item The Euclidean and Frobenius norm will be denoted by $\norm{a}_E, \norm{A}_F$, respectively.
    \item $\mathbf{Sym}(d)$, Set of symmetric matrices with size $(d \times d)$; $\mathbf{GL}(d)$, general liner matrix group of size $(d \times d)$.
    \item Within our mathematical context, the symbols are defined as follows: $D$ and $D_{\varrho}$ for abstract and functional derivatives, respectively; $\nabla_x \coloneqq \nabla$ for the Euclidean gradient;  $\nabla_{\mathcal{P}_2}$ for the Wasserstein gradient; and $\partial_t$ for the temporal derivative. For simplicity, the Jacobian matrix of vector-valued objects $h$ will be interchangeably denoted by $\nabla h \coloneqq \mathcal{J} h$.
    
    \item In the paper, $\mathcal{G}_t$ represents the deviation of score functions for an $N$-particle system, $\mathcal{G}_t^{i, N}$ denotes the projection of these functions onto the $i$-th component, and $\mathcal{G}_t^{\infty}$ corresponds to its mean-field limit.

    \item $\mathcal{L}_p(\mathcal{X}), $ denotes the $L_p$ function space on $\mathcal{X}$, 
    
    \item $\mathbf{Lip}(f)$ is a Lipschitz constant of continuous and bounded function $f$.

    \item $\mathbb{N}$ denotes the index set for cardinality, and $\mathbb{N}^{+}$ is defined as a set of positive integers.
    \item For the maximum and minimum of two real-values, we follow the convention for the notation in literature as $\max(a, b) = a \wedge b, \min(a,b) = a \vee b$. 

\end{itemize}
}
}

\newpage

\subsection{Assumptions and Lemmas}\label{sec:appendix_assumption}

We establish the following assumptions to facilitate existing theoretical frameworks of MFT in analyzing the behavior of the proposed MF-SDEs/dWGFs.

\colorbox{rgb:red!10,0.5;green!10,0.5;blue!10,0.5}
{\parbox{0.98\linewidth}{
\begin{enumerate}
    \item  $\mathbf{(H1)}.$ We always assume the large cardinality in data representation, $\ie N \gg d$. 
    \item $\mathbf{(H2)}.$ For all $j \leq N$, $\mathrm{A}$ and the mean-field interaction $\mathrm{B}$ satisfy the Lipschitz continuity with respect to both $\mathcal{X}$ and $\mathcal{P}_{2, ac}$,
    \begin{align}
    & \norm{[\mathrm{B} * \nu](\mathbf{x}^{j}) - [\mathrm{B} * \nu'](\mathbf{y}^{j})}_E^2 \leq \mathrm{C}_{\mathrm{B}} \left( \norm{\mathbf{x}^j - \mathbf{y}^j}_E^2 + \mathcal{W}_2^2(\nu, \nu')\right),
    \\ & \norm{A(s, \mathbf{x}^j) - A(t, \mathbf{y}^j)}_E^2 \leq \mathrm{C}_{\mathrm{A}}\norm{\mathbf{x}^j - \mathbf{y}^j}_E^2 + \mathrm{C}_A (s \wedge t - s\vee t)^2.
    \end{align}
    By definition and assumptions above, the Lipschitz continuity for the score networks is naturally inferred as
    \begin{equation}
        \norm{\mathbf{s}_{\theta}(s, \mathbf{x}^j, \nu) - \mathbf{s}_{\theta}(t, \mathbf{y}^j, \nu')}_E^2 \leq 2( \mathrm{C}_{\mathrm{A}} \wedge \mathrm{C}_{\mathrm{B}})\norm{\mathbf{x}^j - \mathbf{y}^j}_E^2 + \mathrm{C}_A (s \wedge t - s\vee t)^2 + \mathrm{C}_{\mathrm{B}}\mathcal{W}_2^2(\nu, \nu').
    \end{equation}
    We assume that the second moment of proposed score networks is bounded.
    \begin{equation}
        \norm{\mathbf{s}_{\theta}(t, \mathbf{x}^j, \nu)}_E^2 \leq \mathrm{D}(1 + \norm{\mathbf{x}^j}_E^2).
    \end{equation}
    
    \item $\mathbf{(H3)}.$ There exist real-valued functions $\mathbf{A}, \mathbf{B} \in C^{2}(\mathcal{X})$ and $\mathbf{A}^N, \mathbf{B}^N \in C^{2}(\mathcal{X}^N)$ such that
    \begin{equation}
        \nabla \mathbf{A} = \mathrm{A}, ~\nabla \mathbf{A}^N = \mathrm{A}^N, \quad \nabla \mathbf{B} = \mathrm{B}, ~\nabla \mathbf{B}^N = \mathrm{B}^N,
    \end{equation}
    and those functions are uniformly convex. Equivalently, there exist constants $\gamma_{\mathrm{A}}, \gamma_{\mathrm{B}}, \gamma_{\mathrm{B}}' > 0$ such that  Hessian matrices satisfy following:
    \begin{equation}
        \nabla^2 \mathbf{A} \succeq \gamma_{\mathrm{A}} \mathbf{I}_{d}, \quad \gamma_{\mathrm{B}}' \mathbf{I}_d \succeq \nabla^2 \mathrm{B} \succeq \gamma_{\mathrm{B}} \mathbf{I}_d. 
    \end{equation}

    \item $\mathbf{(H4)}.$ Almost surely, we can always find the score networks $\theta \in \Theta$ that can replace the score function of MF-SDEs. 
    \begin{equation}\label{eq:appendix_NN_score}
        \mathbb{P}\left[ \mathbf{s}_{\theta}(t, \mathbf{x}^N, \nu_t^N) = \nabla \log \varrho_t^N(\mathbf{x}^N) \right] = 1, \quad ~\forall N \in \mathbb{N}.
    \end{equation}
    
    \item $\mathbf{(H5)}.$ For any $t \in [0, T]$, there exist a constant $q > 2, q \neq 4$, such that the solution to non-linear Fokker-Planck equation $\mu_t$ has finite $q$-th moment, $\ie \left(\mathbb{E}_{\mu_t}[\norm{\mathbf{x}}^q]\right)^{1/q}  < \infty$.

    \item $\mathbf{(H6)}.$ For some constant $a > 0$, the following numeric estimation are bounded for any $1 \leq M \leq N$:
    \begin{equation}
        \mathbb{E}_{\mathbf{x} \sim \nu_t^{M, N}}\exp(a ||\mathbf{x}||_E^2) < \infty,~\quad \forall \nu_t^{M, N}(d\mathbf{x}^M) = \varrho_t^{M, N}(\mathbf{x}^M) d\mathbf{x}^M.
    \end{equation}
\end{enumerate}
}}

\colorbox{rgb:red!10,0.5;green!10,0.5;blue!10,0.5}
{\parbox{0.98\linewidth}{
\begin{lemma}\label{lemma:gronwall_ineq}(Grönwall's Lemma, Theorem 5.1~\cite{ethier2009markov}). Assume $h : [0, T] \rightarrow \mathbb{R}$ is bounded non-negative measurable function on $[0,T]$ and $g : [0,T] \rightarrow \mathbb{R}$ is a non-negative integrable function. Let following inequality holds for the constant $\mathrm{a} > 0$, 
\begin{equation}
    h(t) \leq B + \int_0^t g(s)h(s) ds, \quad \longrightarrow \quad 
    h(t) \leq B \exp(\int_0^t g(s) ds), \qquad t \in[0, T].
\end{equation}
\end{lemma}
}}

\newpage

\subsection{Exchangeability, Chaocity, Reducibility}\label{sec:appendix_ECR}
In this section, we discuss three core properties (eg, exchangeability, chaocity, reducibility) of the proposed mean-field $N$-particle system, which will be often referenced in subsequent proofs.

\textbf{Exchangeability}. We first show the universal exchangeability property of sample particles:

\colorbox{rgb:red!10,0.5;green!10,0.5;blue!10,0.5}
{\parbox{0.98\linewidth}{

\begin{proposition}\label{prop:exchangebility}(Exchangebility of $N$-particle system.) Let $\mathbf{X}_t^N \sim \nu_t^N$ be a solution to mean-field SDEs defined in Eq.~\ref{eq:nerual_mean_field_diffusion_2}. Assume that $\mathbf{X}_0^N \sim \mathcal{N}^{\otimes N}[\mathbf{I}_d]$. Then, any particles $\{ \mathbf{X}_t^{i, N}\}_{i \leq N}$ at any time $t \in (0, T]$ are exchangeable.
\end{proposition}
}}
\begin{proof} Since the infinitesimal generator for $N$-particle system lies in the set $\mathcal{L}^N_t \in \{\mathcal{L} ; \tau^{-1} \mathcal{L} \tau = \mathcal{L}, ~\tau \in S_N\}$, all the solutions $\varrho_t^N$ (or $\nu_t^N$) to the Liouville equation in Eq.~\ref{eq:equivalence} are trivially symmetric measures at any time $t \in (0, T]$ when the initial state $\varrho_0^N$ (or $\nu_0^N$) is symmetric. The initial constraint $\varrho_0^N = \mathcal{N}^{\otimes N}$ ensures exchangeability of a set of initial states since samples drawn from two projected components $\pi_i^N \mathcal{N}^{\otimes N}$ and $\pi_j^N \mathcal{N}^{\otimes N}$ are i.i.d for any pairs $(i, j) \in \mathbb{N}^{+} \oplus \mathbb{N}^{+}$, meaning that those random variables are exchangeable.
\end{proof}

The rationale behind the equality in Eq.~\ref{eq:exchangeable_density} is based on the result of Proposition~\ref{prop:exchangebility}, since the initial state of the denoising process assumes i.i.d Gaussianity with the fact that its associated generator $\mathcal{L}^N_t$ is concurrently acting on every particles. 

\textbf{Design of Score Networks, $\mathrm{A}_{\theta}$, $[\mathrm{B}_{\theta} *_{\mathbb{B}} \nu_t^N]$.} We first consider the equi-weighted $N$-product of score networks as following.
\begin{equation}\label{eq:appendix_net_A}
    \mathrm{A}_{\theta}(t, \mathbf{x}^N) = \hat{\mathrm{A}}^{\otimes N}_{\theta}(t, \mathbf{x}^N) = \frac{1}{\sqrt{N}}[\mathrm{A}(t, \mathbf{x}_1, \theta), \cdots, \mathrm{A}(t, \mathbf{x}_N, \theta)]^T \in \mathcal{X}^N. 
\end{equation}
Note that $\mathbf{Lip}(\hat{\mathrm{A}}^{\otimes N}) = \sum_j^N \mathbf{Lip}(\hat{\mathrm{A}}_j^{\otimes N}) = \mathbf{Lip}(\mathrm{A}_{\theta})$. Consequently, we define truncated convolution for $N$-particle system as

\begin{equation}\label{eq:appendix_definition_mean_field_interaction}
[\mathrm{B}_{\theta} *_{\mathbb{B}} \nu_t^{N}](\mathbf{x}^N) = \frac{1}{\sqrt{N}}[[\mathrm{B}_{\theta} * \hat{\nu}_t^N](\mathbf{x}_1), \cdots, [\mathrm{B}_{\theta} * \hat{\nu}_t^N](\mathbf{x}_N)]^T,
\end{equation}
where $\hat{\nu}_t^N = (1/N)\sum_{i'}^N \mathbf{X}_t^{i', N}$ is an empirical projection of $\nu_t^N \in \mathcal{P}_2(\mathcal{X}^N)$ onto $\hat{\nu}_t^N \in \mathcal{P}(\mathcal{P}_2(\mathcal{X}))$. Then, each component in Eq.~\ref{eq:appendix_definition_mean_field_interaction} can be represented as
\begin{equation}
    [\mathrm{B}_{\theta} *_{\mathbb{B}} \hat{\nu}_t^N](\mathbf{x}_j) = \int \mathrm{B}_{\theta}(\mathbf{x}_j - \mathbf{y}_j)d\nu_t^{R}[\mathbf{x}_j](\mathbf{y}_j), \quad 1 \leq j \leq N,
\end{equation}
where $\mathrm{B}_{\theta} : \mathbb{R}^d \rightarrow \mathbb{R}^d$ are score networks parameterized by $\theta \in \Theta$. Here, the truncated measure $\nu_t^R[\mathbf{x}_j](\mathbf{y}_j)$ with respect to the centered particle $\mathbf{x}_j \in \mathbb{R}^d$ is defined as
\begin{equation}
     \quad d\nu_t^{R}[\mathbf{x}_j](\mathbf{y}_j) = \frac{\chi_{\mathbb{B}_R^{\mathbf{x}_j}}\hat{\nu}_t^{N}(d\mathbf{y}_j)}{\hat{\nu}_t^{N}[\mathbb{B}_R^{\mathbf{x}_j}]},
\end{equation}
where $\mathbb{B}_R^{\mathbf{x}_j}$ is a Euclidean ball of radius $R$ centered at $\mathbf{x}_j$ and $\chi_{A}$ represents an indicator function defined on any set $A \subseteq \mathbb{R}^d$.

\textbf{Reducibility.} We say that the function $h : \mathcal{X}^N \rightarrow \mathcal{X}^N$ is \textit{reducible} if there exists at least one $\mathcal{X}$-valued function $\hat{h}: [0, T] \times \mathcal{X} \rightarrow \mathcal{X}$ such that $h = \hat{h}^{\otimes N}$ uniformly, where the function product $\hat{h}^{\otimes N}(t, \mathbf{x}^N) \in \mathcal{X}^N$ is defined in Eq.~\ref{eq:appendix_def_product}. With the definition, the notion of reducibility can be formalized as a kernel of the following functional $\mathcal{R}$ on Sobolev space:
\begin{equation}\label{eq:appendix_functional}
    \mathcal{R}(h) \coloneqq \inf_{\hat{h} \in W^{1,2}}\left[\norm{h(t, \mathbf{x}^N) - \hat{h}^{\otimes N}(t, \mathbf{x}^N)}_W \right].
\end{equation}   
Any functions $h$ in the kernel of functional $\ie \mathbf{Ker}(\mathcal{R}) = \{ h ; \mathcal{R}(h) = 0, h \in W^{1,2}(\mathcal{X}^N)\}$ operates in a particle-wise manner, acting on each particle in parallel. 

By the direct calculation, one can show that our score networks $(\ie \mathbf{s}_{\theta} \coloneqq \mathrm{A}_{\theta} + [\mathrm{B}_{\theta} *_{\mathbb{B}} \nu_t^N])$ are reducible and ready to be implemented for our purpose, as Proposition~\ref{prop:chaos} assures the chaoscity. Furthermore, one can easily show that vector fields $\nabla V^{N}$ of mean-field VP-SDE in~\ref{eq:MF-VP-SDE} also satisfy reducibility. The reducibility condition, particularly, results in substantial computational efficiency in the modeling of score networks $\mathrm{A}_{\theta}, \mathrm{B}_{\theta} \in \mathbf{Ker}(\mathcal{R})$. It permits point-wise operation through GPU-based calculations, thus accelerating the sampling process of the $N$-particle system in high cardinality environments. The reducibility property is critical in our approach, ensuring the particles' chaotic behavior and scalability in the practical application of numerical implementation.

\textbf{Chaocity.} We conclusively demonstrate that our $N$-particle system, modeled by MF-SDEs, not only achieves $\mu_T$-chaos but also exhibits stability in its limit behavior.

\colorbox{rgb:red!10,0.5;green!10,0.5;blue!10,0.5}
{\parbox{0.98\linewidth}{
\begin{proposition}\label{prop:chaos}(Equivalence) Assuming mild Lipschitz continuity, the following three statements are equivalent:
\begin{enumerate}
    \item The $N$-particle entropy Eq.~\ref{eq:relative_entropy_N} becomes chaotic if score networks $\mathbf{s}_{\theta}$ are reducible.
    \item A joint probability density $\varrho_T^N$ solving the Liouville equation in Eq.~\ref{eq:equivalence} is $\mu_T$-chaotic. 
    \item The solution to the dWGF for $N$-particle system in Eq.~\ref{eq:equivalence} becomes $\mu_t$-chaotic if score networks $\mathbf{s}_{\theta}$ are reducible.
\end{enumerate}
\end{proposition}
}}
\begin{proof}
The classical result of the propagation of chaos~\cite{jabin2017mean} with the Lipschitz continuity assumption in $\mathbf{(H2)}$ assures that the denoising dynamics with reducible score networks induce chaoscity as exchangeability is already satisfied by the result of Prop~\ref{prop:exchangebility}. Following by the result suggested in Theorem 1.4~\cite{hauray2014kac}, Kac's chaos $(\ie \mu_T = \lim_{N\rightarrow\infty} \nu_T^N)$ identically implies chaotic entropy given by assumptions of Lipschitz continuity.
\end{proof}

\subsection{Wasserstein Variation Equation}

\textbf{Gradient flows on $\mathcal{P}_{2, ac}$, It\^o's flows of Measures.} With mild assumptions on the regularity of energy functionals (\eg functional differentiability), Wasserstein gradient can be identified with Lions' $L$-derivative~\cite{cardaliaguet2010notes} by utilizing Gâteaux (or Fréchet) derivative of semi-martingale lifting. To be more specific, Theorem~\ref{theorem:appendix_equivalance} reveals the fundamental structure that  Eq.~\ref{eq:flow_measures_primitive} can be rewritten in an alternative form based on a functional analytic perspective.

\colorbox{rgb:red!10,0.5;green!10,0.5;blue!10,0.5}
{\parbox{0.98\linewidth}{
\begin{theorem}\label{theorem:appendix_equivalance}\cite{carmona2018probabilistic} Let us assume that functional $\mathcal{E}$ has a first variation $\partial \mathcal{E} / \partial \delta |_{\mu}$ for any $\mu \in \mathcal{K} \subset \mathcal{P}_{2, ac}$, and define spatial gradient of first variation as
\begin{equation}
    \mathcal{P}_{2, ac} \times \mathbb{R}^d \ni (\mu, \mathbf{x}) \mapsto \nabla_x \frac{\partial \mathcal{E}}{\partial \delta}[\mu](\mathbf{x}) \in \mathbb{R}^d.
\end{equation}
Assume that the mapping is jointly continuous in $(\mu, \mathbf{x})$, and well-defined, at most of the linear growth in $\mathbb{R}^d$, uniformly bounded in subset $\mathcal{K} \subset \mathcal{P}_{2, ac}$. Then Lions' $L$-derivative is identical to the spatial gradient of the first variation.
\end{theorem}
}}

For the a test function $\varphi$ and a solution $\varrho_t$ to dWGFs for $N \rightarrow \infty$ (\eg McKean-Vlasov equation), we apply Gateaux derivative to the infinite-dimensional energy functional $\mathcal{E} : [0, T] \times \mathcal{L}_2(\mathcal{X}) \rightarrow \mathbb{R}$,
\begin{equation}\label{eq:appendix_gaeuaux_derivative}
\begin{split}
    \mathcal{E}(t, \varrho_t) &= \int D_{\varrho}\mathcal{E}(t, \varrho_t, \mathbf{x})\frac{\partial}{\partial t}\varrho_t d\mathbf{x} dt
    \\ &= \mathbb{E}\left[ \nabla_x D_{\varrho}\mathcal{E}(t, \varrho_t, \mathbf{x}) \cdot \nabla V(t, \mathbf{x}, \nu_t) + \frac{1}{2}\mathbf{Tr}[\Sigma(t, \mathbf{x}) \Sigma(t, \mathbf{x})^T\nabla_x^2 D_{\varrho}\mathcal{E}(t, \varrho_t, \mathbf{x})]\right]dt,
\end{split}
\end{equation}
A variety notions for the derivatives of Equation~\ref{eq:appendix_gaeuaux_derivative} have been explored in the literature~\cite{guo2023ito, carmona2018probabilistic, dos2023ito, santambrogio2015optimal}. We examine the identity and details among them as follows:
\begin{equation}\label{eq:appendix_equiv_functional_deriv}
    \nabla_x D_{\varrho}\mathcal{E}|_{t, \mathbf{x}, \varrho = \varrho_t} ~~\xleftrightarrow{~~\text{Sec~7.2~}\text{\cite{santambrogio2015optimal}}~~} ~~\nabla_x \frac{\partial \mathcal{E}}{\partial \delta}|_{t, \mathbf{x}, \varrho_t} ~~\xleftrightarrow{~\text{Theorem~\ref{theorem:appendix_equivalance}}~} ~~\nabla_{\mathcal{P}_2}\mathcal{E}|_{t, \mathbf{x}, \varrho_t} .
\end{equation} 
Assuming the appropriate regularity conditions for each energy functional, we find that three distinct notions of derivatives in Eq~\ref{eq:appendix_equiv_functional_deriv} are congruent. This observation leads us to delve into an alternative definition of the functional derivative and examine its role in defining the evolution of measures over time.

\colorbox{rgb:red!10,0.5;green!10,0.5;blue!10,0.5}
{\parbox{0.98\linewidth}{
\begin{definition}\label{def:ito_flow_measure}(It\^o's Flows of Measures) Given semi-martingale $\mathbf{X}_{(\cdot)}$ with finite variation $\mathbb{E}[\textbf{Var}(V)] < \infty$ and finite quadratic variation $\mathbb{E}[d[\mathbf{X}_{(\cdot)}, \mathbf{X}_{(\cdot)}]] < \infty$, the time-varying energy functional $\mathcal{E} : [0, T] \times \mathcal{P}_{2, ac} \rightarrow \mathbb{R}, ~\mathcal{E} \in \mathcal{C}^{1, 1}(\mathcal{P}_2(\mathcal{X}))$ associated with differential calculus on the Wasserstein space $\mathcal{P}_{2, ac}$ evolves according to dynamics defined as:
\begin{equation}\label{eq:flow_measures_primitive}
    d\mathcal{E}(t, \mathbf{Law}(\mathbf{X}_t)) = \mathbb{E}\left[ \nabla_x\frac{\partial \mathcal{E}}{\partial \delta}(t, \mathbf{Law}(\mathbf{X}_t)) \cdot d\mathbf{X}_t\right] + \mathbb{E}\left[\frac{1}{2}\mathbf{Tr}\left(\nabla_x^2\frac{\partial \mathcal{E}}{\partial \delta}(t, \mathbf{Law}(\mathbf{X}_t)) \cdot d[\mathbf{X}_t, \mathbf{X}_t]\right)\right].
\end{equation}
where $\nabla, \nabla^2$ are gradient and Hessian operators, and the expectation is taken with the law of semi-martingale $\mathbf{X}_{(\cdot)}$.
\end{definition}
}}

Definition~\ref{def:ito_flow_measure} is a pivotal tool in our paper as it offers a closed form for the upper bounds of our variational equation. The following variation equation clearly delineates that the normalized entropy is influenced by fluctuations of Wasserstein metric. Now, we are ready to derive our Wasserstein variation equation of functional $\mathcal{E} = \mathcal{H}_t^N$ with aforementioned notions:

\colorbox{rgb:red!10,0.5;green!10,0.5;blue!10,0.5}
{\parbox{0.98\linewidth}{
\begin{manualtheorem}{3.1}[\textit{Variation Equations for $N$-particle Relative Entropy}]
For arbitrary temporal variables $0 \leq s < t \leq T$, there exist constants $\mathrm{C}_{0}, \mathrm{C}_{1} > 0$ satisfying the following variational equation:
\begin{equation}
    \mathcal{H}^N_t(\nu_t^{N}) \leq \mathcal{H}^N_s(\nu_s^{N}) + \mathrm{C}_{0} \int_s^t \mathcal{O}\left(\mathbb{E}\norm{\nabla_{\mathcal{P}_2}\mathcal{H}^N_u}^{2}_E\right)du + \mathrm{C}_{1} \int_s^t \mathcal{O}\left( \mathbb{E}\norm{\nabla_x\nabla_{\mathcal{P}_2}\mathcal{H}^N_u}^{2}\right) du.
\end{equation} 
\end{manualtheorem}
}}
\begin{proof}
We start by deriving the proposed score-matching objective. Let us consider a semi-martingale $\nu_t \sim \mathbf{X}_t$, $d\mathbf{X}_t = - \nabla V dt + \Sigma_t dW_t$ for $V \coloneqq V^1$ in Eq.~\ref{eq:appendix_V_N} with progressively measurable processes $f_t, \nabla \log \zeta_{T-t}$. We define the time-varying energy functional $\mathcal{E}$ as relative entropy
\begin{equation}
    \mathcal{E}(t, \mu_t) = \mathcal{H}(\mu_t | \zeta_{T - t}) \coloneqq \mathcal{H}(t, \mu_t) \coloneqq \mathcal{H}_t    
\end{equation}
With the notation $\nu_t^N = \mathbf{Law}(\mathbf{X}_t^N)$, the functional $\mathcal{H}_t$ evolves with differential calculus by It\^o's flow of measures introduced in Definition~\ref{def:ito_flow_measure} associated with Wasserstein gradient flow in Eq.~\ref{eq:path_measures}:
\begin{equation}
    d\mathcal{H}(t, \nu_t^N) = \mathbb{E}\left[ \nabla_x\frac{\partial \mathcal{H}}{\partial \delta}(t, \nu_t^N) \cdot d\mathbf{X}^N_t\right] + \mathbb{E}\left[\frac{1}{2}\mathbf{Tr}\left(\nabla_x^2\frac{\partial \mathcal{H}}{\partial \delta}(t, \nu_t^N) \cdot d[\mathbf{X}^N_t, \mathbf{X}^N_t]\right)\right].
\end{equation}
where $\nabla, \nabla^2$ are gradient and Hessian operators, and the expectation is taken with respect to the law of semi-martingale $\mathbf{X}_{(\cdot)}$. Then, the direct application of variation equation in Definition~\ref{def:ito_flow_measure} to entropy $\mathcal{H}_t$ gives
\begin{equation}\label{eq:appendix_ito_flows_1}
\begin{split}
    \mathcal{H}_t &= \mathcal{H}_s +  \int_s^t\mathbb{E}_{\nu_u^N}\left[\frac{1}{N} \nabla \left(\log \frac{\varrho_u^N}{\zeta_{T-u}^{\otimes N}} \right) \cdot d\mathbf{X}_u^N \right]du 
    + \int_s^t \mathbb{E}_{\nu_u^N}\left[\frac{1}{2N}\nabla^2\left(\log \frac{\varrho_u^N}{\zeta_{T-u}^{\otimes N}}\right) d[\mathbf{X}^N_u, \mathbf{X}^N_u]^T\right]
    \\ & \leq \mathcal{H}_s +  \int_s^t\mathbb{E}_{\nu_u^N}\left[\frac{1}{N}\left| \nabla \left(\log \frac{\varrho_u^N}{\zeta_{T-u}^{\otimes N}} \right) \cdot \nabla V^N \right|\right]du 
    + \int_s^t \mathbb{E}_{\nu_u^N}\left[\frac{1}{2N}\left|\nabla^2\left(\log \frac{\varrho_u^N}{\zeta_{T-u}^{\otimes N}}\right) \cdot \left(\int_s^t \Sigma_u \Sigma_u^T du\right) \right|\right]
    \\ & \leq \mathcal{H}_s + \int_s^t \mathbb{E}_{\nu_u^N}\left[\frac{1}{N}\norm{\nabla \left(\log \frac{\varrho_u^N}{\zeta_{T-u}^{\otimes N}}\right)}_E\norm{\nabla V^N}_E\right]du + \int_s^t \mathbb{E}_{\mu_u}\left[ \frac{1}{2N}\norm{\nabla^2 \left(\log \frac{\varrho_u^N}{\zeta_{T-u}^{\otimes N}} \right)}_F \norm{\left( \int_s^t \Sigma_u \Sigma_u^{T} du\right)}_F\right],
\end{split}
\end{equation}
where $\nabla, \nabla^2$ denote Euclidean gradient and Hessian operators with respect to the spatial axis, and $\norm{\cdot}_F$ is Frobenius norm. The first equality holds as the Wasserstein gradient is identified with
spatial gradient of the first variation. Note that first variation of entropy-type functionals can be directly obtained from Section 8.2~\cite{santambrogio2015optimal}.
\begin{align}
    & \frac{\partial \mathcal{H}[\nu_t^{N} | \zeta_{T - t}^{\otimes N}]}{\partial \delta}(\mathbf{x}^N) = \log \varrho_t^{N}(\mathbf{x}^N)  - \log \zeta_{T - t}^{\otimes N}(\mathbf{x}^N)  + 1,
    \\ & \nabla \frac{\partial \mathcal{H}[\nu_t^{N} | \zeta_{T - t}^{\otimes N}]}{\partial \delta}(\mu) 
     = \nabla_{\mathcal{P}_2}\mathcal{H}_t[\nu_t^{N}] = \nabla \log \varrho_t^N(\mathbf{x}^N) - \nabla \log \zeta_{T - t}^{\otimes N}(\mathbf{x}^N), \\
    & \nabla_x \nabla_{\mathcal{P}_2}\mathcal{H}_t[\nu_t^N] = \nabla^2 \log \varrho_t^N(\mathbf{x}^N) - \nabla^2 \log \zeta_{T - t}^{\otimes N}(\mathbf{x}^N).
\end{align}
For the deterministic log-probabilities $\log \varrho_t$ and $\log \zeta_{T-t}$, the expectation of martingale terms vanishes
\begin{align*}
    &\mathbb{E}\left[\nabla \log \varrho_{u}^N \Sigma_u dW_u\right] = \mathbb{E}\mathbb{E}\left[\nabla \log \varrho_{u}^N \Sigma_u dW_u | \mathcal{F}_u\right] = 0, \\ &\mathbb{E}\left[-\nabla \log \zeta_{T-u}^{\otimes N} \Sigma_u dW_u\right] = \mathbb{E}\mathbb{E}\left[-\nabla \log \zeta_{T-u}^{\otimes N} \Sigma_u dW_u | \mathcal{F}_u\right]  = 0.
\end{align*}
For the time-varying diffusion matrix $\Sigma_t$, quadratic variation can be calculated as
\begin{equation}
    d[\mathbf{X}^N_{(\cdot)}, \mathbf{X}^N_{(\cdot)}]^T= \left(\int \Sigma_{(\cdot)} \Sigma_{(\cdot)}^T dt\right)^T = \int (\Sigma_{(\cdot)} \Sigma_{(\cdot)}^T)^T dt = \int (\Sigma_{(\cdot)} \Sigma_{(\cdot)}^T) dt, \quad \Sigma_{(\cdot)} \Sigma_{(\cdot)}^T \in \mathbf{Sym}(d).
\end{equation}

Let us define $\mathcal{X}$-valued function $\mathcal{G}_t = \mathbf{s}_{\theta} - \nabla \log \zeta_{T-t}$. Recall the definition of weighted Sobolev space, and its canonical norm with respect to multi-index $\boldsymbol{\alpha}$, recall the definition of the norm on the weighted Sobolev space $W_{\alpha, p}^{w}(\mathcal{X}^N)$ as
\begin{equation}
    \norm{\mathcal{G}_t}_{W^{\boldsymbol{\alpha}}_p} = \left(\int \norm{\mathcal{G}_t}_E^p w_{0} d\nu_t\right)^{1/p} + \sum_{\boldsymbol{\alpha}}\left(\int \norm{D^{\boldsymbol{\alpha}}\mathcal{G}_t}^{\boldsymbol{\alpha}} w_{\boldsymbol{\alpha}} d\nu_t\right)^{1/p}.
\end{equation}
where $D^{\alpha}$ stands for higher-order weak partial derivatives at most $L$ degree $D^{\boldsymbol{\alpha}}\varphi = \partial^{L}\varphi/\partial\mathbf{x}_1^{\alpha_1} \cdots \partial\mathbf{x}_L^{\alpha_L}$ defined as:
\begin{align}
    \int u D^{\boldsymbol{\alpha}}\varphi d\mathbf{x}^N = (-1)^{K \leq |\boldsymbol{\alpha}|} \int \varphi D^{\boldsymbol{\alpha}}u d\mathbf{x}^N.
\end{align}   

With aforementioned notations and definitions for $|\boldsymbol{\alpha}| = 1, p = 2$, the right-hand side can be rewritten by the weighted Sobolev norm.
\begin{equation}\label{eq:appendix_Corollary_1}
\begin{split}
    \mathcal{H}_t & \leq \mathcal{H}_s + \int_s^t \left(\int \norm{\nabla \log \varrho_u^N - \nabla \log \zeta_{T - u}^{\otimes N} }^2_E w_0(\mathbf{x}^N)d\nu_u^N(\mathbf{x}^N)\right)du 
    \\ & \qquad \qquad \qquad + \int_s^t \left( \norm{ \nabla^2 \log \varrho_u^N - \nabla^2 \log \zeta_{T - u}^{\otimes N}}_{F}^2 w_1(\mathbf{x}^N) d\nu_u(\mathbf{x}^N)\right) du
    \\ & = \mathcal{H}_s + \int_s^t \norm{\mathcal{G}}_{W_{1, 2}^{w}} du,
\end{split}
\end{equation}
with following weight functions $w_0, w_1$:
\begin{align}
    & w_0(t, \mathbf{x}^N) = \norm{\nabla V^N}_E, \quad w_1(t) = \mathbb{E}\left[\int_0^T \norm{\Sigma_t \Sigma_t^{T}}_F dt\right] = \int_0^T \norm{\Sigma_t \Sigma_t^{T}}_F dt.
\end{align}
To simplify the weighted norm to derive Eq.~\ref{eq:flow_measures_primitive_main}, we apply Hölder's inequality to the first term in the last line of Eq.~\ref{eq:appendix_ito_flows_1}.
\begin{equation}
    \mathbb{E}_{\nu_t^N}\left[ \norm{\frac{1}{\sqrt{N}}\nabla \left(\log \frac{\varrho_u^N}{\zeta_{T-u}^{\otimes N}}\right)}_E\norm{\frac{1}{\sqrt{N}}\nabla V^N}_E \right] \leq \mathrm{C}_0\left(\mathbb{E}_{\nu_t^N}\left[ \norm{ \frac{1}{N} \nabla \left(\log \frac{\varrho_u^N}{\zeta_{T-u}^{\otimes N}}\right)}^2_E \right]\right)^{1/2}.
\end{equation}
The constant $\mathrm{C}_0$ can be controlled by
\begin{equation}
\begin{split}
    \mathrm{C}_0 & = \frac{1}{\sqrt{N}} \left[\int w_0^2(t, \mathbf{\mathbf{X}}_t^N) d\nu_t(\mathbf{X}_t^N)\right]^{1/2}
    \\ & \leq \frac{1}{\sqrt{N}} \left[\frac{\beta_t}{2}\mathbb{E}_{\nu_t^N}\norm{\mathbf{X}_t^N}^2_E + \mathbb{E}_{\nu_t^N}\norm{\log \zeta_{T - t}^{\otimes N}(\mathbf{X}_t^N)}^2_E\right]^{1/2}
    \\ & \leq \frac{1}{\sqrt{N}}\left[\sup_{t \in [0, T]}\left(\frac{\beta_t}{2} + \frac{1}{\sigma_{\zeta}(T-t)}\right) \sup_{t \in [0, T]} \mathbb{E}_{\nu_t^N}\norm{\mathbf{X}_t^N}^2_E + \mathbb{E}\norm{\mathbf{Y}^N}^2_E \sup_{t \in [0, T]} \frac{1}{N}\sum_j^{N}\frac{\mathbf{m}_{\zeta}(T-t) }{\sigma_{\zeta}(T-t)}\right]^{1/2}
    \\ &  \leq \frac{1}{\sqrt{N}}\left[\mathrm{C}_3 \mathrm{C}_2 e^{\mathrm{C}_2 T}(1 + \mathbb{E}\norm{\mathbf{X}_0^N}^2_E)
    + \mathbb{E}\norm{\mathbf{Y}^N}^2_E \mathrm{C}_4 \right]^{1/2}
    \\ & \leq \left[\frac{\cancel{\mathrm{C}_3 \mathrm{C}_2 e^{\mathrm{C}_2 T}(1} + d\cancel{N})
    + \cancel{N}\mathcal{M}^2(2, \zeta_0)\mathrm{C}_4}{\cancel{N}}\right]^{1/2} 
    \\ & \overset{N \gg d}{\approx}\sqrt{d} + \frac{1}{2\sqrt{d}}\mathcal{M}^2(2, \zeta_0)
\end{split}
\end{equation}
where we denote by $\mathcal{M}(r, \nu)$ the $r$-th moment of measure $\nu$, and used the fact $\mathbb{E}\norm{\mathbf{Y}^N}_E^2 = N \mathbb{E}|\mathbf{Y}|^2 = N\mathcal{M}^2(2, \zeta_0)$. The constants $\mathrm{C}_3$, $\mathrm{C}_4$ are dependent on the choice of hyperparameters $\beta_{\text{min}}$ and $\beta_{\text{max}}$. 
\begin{align}
    & \mathrm{C}_3 = \sup_{t \in [0, T]} \left(\frac{\beta_t}{2} + \frac{1}{\sigma_{\zeta}(T-t)}\right), \quad \mathrm{C}_4 = \sup_{t \in [0, T]}\frac{\mathbf{m}_{\zeta}(T-t)}{\sigma_{\zeta}(T-t)} \ll 1.
\end{align}

Applying Hölder's inequality again for the quadratic variation term, we have the following upper bound:
\begin{equation}
    \mathbb{E}_{\nu_u^N}\left[ \norm{\frac{1}{\sqrt{2N}} \nabla^2 \left(\log \frac{\varrho_u^N}{\zeta_{T-u}^{\otimes N}} \right)}_F \norm{\frac{1}{\sqrt{2N}} \left( \int_s^t \Sigma_u \Sigma_u^{T} du\right)}_F\right] \leq \mathrm{C}_1\left(\mathbb{E}_{\nu_u^N}\left[ \norm{\frac{1}{2N}\nabla^2 \left(\log \frac{\varrho_u^N}{\zeta_{T-u}^{\otimes N}} \right)}_F^2 \right]\right)^{1/2},
\end{equation}
where $\mathrm{C}_1$ can be bounded by
\begin{equation}
\begin{split}
    \mathrm{C}_1 & = \frac{1}{\sqrt{2N}}\int \norm{\Sigma_s \Sigma_s^T}_F ds \leq \frac{1}{\sqrt{2}}\cancel{\frac{1}{\sqrt{N}}}T\sup_{t \in [0, T]}\sqrt{d} \cancel{\sqrt{N}} \beta_t = \sqrt{\frac{d}{2}}T((1-T)\beta_{\text{min}} + T\beta_{\text{max}})
\end{split}
\end{equation}
By using these results, Eq.~\ref{eq:appendix_Corollary_1} can be further improved as follows:
\begin{equation}\label{eq:appendix_Corollary_2}
\begin{split}
    \mathcal{H}_t & \leq \mathcal{H}_s \int \norm{\mathcal{G}_t}_{W^{1, 2}_{w}}  dt
    \\ & \leq \mathcal{H}_s + \mathrm{C}_0 \int_s^t \frac{1}{\sqrt{N}}\mathbb{E}\norm{\nabla \log \varrho_t^N - \nabla \log \zeta_{T-t}^{\otimes N}}_{E}^2 dt + \mathrm{C}_1 \int_s^t \frac{1}{\sqrt{N}} \mathbb{E}\norm{\nabla^2 \log \varrho_t^N - \nabla^2 \log \zeta_{T-t}^{\otimes N}}_{F}^2 dt.
\end{split}
\end{equation}
By rewriting the inequality above, the proof is complete.
\end{proof}

\begin{remark}
Following by the Sobolev embedding theorem~\cite{brezis2011functional}, it is trivial to observe that the Sobolev space can be embedded into $L^{6}$-space, $\ie W^{1,2} \hookrightarrow L^{6}$, assuring a  lower-bound $\norm{\mathcal{G}_t}_{L^{6}} \leq \mathrm{S}\norm{\mathcal{G}_t}_{W^{1,2}}$ with numerical constant $\mathrm{S} > 0$ related to the diameter of $\Omega_{\mathcal{X}}$, when one restricts on the open and bounded subset $\Omega_{\mathcal{X}} \subset \mathcal{X}$. Since Hölder's inequality naturally gives another embedding $L^{6} \hookrightarrow L^2$, the chain of two embeddings bridges the gap between conventional score-matching and the proposed MF-SM.    
\end{remark} 

\colorbox{rgb:red!10,0.5;green!10,0.5;blue!10,0.5}
{\parbox{0.98\linewidth}{
\begin{manualcorollary}{3.2}{\textit{(Sobolev Score Matching)}}  Let $\norm{\cdot}_{W}$ be a norm defined on Sobolev space $W^{1, 2}(\mathcal{X}^N, \nu_t^N)$ and $\mathcal{M} \coloneqq \mathcal{M}(\zeta_0) < \infty$ be a second moment of target data instance $\mathbf{Y} \sim \zeta_0$. Then, we have the following 
\begin{equation}
    \mathcal{H}_T^N(\nu_t^N) \precsim \frac{\mathcal{M}}{\sqrt{Nd}}\int_0^T \norm{\nabla \log \varrho_{t}^N - \nabla \log \zeta_{T - t}^{\otimes N}}_W dt.
\end{equation}
\end{manualcorollary}
}}
\begin{proof}
The proof is direct consequence from Theorem~\ref{prop:objective_flows} and the subsequent inequalities:
\begin{equation}\label{eq:appendix_Corollary_3}
\begin{split}
    \mathcal{H}^N_T & \leq \int_0^T \norm{\mathcal{G}_t}_{W^{1, 2}_{w}}  dt + \mathcal{H}_0^N
    \\ & \leq \mathrm{C}_0 \int_0^T \frac{1}{\sqrt{N}}\mathbb{E}\norm{\nabla \log \varrho_t^N - \nabla \log \zeta_{T-t}^{\otimes N}}_{E}^2 dt + \mathrm{C}_1 \int_0^T \frac{1}{\sqrt{N}} \mathbb{E}\norm{\nabla^2 \log \varrho_t^N - \nabla^2 \log \zeta_{T-t}^{\otimes N}}_{F}^2 dt
    \\ & \leq \frac{2(\mathrm{C}_0 \wedge \mathrm{C}_1)}{\sqrt{N}} \int_0^T \norm{\mathcal{G}_t}_{W^{1, 2}}^2 dt 
    \\ & \precsim \frac{2}{\sqrt{N}}\left(\sqrt{d} + \frac{1}{2\sqrt{N d}}\mathcal{M}^2(2, \zeta_0) \wedge [\beta_{\text{min}}T(1-T) + T^2\beta_{\text{max}}]^2 \right) \int_0^T \norm{\mathcal{G}_t}_{W^{1, 2}}^2 dt
    \\ & \approx \frac{\mathcal{M}^2(2, \zeta_0)}{\sqrt{Nd}}\int_0^T \norm{\mathcal{G}_t}_{W^{1, 2}}^2 dt 
    \\ & = \frac{\mathcal{M}^2(2, \zeta_0)}{\sqrt{Nd}}\int_0^T \norm{\nabla \log \varrho_{t}^N - \nabla \log \zeta_{T - t}^{\otimes N}}_{W^{1,2}}^2 dt \xrightarrow{N \rightarrow \infty} 0,
\end{split}
\end{equation}
where we assume $T = 1.0, d = 3, \mathcal{M}^2 \gg \beta_{\text{max}} \wedge d$ in the last line. The relative entropy of an $N$-particle system approaches zero when the right-hand side of Eq.~\ref{eq:appendix_Corollary_3} also converges to zero. With the fact that $\mathcal{H}_0^N = 0$ by the assumptions on initial states of dWGFs, the proof is complete.
\end{proof}

\subsection{Variation Equation in Infinite Particle System}

\textbf{Time-inhomogenous Markov Process for $N$-particle System.} While the proposed denoising MF VP-SDEs are modeled as time-inhomogenous Markovian dynamics, this section starts by 
 providing basic materials for further understanding and analysis of the asymptotic behavior of proposed MF-SDEs/dWGFs. Given the structure of MF-SDEs with joint density $\varrho_t^N$, the entire $N$-particle system possesses a $\mathcal{P}_N(\mathcal{X})$-valued Markovian property, where its semi-group and infinitesimal generator are given by:
\begin{align}
    & \mathcal{L}_t^N \varrho_t^N(\mathbf{x}^N) = \sum_i^N \mathcal{L}_t^i\varrho_t^N(\mathbf{x}_1, \cdots, \mathbf{x}_i, \bar{\mathbf{x}}_{i+1}, \cdots \bar{\mathbf{x}}_N)(\mathbf{x}), \\
    & \mathcal{L}_t^i\varrho_t^{i, N} = \nabla_{\mathcal{P}_2}\mathcal{E}_t\varrho_t^{i, N} =  -\nabla\varrho_t^{i, N} \cdot \partial_{\mathbf{x}_i} \nabla V^N -\frac{\beta}{2} \partial^2_{\mathbf{x}_i} \varrho_t^{i, N}.
\end{align}
Note that the Liouville equation in Sec~\ref{sec:wgf} representing the probabilistic formulation of MF-SDEs is based on the infinitesimal generator defined as above. For the function families $f, g \in \mathbf{Dom}(\mathcal{L}_t^N)$, we associate the infinitesimal generator with its first and second order carr\^e du champ operator~\cite{bakry1997sobolev}    $\boldsymbol{\Gamma}$, $\boldsymbol{\Gamma}_2$ defined by 
\begin{align}
    & \boldsymbol{\Gamma}(t)(f, g) \coloneqq \frac{1}{2}\left( \mathcal{L}_t^N(f g ) - f \mathcal{L}_t^N g - g \mathcal{L}_t^N f \right), \\
    & \boldsymbol{\Gamma}_2(t)(f) \coloneqq \frac{1}{2}\left( \mathcal{L}_t^N \boldsymbol{\Gamma}(f) - 2\boldsymbol{\Gamma}(f, \mathcal{L}_t^N f)\right).
\end{align}
Recall that we say that the diffusion $\mathcal{L}^N$ for probability measure of time-homogeneous Markov process enjoys the logarithmic Sobolev inequality: $\boldsymbol{\Gamma}_2(f) \geq \upsilon \boldsymbol{\Gamma}(f, f)$ for arbitrary $\upsilon \in \mathbb{R}^{+}$. The goal is to generalize this type of functional inequality to time-inhomogenous dynamics. For this, consider a diffusion process, which has a infinitesimal generator $\mathcal{L}_t$ as follows:
\begin{equation}
    \mathcal{L}_t^1 f = \sum_{a, b \leq d} [\sigma \sigma^T]_{ab}(t)\partial_{ab}f + \sum_{a \leq d} v_i(t, \mathbf{x})\partial_a f,
\end{equation}
where the infinitesimal generator $\mathcal{L}_t^{1}$ is associated with SDE of following type:
\begin{equation}
    d\mathbf{X}_t^1 = v(t, \mathbf{X}_t^1, \nu_t^1)dt + \sigma(t) dW_t.
\end{equation}
Let $\mathbf{P}_t^N(\mathbf{x}) = \mathbb{E}[\mathbf{X}_t^N | \mathbf{X}_0 = \mathbf{x}]$ be a semi-group related to $\mathcal{L}_t^N$. By direct calculations, first and second-order carr\^e du champ operators can be estimated as
\begin{align}\label{eq:appendix_carre_estimations}
    & \boldsymbol{\Gamma}(t)(f,f) = [\sigma \sigma^T](t)\norm{\nabla f}_E^2, 
    \\ & \boldsymbol{\Gamma}_2(t)(f) = \norm{\nabla^2 f }_F^2 - \nabla f \cdot \boldsymbol{J}(v_t)\nabla f, 
    \\ & \partial_t \boldsymbol{\Gamma}(t)(f) = \partial_t [\sigma \sigma^T](t)\norm{\nabla f}_E^2, \quad \quad f \in \mathbf{Dom}(\mathcal{L}_t^N),
\end{align}
where $\boldsymbol{J}$ denotes a Jacobian operator. Then, the time-inhomogeneous semigroup $\mathbf{P}_t$ is said to satisfy log-Sobolev inequality if Bakry-Émery criterion in Eq.~\ref{eq:appendix_bakry_emery_criterion} holds for any suitable $f$:
\begin{equation}\label{eq:appendix_bakry_emery_criterion}
    \boldsymbol{\Gamma}_2(t)(f) + \frac{1}{2}\partial_t \boldsymbol{\Gamma}(t)(f) \geq \kappa(t)\boldsymbol{\Gamma}(t)(f),
\end{equation}

\textbf{Generalized Logarithmic Sobolev inequality.} Under the conditions desribed in Eq.~\ref{eq:appendix_bakry_emery_criterion}, Theorem 3.10~\cite{collet2008logarithmic} ensures the existence of $\Phi$-logarithmic Sobolev inequality.
\begin{equation}\label{eq:appendix_phi_log_sobolev}
    \mathbf{Ent}^{\Phi}_{\nu_t}(g) \leq c(t)P_{t}\left(\Phi''(g) \boldsymbol{\Gamma}(t)(g) \right), \quad c(t) = \int_0^t \exp \left( -2 \int_v^t \kappa(u) du \right) dv,
\end{equation}
where $\Phi : \mathbb{R} \rightarrow \mathbb{R}$ is a smooth convex function and the $\Phi$-entropy is given by
\begin{equation}
    \mathbf{Ent}^{\Phi}_{\nu}(f) = \int \Phi(f)d\mu - \Phi \int f d\nu.
\end{equation}

Define $LS^{\Phi}[c(t)]$ with respect to $c(t)$ in Eq.~\ref{eq:appendix_phi_log_sobolev} as the constant related to the generalized $\Phi$-log Sobolev inequality. Then the constant $LS^{\Phi}$ associated with product measure is readily derived using the subsequent result:

\colorbox{rgb:red!10,0.5;green!10,0.5;blue!10,0.5}
{\parbox{0.98\linewidth}{
\begin{lemma}\label{lemma:stability_product_LS}(Stability under product)~\cite{bakry2014analysis} If $(\mathcal{X}^N,\mu_1, \mathcal{L}_t^{1, N})$ and $(\mathcal{X}^N, \mu_2, \mathcal{L}_t^{2, N})$ satisfy logarithmic Sobolev inequalities $LS^{\Phi}[c_1(t)]$ and $LS^{\Phi}[c_2(t)]$ respectively, then
the product $(\mathcal{X}^{N} \times \mathcal{X}^{N}, \mu_1 \otimes \mu_2, \mathcal{L}_t^{1, N} \oplus \mathcal{L}_t^{2, N})$ satisfies a logarithmic Sobolev inequality $LS^{\Phi}[\max(C_1(t),C_2(t))]$.
\end{lemma}
}}

By the result of Lemma~\ref{lemma:stability_product_LS}, it is straightforward to show that $N$-product of Gaussian measures in forward noising process $\zeta_{t}^{\otimes N}$ preserve the log-Sobolev constant of its single component $\zeta_t$.

\colorbox{rgb:red!10,0.5;green!10,0.5;blue!10,0.5}
{\parbox{0.98\linewidth}{
\begin{theorem}\label{theorem:HWI}(HWI inequality)~\cite{otto2000generalization} Let $d\nu \propto e^{- W}d\mathbf{x}$ be a probability measure on $\mathcal{X}$, with finite second moments, such that $W \in C^{2}(\mathcal{X})$, $\nabla^2 W \succeq \kappa \mathbf{I}_d$, $\kappa \in \mathbb{R}$. Then, $\nu$ satisfies the log-Sobolev inequality with constant $LS(\kappa, \infty)$. For any other absolutely continuous measures $\nu_0$, the following inequality holds:
\begin{equation}
    \tilde{\mathcal{H}}(\nu_0 | \nu) \leq \mathcal{W}_2(\nu_0, \nu)\sqrt{\tilde{\mathcal{I}}(\nu_0 | \nu)} - \frac{\kappa}{2}\mathcal{W}_2^2(\nu_0, \nu).
\end{equation}
The inequality above equally indicates that
\begin{equation}
    \tilde{\mathcal{H}}(\nu_0 | e^{-W}) - \tilde{\mathcal{H}}(\nu_1 | e^{-W}) \leq \mathcal{W}_2(\nu_0, \nu_1)\sqrt{\tilde{\mathcal{I}}(\nu_0 | e^{-W})} - \frac{\kappa}{2}\mathcal{W}^2_2(\nu_0, \nu_1).
\end{equation} $\tilde{\mathcal{H}},\tilde{\mathcal{I}}$ denotes non-normalized relative entropy and relative Fisher information, respectively. 
\end{theorem}
}}

\begin{remark}
It should be emphasized that the functionals described in Theorem~\ref{theorem:HWI} are presented in non-normalized forms while the $N$-particle entropy in Eq.~\ref{eq:relative_entropy_N} is defined as its normalized counterpart. This distinction in notation, while subtle, is made explicit in the context and is intentionally simplified here for brevity.
\end{remark}

\colorbox{rgb:red!10,0.5;green!10,0.5;blue!10,0.5}
{\parbox{0.98\linewidth}{
\begin{manualproposition}{3.4} The $N$-particle entropy for infinity cardinality $N \rightarrow \infty$ have upper bound as
\begin{equation}
    \mathcal{H}_T(\mu_{T}) \leq \frac{\mathcal{M}}{\sqrt{Nd}}\mathcal{J}_{MF}^N( \theta, [0, T]) + \kappa_{\zeta}\mathcal{O}\left(\frac{\mathrm{C}_2}{N}  + \frac{\mathrm{C}_3}{N^{1/2}} + \frac{\mathrm{C}_4}{N^{3/2}}\right) \xrightarrow{N \rightarrow 0} 0.
\end{equation}
We define numerical constants $\mathrm{C}_2 \coloneqq \mathrm{C}_2[\beta_T, \mathrm{C}_B, \sigma_{\zeta}(T)]$, $\mathrm{C}_3 \coloneqq \mathrm{C}_3[\mathrm{D}, \sigma_{\zeta}(T), \mathcal{M}, \beta_T, \mathbf{m}_{\zeta}(T)]$, where each $\beta_T, \mathrm{C}_B, \sigma_{\zeta}, \mathrm{D}, \mathcal{M}(2, \zeta_0), \mathbf{m}_{\zeta}$ is independent on the data cardinality $N$.
\end{manualproposition}
}}
\begin{proof} 
For any fixed $t \in [0, T]$, let us repurpose the stationary density of the time-varying Ornstein-Uhlenbeck process for VP SDE. 
\begin{equation}
    \mathfrak{m}(\mathbf{x}) \propto e^{-W_{\zeta}(t, \mathbf{x})} = e^{-\sum_j^N \kappa_{\zeta}\norm{\mathbf{x}_j - \mathbf{Y}\mathbf{m}_{\zeta}(t)}^2_E},
\end{equation}
where we denote $\kappa_{\zeta} = \sigma_{\zeta}^{-2}(t)$.
Following direct calculation, one has
\begin{equation}
    \nabla^2 W_{\zeta}(t, \mathbf{x}_j) \succeq
\kappa_{\zeta} \mathbf{I}_d, \quad \quad \nabla^2 W^{\otimes N}_{\zeta}(t, \mathbf{x}^N) \succeq \kappa_{\zeta} \mathbf{I}_{Nd},
\end{equation}

Consider $\mu_t$ as a solution to Liouville equation in Eq.~\ref{eq:equivalence} for limitation $N \rightarrow \infty$, and let $\mu_t^{\otimes N}$ be a $N$-product of $\mu_t$. For any $N \in |\mathbb{N}|$, the normalized variant of HWI inequality in Theorem~\ref{theorem:HWI} shows the following inequality holds for any $N$:
\begin{equation}\label{eq:appendix_HWI_adapted}
\begin{split}
    N e_N \coloneqq N\mathcal{H}(\mu_t^{\otimes N} | \mathfrak{m} d\mathbf{x}^N) - N\mathcal{H}(\nu_t^{N} | \mathfrak{m} d\mathbf{x}^N)
    \leq \underbrace{\mathcal{W}_2(\nu_t^{N}, \mu_t^{\otimes N})}_{\mathbf{(A)}}\underbrace{\sqrt{\mathcal{I}(\nu_t^N | \mathfrak{m} d\mathbf{x}^N)}}_{\mathbf{(B)}} - \underbrace{\frac{\kappa_{\zeta}}{2 }\mathcal{W}_2^2(\nu_t^N, \mu_t^{\otimes N})}_{\mathbf{(A')}}
\end{split}
\end{equation}
We first derive the upper bounds of $\mathbf{(B)}$ by estimating $\mathcal{I}$, which stands for the relative fisher information. Assuming $\mathbf{s}_{\theta}^{\otimes N} \in \mathbf{Ker}(\mathcal{G})$ and Eq.~\ref{eq:appendix_NN_score}, we have
\begin{equation}\label{eq:appendix_fisher_inform_bound}
\begin{split}
    \mathcal{I}(\nu_t^N | \mathfrak{m} d\mathbf{x}^N) & \coloneqq \int \norm{\nabla \log \frac{\varrho_t^N}{e^{-W}} }^2_E d\nu_t^N
    \\ & \leq \int \norm{\nabla \log \varrho_t^N}^2_E d\nu_t^N + \sum_j^N \int \norm{\nabla W_{\zeta}}^2_E d\nu_t^N 
    \\ & \leq \mathrm{D}(1 + \norm{\mathbf{X}_t^N}_E^2) + 4\kappa_{\zeta}^2 \left( \mathbb{E}\norm{\mathbf{X}_t^{N}}_E^2 + N \mathcal{M}^2(2, \zeta_0)\mathbf{m}_{\zeta}^2(t) \right)
    \\ & \leq \mathrm{D} + 4\kappa_{\zeta}^2 \left[(1 + \mathrm{D})[N \mathcal{M}^2(2, \zeta_0) + N d \beta_T^2 + \mathrm{D}]e^{\mathrm{D}} + N \mathcal{M}^2(2, \zeta_0)\mathbf{m}_{\zeta}^2(t)\right]
\end{split}
\end{equation}

As a next step, we investigate the upper bound of 2-Wasserstein distance involved in $\mathbf{(A)}$ and $\mathbf{(A')}$. First, we define two dynamics $(\mathbf{X}_t^{N}, \bar{\mathbf{X}}_t^N)$ as
\begin{equation}
(\mathbf{X}_t^{N}, \bar{\mathbf{X}}_t^N) = ~
\begin{cases}
    d\mathbf{X}_t^{N} = f^{\otimes N}_t(\mathbf{X}_t^{N}) dt - \beta_t \mathbf{s}_{\theta}(t, \mathbf{X}_t^{N}, \nu_t^N)dt + \sqrt{\beta_t} dB_t^N, \\
    d\bar{\mathbf{X}}_t = f^{\otimes N}_t(\bar{\mathbf{X}}_t^N) dt - \beta_t \mathbf{s}_{\theta}(t, \bar{\mathbf{X}}_t^N, \mu_t^{\otimes N})dt + \sqrt{\beta_t} dB_t^N.
\end{cases}
\end{equation}

By using It\^o's formula and Burkholder-Davis-Gundy inequality, one can induce that
\begin{equation}\label{eq:appendix_sup_X_X_bar}
\begin{split}
    \mathbb{E}\left[\sup_{t} \norm{\mathbf{X}_t^{N} - \bar{\mathbf{X}}_t^N}_E^2 \right] & \leq 2 T \int_{0}^{T} \beta_t^2\mathbb{E} \norm{\mathbf{s}^{\otimes N}_{\theta}(t, \mathbf{X}_t^{N}, \nu_t^N) - \mathbf{s}_{\theta}^{\otimes N}(t, \bar{\mathbf{X}}_t^N, \mu_t^{\otimes N})}^2_E dt
    \\ &\leq 4 T \sup_{t \in [0, T]}\beta_t^2 \bigg( \int_0^T \mathbb{E}\norm{\mathbf{s}^{\otimes N}_{\theta}(t, \bar{\mathbf{X}}_t^N, \nu_t^N) - \mathbf{s}^{\otimes N}_{\theta}(t, \bar{\mathbf{X}}_t^N, \mu_t^{\otimes N})}^2_E dt 
    \\ & \qquad \qquad + \int_0^T \mathbb{E}\norm{\mathbf{s}^{\otimes N}_{\theta}(t, \bar{\mathbf{X}}_t^N, \nu_t^N) - \mathbf{s}^{\otimes N}_{\theta}(t, \mathbf{X}_t^N, \nu_t^{N})}^2_E dt\bigg)
    \\ & \leq 4T\beta_T^2 \left( \mathrm{C}_{\mathrm{B}} \int_{0}^{T} \mathcal{W}_2^2(\nu_t^N, \mu_t^{\otimes N})dt + 2( \mathrm{C}_{\mathrm{A}} \wedge \mathrm{C}_{\mathrm{B}})\int_{0}^{T} \mathbb{E}\left[\sup_{s \leq t}\norm{\bar{\mathbf{X}}_t^N - {\mathbf{X}_t^N}}_E^2\right]dt\right)
\end{split}
\end{equation}
With the fact that $\mathbf{Lip}(\mathbf{s}_{\theta}) = \mathbf{Lip}(\mathbf{s}^{\otimes N}_{\theta})$, and applying Grönwall's Lemma gives
\begin{equation}\label{eq:appendix_W2_sup}
\begin{split}
    \sup_{t} \mathcal{W}^2_2(\nu_t^N, \mu_t^{\otimes N}) & \leq \mathbb{E}\left[\sup_{t} \norm{\mathbf{X}_t^{N} - \bar{\mathbf{X}}_t^N}_E^2 \right] 
    \\ & \leq 4 \beta_T^2 T \mathrm{C}_{\mathrm{B}} \exp(8 \beta_T^2 T^2( \mathrm{C}_{\mathrm{A}} \wedge \mathrm{C}_{\mathrm{B}}))\left(\int_{0}^{T} \mathcal{W}_2^2(\nu_t^N, \mu_t^{\otimes N}) dt\right)     
    \\ & \leq \mathfrak{a} + 4 \beta_T^2 T \mathrm{C}_{\mathrm{B}}\exp(8 \beta_T^2 ( \mathrm{C}_{\mathrm{A}} \wedge \mathrm{C}_{\mathrm{B}}))\left(\int_{0}^{T} \sup_{s \leq t}\mathcal{W}_2^2(\nu_t^N, \mu_t^{\otimes N}) dt\right) \\ & \leq \mathcal{W}^2_2(\nu_t^N, \mu_t^{\otimes N}) 
    \leq e^{4 \beta_T^2 \mathrm{C}_{\mathrm{B}}}.
\end{split}
\end{equation}
Since $\mathfrak{a} > 0$ is an arbitrary positive constant. Optimization of the final term is achieved by setting $\mathfrak{a} = \exp(\exp(-8\beta_T ( \mathrm{C}_{\mathrm{A}} \wedge \mathrm{C}_{\mathrm{B}}))) \approx 1$ in third inequality and we apply Grönwall's Lemma again. By rewriting the HWI inequality in Eq.~\ref{eq:appendix_HWI_adapted} and setting $t = T$, we have
\begin{equation}\label{eq:appendix_HWi_e_N}
\begin{split}
    \mathcal{H}_T(\mu_{T}) & \leq \underbrace{\mathcal{H}_T(\nu_{T}^{N})}_{\text{Corollary~\ref{corollary:sobolev_upper_bound}}} ~ + ~ \mathrm{e}_N.
\end{split}
\end{equation}
It is noteworthy that the first term on the right-hand side can be controlled by Corollary~\ref{corollary:sobolev_upper_bound}. The error term $\mathrm{e}_N$  called \textit{cardinality errors}, is determined by aggregating Eq.~\ref{eq:appendix_W2_sup}, Eq.~\ref{eq:appendix_fisher_inform_bound}, being inversely proportional to cardinality $N$.
\begin{equation}
\begin{split}
    \mathrm{e}_N & = \frac{1}{N}e^{2\beta_T^2 \mathrm{C}_{\mathrm{B}}}\left(
    \frac{\kappa_{\zeta}}{2}e^{2\beta_T^2 \mathrm{C}_{\mathrm{B}}} + \sqrt{\mathrm{D} + 4\kappa_{\zeta}^2 \left[(1 + \mathrm{D})[N \mathcal{M}^2(2, \zeta_0) + N d \beta_T^2 + \mathrm{D}]e^{\mathrm{D}} + N \mathcal{M}^2(2, \zeta_0)\mathbf{m}_{\zeta}^2(T)\right]}\right)
    \\ & \approx \frac{\kappa_{\zeta}}{2N}e^{4\beta_T^2 \mathrm{C}_{\mathrm{B}}} + 4\kappa_{\zeta}\sqrt{\left( (1+D)\mathcal{M}^2 + d\beta_T^2 + \mathcal{M}^2 \mathbf{m}_{\zeta}^2(T) \right)} \frac{1}{N^{1/2}} \\ & \qquad\qquad\qquad\qquad\qquad\qquad\qquad\qquad + \frac{D(4\kappa_{\zeta}^2 + e^D)}{4\kappa_{\zeta}\sqrt{\left( (1+D)\mathcal{M}^2 + d\beta_T^2 + \mathcal{M}^2 \mathbf{m}_{\zeta}^2(T) \right)}}\frac{1}{N^{3/2}} 
    \\ & \approx \kappa_{\zeta}\mathcal{O}\left(\frac{\mathrm{C}_2}{N}  + \frac{\mathrm{C}_3}{N^{1/2}} + \frac{\mathrm{C}_4}{N^{3/2}}\right) \xrightarrow{N \rightarrow 0} 0, \quad \kappa_{\zeta}(T) \coloneqq \sigma^{-2}(T),
\end{split}
\end{equation}
where $\mathrm{C}_2 \coloneqq \mathrm{C}_2[\beta_T, \mathrm{C}_B]$, $\mathrm{C}_3 \coloneqq \mathrm{C}_3[\mathrm{D}, \mathcal{M}, \beta_T, \mathbf{m}_{\zeta}(T)]$ and $\mathrm{C}_4 \coloneqq \mathrm{C}_4[\mathrm{D}, \mathcal{M}, \beta_T, \mathbf{m}_{\zeta}(T)]$. The proof is complete by rewriting Eq.~\ref{eq:appendix_HWi_e_N} for $\mathrm{e}_N$ computed above.
\end{proof}

This section explicates the division of chaotic entropy into $K$ smaller sub-problems, each with a notably low cardinality ${N_k}_{k \leq K}$. The foundation of the proof relies on the strategic use of the HWI inequality.

\colorbox{rgb:red!10,0.5;green!10,0.5;blue!10,0.5}
{\parbox{0.98\linewidth}{
\begin{manualproposition}{4.1}{\textit{(Subdivision of Chaotic Entropy)}} Let $\mathbb{N} = \{ N_k \}$ be a set of strictly increasing cardinality, and $\mathbb{T} = \{ t_k \}$ be a partition of the interval $[0, T]$, where $k \in \{0, \ldots, K\}$. Under the conditions $\mathbf{s}_{\theta} \in \mathbf{Ker}(\mathcal{G})$, the chaotic entropy can be split into $K$ sub-problems. 
\begin{equation}\label{eq:appendix_subdivision}
\mathcal{H}_T(\mu_{T}) \propto \lim_{K \rightarrow \infty} \sum_{k=0}^K \bigg[ \mathcal{O}\left(\frac{\mathrm{C}_2}{N_{k+1}} + \frac{\mathrm{C}_3}{N_{k+1}^{1/2}} +\frac{\mathrm{C}_4}{N_{k+1}^{3/2}}\right) + \bigg( \frac{\mathrm{C}_5}{{\boldsymbol{\mathfrak{b}}}\sqrt{N_{k+1}}}\bigg)^{k}\mathcal{J}_{MF}(N_k, \theta, [t_k, t_{k+1}]) \bigg].
\end{equation}
The damping ratio $\mathfrak{b} \in \mathbb{N}^{+}, N_{k+1} = \mathfrak{b}N_k$ controls the influence of each sub-problem.
\end{manualproposition}
}}

\begin{proof}

Let us specify the cardinality set as $\mathbb{N} = \{N_k ; N_{k+1} = \boldsymbol{\mathfrak{b}} N_k, k \in \{1, \cdots K\}, \boldsymbol{\mathfrak{b}} \in \mathbb{N}^{+} \}$, where we set $\sup \mathbb{N} = N$.

\begin{equation}\label{eq:appendix_HWi_decomposition}
    \mathcal{H}(\nu_t^{N_{k+1}} | \mathfrak{m} d\mathbf{x}^{N_{k+1}}) - \boldsymbol{\mathfrak{b}} \mathcal{H}(\nu_t^{N_k} | \mathfrak{m} d\mathbf{x}^N_{k})
    \leq \mathrm{e}_{N_{k+1}}
\end{equation}
Note that the $N$-particle relative entropy for measure product can be decomposed $\mathbf{b}$ copy of the original measure.
\begin{equation}
\begin{split}
    \mathcal{H}(\nu_t^{N_{k+1}}) & = \mathcal{H}([\nu_t^{{N_k}}]^{\otimes \boldsymbol{\mathfrak{b}}}) 
    \\ & = \int \log [\varrho_t^{N_k}]^{\otimes \boldsymbol{\mathfrak{b}}}(\mathbf{x}^{N_{k+1}})d[\nu_t^{{N_k}}]^{\otimes \boldsymbol{\mathfrak{b}}}(\mathbf{x}^{N_{k+1}})  - \int \log \zeta_{T-t}^{\otimes {N_{k+1}}}(\mathbf{x}^{N_{k+1}}) d[\nu_t^{{N_k}}]^{\otimes \boldsymbol{\mathfrak{b}}}(\mathbf{x}^{N_{k+1}}) 
    \\ & = \mathbf{b}\mathcal{H}([\nu_t^{N_k}])    
\end{split}
\end{equation}
The equality can be easily seen by showing that
\begin{equation}
\begin{split}
    \int \log [\varrho_t^{N_k}]^{\otimes \boldsymbol{\mathfrak{b}}}(\mathbf{x}^{N_{k+1}}) & = \int_{\mathcal{X}^{N_{k+1}}} \left(\sum_{i = 1}^{\boldsymbol{\mathfrak{b}}} \log \varrho_t^{N_{k+1}}(\pi_{N_k}^i \mathbf{x}^{N_{k+1}}) \right)d[\nu_t^{N_k}]^{\otimes \boldsymbol{\mathfrak{b}}}(\mathbf{x}^{N_{k+1}}) 
    \\ & = \boldsymbol{\mathfrak{b}} \int_{\mathcal{X}^{N_k}}  \log \varrho_t^{N_k} d\nu_t^{N_k}(\mathbf{x}^{N_k}),    
\end{split}
\end{equation}
and the log-probability with the projected component can be calculated as
\begin{equation}
    \int_{\mathcal{X}^{N_{k+1}}} \left(\sum_{i = 1}^{\boldsymbol{\mathfrak{b}}} \log \zeta_{T-t}^{\otimes N_k}(\pi_{N_k}^i \mathbf{x}^{N_{k+1}}) \right)d[\nu_t^{N_k}]^{\otimes \boldsymbol{\mathfrak{b}}}(\mathbf{x}^{N_{k+1}}) = \boldsymbol{\mathfrak{b}} \int_{\mathcal{X}^{N_k}}  \log \zeta_{T-t}^{\otimes {N_k}}(\mathbf{x}^{N_k}) d\nu_t^{N_k}(\mathbf{x}^{N_k}).
\end{equation}
The above calculations are correct for any subsequent elements $N_k < {N_{k+1}} \in \mathcal{T}$ and $\pi_{N_k}^i \mathbf{x}^{N_{k+1}} = (\mathbf{x}_{i \mathfrak{b}}, \cdots, \mathbf{x}_{(i+1)\mathfrak{b}}) \in \mathcal{X}^{N_k}$. By rewriting Eq.~\ref{eq:appendix_HWi_decomposition}, we have
\begin{equation}
    \mathcal{H}(N_{k}, t, \nu_{t}^{N_{k}} | \mathfrak{m} d\mathbf{x}^{N_{k}}) \geq \frac{1}{\boldsymbol{\mathfrak{b}}}\mathcal{H}(N_{k+1}, t, \nu_{t}^{N_{k+1}} | \mathfrak{m} d\mathbf{x}^{N_{k+1}}) - \mathrm{e}_{N_{k+1}}
\end{equation}

Let $t_k \leq t_{k+1}$ be subsequent elements in the partition $\mathcal{T}$. Combining Eq.~\ref{eq:appendix_HWi_decomposition} and Eq.~\ref{eq:appendix_Corollary_3}, we can show the following:
\begin{equation}
\begin{aligned}
    \mathcal{H}(N_0, t_0, \nu_{t_0}^{N_0} | \zeta_0^{\otimes N_0}) & \geq \frac{1}{{\boldsymbol{\mathfrak{b}}}}\mathcal{H}(N_1, t_0, \nu_{t_0}^{N_1} | \zeta_0^{\otimes N_1}) - \mathrm{e}_{N_{k+1}} &&& \text{Eq.~\ref{eq:appendix_HWi_decomposition}}
    \\ & \geq \frac{1}{{\boldsymbol{\mathfrak{b}}}}\mathcal{H}(N_1, t_1, \nu_{t_1}^{N_1} | \zeta_0^{\otimes N_1}) - \frac{\mathcal{M}^2(2, \zeta_0)}{\sqrt{d N_{1}}} \frac{1}{{\boldsymbol{\mathfrak{b}}}}\int_{t_0}^{t_1} \norm{\mathcal{G}_t}_{W^{1, 2}}^{2, \nu_{t_{1}}^{N_{1}}} dt - \mathrm{e}_{N_{1}}. &&& \text{Eq.~\ref{eq:appendix_Corollary_3}}
\end{aligned}
\end{equation}
Note that the Sobolev norm is taken to the law of temporal marginals for Cauchy sequence $(\mathbf{X}_{(\cdot)}^{k, N})(N)$ at timestamp $t = t_{k+1}$ with cardinality condition $N = N_{k+1}$, $\ie \nu_{t_{k+1}}^{N_{k+1}}$. Given the fact $t_{N_K} = T$, one can show the recursion until reaching the target cardinality $N_k \rightarrow N_K$.
\begin{equation}
\begin{aligned}\label{eq:appendix_before_rearranging}
   \underbrace{\mathcal{H}(N_0, 0, \nu_{0}^{N_0} | \zeta_0^{\otimes N_0})}_{= 0} & \geq \mathcal{H}(N_K, T, \nu_T^{N_K} | \zeta_{T}) - \frac{\mathcal{M}^2(2, \zeta_0)}{\sqrt{d}} \sum_{k=0}^K \left( \frac{1}{{\boldsymbol{\mathfrak{b}}}\sqrt{N_{k+1}}}\right)^{k}\int^{t_{k+1}}_{t_k}  \norm{\mathcal{G}_t}_{W^{1, 2}}^{2, \nu_{t_{k+1}}^{N_{k+1}}}dt
\end{aligned}    
\end{equation}
We show that the left-hand side is equal to $0$ by the assumption that the initial states are distributed by standard Gaussian $\mathcal{N}$,
\begin{equation}
    \mathcal{H}(N, 0, \nu_{0}^{N} | \zeta_0^{\otimes N}) = \mathcal{H}( \mathcal{N}^{\otimes N}[\mathbf{I}_{d}] ~|~ \mathcal{N}^{\otimes N}[\mathbf{I}_{d}]) = 0, \quad \forall~N \in \mathbb{N}
\end{equation}
Combining this result and rearranging the terms on both sides of Eq.~\ref{eq:appendix_before_rearranging} yields the inequality
\begin{equation}\label{eq:appendix_kac_chaos_chasotic_entropy}
\begin{split}
    \mathcal{H}(\nu_T^{N_K} | \zeta_0^{\otimes N_K}) & = \mathcal{H}(N_K, T, \nu_T^{N_K} | \zeta_{T})
    \\ & \leq \frac{\mathcal{M}^2(2, \zeta_0)}{\sqrt{d}} \sum_{k=0}^K \left( \frac{1}{{\boldsymbol{\mathfrak{b}}}\sqrt{N_{k+1}}}\right)^{k} \mathcal{J}_{MF}(N_k, \theta, [t_k, t_{k+1}]) + \mathrm{e}_{N_{k+1}}.    
\end{split}
\end{equation}
Recall the following fact that the chaotic entropy converges as PoC is guaranteed by. 
\begin{equation}
\mathcal{H}(\mu_T | \zeta_0) = \lim_{N \rightarrow \infty} \mathcal{H}(\nu_T^{N} | \zeta_0^{\otimes N}).    
\end{equation}

To summarize, we have the desired result and complete the proof.
\begin{equation}
\begin{split}
    \mathcal{H}_T(\mu_{T}) \propto \frac{\mathcal{M}^2(2, \zeta_0)}{\sqrt{d}} \lim_{K \rightarrow \infty}\sum_{k=0}^K \left( \frac{1}{{\boldsymbol{\mathfrak{b}}}\sqrt{N_{k+1}}}\right)^{k} \mathcal{J}_{MF}(N_k, \theta, [t_k, t_{k+1}]) + \kappa_{\zeta}\mathcal{O}\left(\frac{\mathrm{C}_2}{N_{k+1}}  + \frac{\mathrm{C}_3}{N^{1/2}_{k+1}} + \frac{\mathrm{C}_4}{N^{3/2}_{k+1}}\right).
\end{split}
\end{equation}
\end{proof}

\subsection{Comparison of Variational Equations}\label{sec:comparison_VEs}

With the definition of the non-normalized relative entropy $\tilde{\mathcal{H}}$, we derive the variational equation to the temporal derivative. Let $\varrho_t, \zeta_t$ be density representations of forward and reverse diffusion dynamics of FR-SDEs. Taking a temporal derivative $(\ie \partial_t)$ gives the following equality:
\begin{equation}
     \tilde{\mathcal{H}}(\rho_0 | \zeta_T) = -  \int_0^T \partial_t \tilde{\mathcal{H}}(\rho_t | \zeta_{T - t})dt + \tilde{\mathcal{H}}(\rho_T | \zeta_0).
\end{equation}
By rearranging both terms above and using the divergence theorem, one can obtain the closed-form of relative entropy as
\begin{equation}\label{eq:appendix_compare_1}
     \tilde{\mathcal{H}}(\nu_0| \zeta_T) = - \frac{\sigma^2}{2} \int_0^T \mathbb{E}_{\mathbf{Y}_t \sim \rho_t d\mathbf{x}}\left[ \norm{\nabla \log \rho_t - \nabla \log \zeta_{t}}^2 \right]dt, \quad \textbf{VP SDE}\text{,~\cite{song2021maximum}}.
\end{equation}
On the other hand, the proposed Wasserstein variational equation gives the inequality as
\begin{equation}\label{eq:appendix_compare_2}
    \mathcal{H}(\nu_0^N | \zeta_T^{\otimes N}) \precsim \frac{\mathcal{M}}{\sqrt{Nd}}\int_0^T \norm{\nabla \log \varrho_{t}^N - \nabla \log \zeta_{T - t}^{\otimes N}}_W dt, \quad \textbf{MF-CDMs}.
\end{equation}

Given the definition above, we detail three notable differences here:

\colorbox{rgb:red!10,0.5;green!10,0.5;blue!10,0.5}
{\parbox{0.98\linewidth}{
\begin{itemize}
    \item \textbf{Impact of Cardinality $N$.} In contrast to conventional score-matching objectives which are incapable of revealing the impact of data cardinality, our score-matching formula in Eq.~\ref{eq:appendix_compare_2} derived from Wasserstein variational equation explicitly shows the detailed association of particle counts.

    \item \textbf{Cardinality Adaptive Discrepancy.} As can be seen, existing approaches in Eq.~\ref{eq:appendix_compare_1} based on temporal derivative overlook the influence of data dimensionality in the estimation of discrepancy. In contrast, the proposed new variational equation based on the Itô-Wentzell-Lions (known as It\'o's flow of measures) formula in Eq.~\ref{eq:appendix_compare_2}, effectively cancels the dimensionality effect. Moreover, the proposed parameterization of the score function endowed with a reducible structure outlined in the preceding section provides clarity on the architecture's scalability for an increasing $N$, contrasting with the heuristic model choices prevalent in existing architecture modeling.

    \item \textbf{Higher-order Information.} As a result of the geometric deviation induced by Itô-Wentzell-Lions formula, our methodology adopts the Sobolev norm on $W^{1,2}$. It additionally compares the second derivatives of score functions, applying more stringent constraints to achieve a higher level of accuracy in estimating the discrepancy. Meanwhile, the computational overhead remains minimal, as the Hessian of the log-probability exhibits utmost constant complexity. $\ie \nabla^2 \zeta_{T - t}^{\otimes N} \propto \sigma^{-2}_{\zeta} \approx \mathcal{O}(1)$. This simplicity in computation ensures efficiency in practical applications.  
\end{itemize}
}}

\subsection{Particle Branching and Monge-Ampère equation}\label{sec:appendix_particle_branching}

The following result shows that the Monge-Ampère equation sheds light on the precise way in which the optimal particle branching modifies the score function especially when the score networks solve the proposed MF-SM objective optimally.

\colorbox{rgb:red!10,0.5;green!10,0.5;blue!10,0.5}
{\parbox{0.98\linewidth}{
\begin{proposition}\label{prop:optimal_particle_branching} For the optimal parameter profiles $\theta = \theta^{*}$ solving the proposed MF-SM objecitve, then we have
\begin{equation}
\log \varrho_t^{\otimes N_k}(\mathbf{x}^{N_k}) = 
\begin{cases}
     \nabla \log \varrho_t^{N_k, \mathfrak{b}N_k}(\Phi^{\theta^{*}})(\mathbf{x}^{N_k}) + \nabla\mathbf{logdet}(\mathcal{J} \Phi^{\theta *})(\mathbf{x}^{N_k}), \\
     \nabla \log \varrho_t^{\mathfrak{c}N_k, \mathfrak{b}N_k}(\mathbf{x}^{N_k}), ~\forall 1 < \mathfrak{c} \leq \mathfrak{b} \in \mathbb{N}^{+}.
\end{cases}
\end{equation}
For the affine transforms $\Phi^{\theta}(\mathbf{x}) = \mathrm{F}_{\theta}\mathbf{x} + \mathfrak{e}_{\theta}$ with any neural parameters $\mathrm{F}_{\theta} \in \mathbf{GL}(d)$ and $\mathfrak{e}_{\theta} \in \mathbb{R}^d$, the gradient of log-determinant vanishes $(\ie \nabla \mathbf{logdet}(\mathcal{J}\Phi) = 0)$ almost every where $[\nu_t^{N_k}]$.
\end{proposition}
}}
\begin{proof}
Assume the scenario that the branching ratio is $\mathfrak{b} = 2$, where the number of particle is doubled after branching. Considering the necessity for the push-forward mapping to be optimal, in the case of optimal parameter $\theta^{*}$, which solves the problem $\mathbf{(P2})$, one has a representation as follows for arbitrary $M, N$ satisfying $N_k < \mathfrak{b}N_k \leq N$.
\begin{equation}
    (\mathbf{Id}^{\mathfrak{b} - 1} \otimes \Psi^{\theta*})_{\#} \nu_t^{N_{k}} = \nu_t^{\otimes \mathfrak{b}N_k}.
\end{equation}

Following by Brenier's theorem on optimal transport mapping, there exists a convex $\phi$ such that $\nabla \phi$ optimally transports $\nu_t^{M}$ to $\zeta_t^{\otimes M}$, $\ie (\nabla \phi)_{\#}\nu_t^{N_k} = \nu_t^{N_k, \mathfrak{b}N_k}$. On the other hand, the optimal particle branching function needs to assure the following equality:
\begin{equation}
    \Phi^{\theta^{*}}_{\#} \nu_t^{N_k} = \nu_t^{N_k, \mathfrak{b}N_k}, \quad (\Phi^{\theta^{*}})^{-1}_{\#} \nu_t^{N_k, \mathfrak{b}N_k} = \nu_t^{N_k}.
\end{equation}
Whenever we can specify $\Phi_{\#}^{\theta*} = \nabla \phi$ almost everywhere, we have the second-order partial differential equation, so-called \textit{Monge-Ampère equation} as following:
\begin{equation}
    \varrho_t^{\otimes N_k} = \varrho_t^{N_k, \mathfrak{b}N_k}(\Phi^{\theta^{*}}) \mathbf{det}(\mathcal{J} \Phi^{\theta^{*}}), \quad \nabla \log \varrho_t^{\otimes N_k}(\mathbf{x}^{N_k}) = \nabla \log \varrho_t^{N_k, \mathfrak{b}N_k}(\Phi^{\theta^{*}}) + \nabla\mathbf{logdet}(\mathcal{J} \Phi^{\theta *})
\end{equation}
The result is a restatement of the above equality. For the affine transformation $\Phi^{\theta}(\mathbf{x}) = \mathrm{F}_{\theta}\mathbf{x} + \mathfrak{e}_{\theta}$ with neural parameters $\mathrm{F}_{\theta} \in \mathbf{GL}(d)$ and $\mathfrak{e}_{\theta} \in \mathbb{R}^d$, it is trivial that $\mathbf{log} \mathbf{det} > 0$ is a positive constant and the result follows.
\end{proof}

\subsection{Chaotic Convergence of dWGFs}\label{sec:appendix_convergence_dwgf}

This section provides comprehensive proofs for two concentration results presented in Sec~\ref{sec:sampling}.

\colorbox{rgb:red!10,0.5;green!10,0.5;blue!10,0.5}
{\parbox{0.98\linewidth}{

\begin{manualtheorem}{4.2} (Concentration of Chaotic Dynamics) For the constant $\mathfrak{f} \propto \kappa$ dependent on the Log-Sobolev constant $\kappa$ and $\mathfrak{h} \propto R$ dependent on radius of convolution, we have
\begin{equation}\label{eq:appendix_prob_concentration}
    \mathbb{P}\left[ \mathcal{H}(\nu_t^{M, N} | \mu_t^{\otimes M})  \geq \varepsilon \right] \precsim \mathcal{O}(\varepsilon^{-\epsilon^{-d}}) \cdot 
     \mathcal{O}\left(\exp[-M \mathfrak{f}(\kappa)\varepsilon^2 -M \mathfrak{f}(\kappa)\mathfrak{h}(R)]\right).
\end{equation}
    
\end{manualtheorem}
}}
\begin{remark}
Since the proof is a direct modification of results in~\cite{bolley2007quantitative, bolley_2010}, for the sake of simplicity, we only provide the modified descriptions, where the details can be found in the literature. 
\end{remark}

\begin{proof} 

We first assume $N \geq M$ and define the deviation between two vector-fields for $N$- and $M$-particle systems: $\mathcal{X}^M \ni \delta V_t \coloneqq \mathbf{s}_{\theta}^{M, N}(t, \cdot, \nu_t^N) - \mathbf{s}_{\theta}(t, \cdot, \nu_t^M) $, where $\mathbf{s}_{\theta}^{M, N}$ denotes the first $M$-components of $\mathbf{s}_{\theta}$ among $N$ components. By Girsanov theorem~\cite{oksendal2003stochastic} and induced exchangeability due to the fact that $\mathbf{s}_{\theta}$ is reducible, the Radon-Nikodym derivative can be represented as
\begin{equation}
    \frac{d\nu_t^{M}}{d\nu_t^{M, N}} = \exp \left( \frac{1}{\sigma_t} \sum_i^M \int_{[0, t]} \delta V^i_s \cdot dW_s - \frac{1}{2 \sigma_t^2}\norm{\delta V^i_s}^2_E ds\right), \quad 1 \leq i \leq M,
\end{equation}
where $\delta V^i_t$ is the $i$-th component of $\delta V_t$, $W_{(\cdot)}$ is $\nu_{[0, T]}^{M, N}$-adapted Brownian motion, and thus $d\nu_t^N/d\nu_t^M$ is $\nu_t^{M, N}$-martingale. 

Assuming $(2 \sigma_t)^{-1} \leq \mathrm{D}_{\sigma}$ for numerical constant $\mathrm{D}_{\sigma}$, the definition of normalized entropy gives following:
\begin{equation}\label{eq:appendix_relative_entropy_girsanov}
\begin{split}
    \mathcal{H}(\nu_t^{M} |  \nu_t^{M, N}) &= \frac{1}{M}\mathbb{E}_{\nu_t^{M, N}}\left[ \frac{d\nu_t^{M}}{d\nu_t^{M, N}} \right] = \frac{1}{2M}\int_{[0, T]} \mathbb{E}_{\nu_t^{M}}\left[\frac{1}{\sigma_s}\sum_i^M \norm{\delta V^i_s}_E^2\right] ds 
    \\ & = \frac{1}{2}  \int_{[0, T]} \mathbb{E}_{\nu_t^{M}}\left[\frac{1}{\sigma_t}\norm{\delta V_t}_E^2\right] ds \leq \mathrm{D}_{\sigma} \int_{[0, T]} \mathbb{E}_{\nu_t^{M}}\left[\norm{\delta V_t}_E^2\right] ds \leq \mathrm{D}_{\sigma} \sup_{t \in [0,T]} \mathbb{E}_{\nu_t^{M}}\left[\norm{\delta V_t}_E^2\right],    
\end{split}
\end{equation}
where the last equality is induced by the exchangeability of the system. Let us define two empirical projections  $\hat{\nu}_t^{M, N}$ as $\hat{\nu}_t^{M}$ follows:
\begin{equation}
    \hat{\nu}_t^{M, N} \coloneqq \frac{1}{M}\sum^{M}_m \delta_{\mathbf{X}_t^{m, N}},  \quad \hat{\nu}_t^{M} \coloneqq \frac{1}{M}\sum^{M}_m\delta_{\mathbf{X}_t^{m}}.
\end{equation}
For the $d$-dimensional Euclidean ball $\mathbb{B}_{R}^{\mathbf{x}} = \mathbb{B}(\mathbf{x}, R)$ of radius $R$ centered at $\mathbf{x}$, we consider the truncated measures as follows:
\begin{equation}
    \mathbf{X}_t^{j, R} \sim \nu_t^{j, N, R}(d\mathbf{x}) \coloneqq \frac{\chi_{\mathbb{B}_R^{\mathbf{X}^{j, M}}}\nu_t^{j, N}(d\mathbf{x})}{\nu_t^{j, N}[\mathbb{B}_R^{\mathbf{X}^{j, M}}]}, \quad \mathbf{Y}_t^{i, R} \sim  \nu_t^{i, M, R}(d\mathbf{y}) \coloneqq \frac{\chi_{\mathbb{B}_R^{\mathbf{X}^{j, M}}}\nu_t^{i, M}(d\mathbf{y})}{\nu_t^{i, M}[\mathbb{B}_R^{\mathbf{X}^{j, M}}]}, \quad i \neq j \leq M \leq N.
\end{equation}
and we define an empirical measure for truncated representations above:
\begin{equation}
    \hat{\nu}_t^{M, N, R} \coloneqq \frac{1}{M}\sum_{j}^{M}\delta_{\mathbf{X}_t^{j, R}},  \quad \hat{\nu}_t^{M, R} \coloneqq \frac{1}{M}\sum_{i}^{M}\delta_{\mathbf{Y}_t^{i, R}}.
\end{equation}

Next, our objective is to demonstrate the probability inequality concerning the Euclidean norm of the deviation $\delta V_t$ for any given $t \in [0, T]$:
\begin{equation}\label{eq:appendix_randomness}
\begin{split}
    \mathbb{P}\left[ \mathrm{D}_{\sigma}\mathbb{E}_{\nu_t^M}\norm{\delta V_t}_E^2 \geq \varepsilon \right] & \leq \mathbb{P}\left[ \mathrm{D}_{\sigma}\mathbb{E}_{\nu_t^M}\left[ \norm{[\mathrm{B}_{\theta} *_{\mathbb{B}} \nu_t^{M, N}] - [\mathrm{B}_{\theta} *_{\mathbb{B}} \nu_t^{M}]}_E^2\right] \geq \varepsilon \right]
    \\ & = \mathbb{P}\left[ M^{-1}\mathrm{D}_{\sigma} \sum^M_{l} \mathbb{E}_{\nu_t^{l, M}}\left[ \norm{[\mathrm{B}_{\theta} *_{\mathbb{B}} \hat{\nu}_t^{M, N}](\mathbf{X}_t^{l, M}) - [\mathrm{B}_{\theta} *_{\mathbb{B}} \hat{\nu}_t^{M}](\mathbf{X}_t^{l, M})}_E^2\right] \geq \varepsilon \right]
    \\ & \leq \mathbb{P}\left[ \mathrm{C}_{\mathrm{B}}^{\sigma} \sup_{l}\mathcal{W}_2^2(\hat{\nu}_t^{M, N, R}, \hat{\nu}_t^{M, R}) |_{l} \geq \varepsilon \right] = \mathbb{P}\left[ \mathrm{C}_{\mathrm{B}}^{\sigma} \mathcal{W}_2^2(\hat{\nu}_t^{M, N, R}, \hat{\nu}_t^{M, R})|_{l = \bar{l}} \geq \varepsilon \right],
\end{split}
\end{equation}
where we define the index $\bar{l}$ that gives the maximal Wasserstein distance and scale the constant $\mathrm{C}_B^{\sigma} = \mathrm{D}_{\sigma}\mathrm{C}_B$. It is worth noting that the term in the last line of Eq~\ref{eq:appendix_randomness} contains randomness since these two representations $\nu_t^{M, N, R}$ and $\nu_t^{M, R}$ are empirical projections defined in the space $\mathcal{P}(\mathcal{P}_2(\mathbb{R}^d))$. Setting $\varepsilon'' = \varepsilon (\mathrm{C}_B^{\sigma})^{-1}$, there exist constants $\alpha_0, \alpha_1, \alpha_2 > 0$ such that the following can be obtained by triangle inequality of $2$-Wasserstein distance and the Lipschitzness assumption on $(\mathbf{H2})$.
\begin{equation}\label{eq:appendix_prob_ineq_1}
\begin{split}
    & \mathbb{P}\left[ \mathrm{C}_B^{\sigma} \mathcal{W}_2^2(\hat{\nu}_t^{ M, N, R}, \hat{\nu}_t^{M, R})|_{\bar{l}}
    \geq \varepsilon \right] 
    \\ & \leq  \mathbb{P}\left[\mathcal{W}_2^2(\hat{\nu}_t^{M, N, R}, \nu_t^{j, N, R}) + \mathcal{W}_2^2(\nu_t^{j, N, R}, \nu_t^{j, N}) + \mathcal{W}_2^2(\nu_t^{j, N}, \nu_t^{i, M}) + \mathcal{W}_2^2(\nu_t^{i, M}, \nu_t^{i, M, R}) + \mathcal{W}_2^2(\nu_t^{i, M, R}, \hat{\nu}_t^{M, R}) \geq \varepsilon''\right]
    \\ & \leq \mathbb{P}\bigg[\mathcal{W}_2^2(\hat{\nu}_t^{M, N, R}, \nu_t^{j, N, R}) + \mathcal{W}_2^2(\nu_t^{i, M, R}, \hat{\nu}_t^{M, R}) \geq \varepsilon'' - 4 (|\mathcal{E}^\mathbf{x} - \mathcal{E}^\mathbf{y}|)R^2 \exp(-\alpha_0 R^2) - 4\alpha_1 \exp (\alpha_2)\bigg]
    \\ & \leq \mathbb{P}\left[\mathcal{W}_2^2(\hat{\nu}_t^{M, N, R}, \nu_t^{j, N, R}) \geq \varepsilon'' a_0 - 2(|\mathcal{E}^\mathbf{x} - \mathcal{E}^\mathbf{y}|)R^2 \exp(-\alpha_0 R^2) - 4 \alpha_1 \exp(\alpha_2)\right] 
    \\ & \qquad \qquad \qquad + \mathbb{P}\left[\mathcal{W}_2^2(\nu_t^{i, M, R}, \hat{\nu}_t^{M, R}) \geq \varepsilon''(1 - a_0) - 2(|\mathcal{E}^\mathbf{x} - \mathcal{E}^\mathbf{y}|)R^2 \exp(-\alpha_0 R^2)\right],    
\end{split}
\end{equation}
where we define a bounded second moment of empirical measures as follows:
\begin{equation}
    \mathcal{E}^{\mathbf{x}} \coloneqq \mathbb{E}_{\mathbf{x} \sim \nu_t^{M, N}}\exp(a_0 ||\mathbf{x}||_E^2) < \infty, \quad 
    \mathcal{E}^{\mathbf{y}} \coloneqq \mathbb{E}_{\mathbf{y} \sim 
 \nu_t^{M}}\exp(a_0 ||\mathbf{y}||_E^2) < \infty.
\end{equation}
Note that $\mathbf{(H6)}$ assures the boundness of the above terms. Following the analogous calculation in prior proofs with the Lipschitz constraints of $\mathbf{s}_{\theta}$, invoking the Burkholder-Davis-Gundy inequality leads to the following result, where those constants $\alpha_1$ and $\alpha_2$ are dependent on $\mathrm{C}_{\mathrm{A}}$, $\mathrm{C}_{\mathrm{B}}$ and $\beta_t$.
\begin{equation}
    \mathcal{W}_2^2(\nu_t^{j, N}, \nu_t^{i, M}) \leq \sup_t \mathbb{E}\norm{\mathbf{X}_t^{j, N} - \mathbf{X}_t^{i, M}}_E^2 \leq \alpha_1(\beta_t, \mathrm{C}_{\mathrm{A}}, \mathrm{C}_{\mathrm{B}}, \mathrm{D}_{\sigma}) \exp(\alpha_2(\beta_t, \mathrm{C}_{\mathrm{A}}, \mathrm{C}_{\mathrm{B}}, \mathrm{D}_{\sigma}) T).
\end{equation}

Consider the compact subset lying in Polish space $\mathbb{B}_R \subset \mathcal{X}$ and its corresponding probability space $\mathcal{A} \subset \mathcal{P}(\mathbb{B}_R)$. Exercise 6.2.19~\cite{dembo2009large} has shown that the following probability inequality holds;
\begin{equation}
\begin{split}
    \mathbb{P}[\hat{\nu}_t^{M, R} \in \mathcal{A}^1] & \leq \mathrm{M}(\mathcal{A}^1, \delta')\exp\left( -M \inf_{\nu^{\mathcal{A}} \in \mathcal{A}_{\delta'}^1}\tilde{\mathcal{H}}(\nu^{\mathcal{A}} | \nu_t^{i, M, R})\right),
\end{split}
\end{equation}
where $\mathrm{M}(\mathcal{A}^1, \delta')$ stands for the metric entropy, referring to the smallest number of $\delta'$-Wasserstein balls (for the $\mathcal{W}_2$ metric) that are necessary to cover the subset $\mathcal{A}$. Similarly, we have
\begin{equation}
\begin{split}
    \mathbb{P}[\hat{\nu}_t^{M, N, R} \in \mathcal{A}^2] & \leq \mathrm{M}(\mathcal{A}^2, \delta')\exp\left( - M \inf_{\nu^{\mathcal{A}} \in \mathcal{A}_{\delta'}^2}\tilde{\mathcal{H}}(\nu^{\mathcal{A}} | \nu_t^{j, N, R})\right), \quad j \leq M
\end{split}
\end{equation}
For the purpose of deriving the upper bound of Wasserstein distance, we specify the Wasserstein subspace $\mathcal{A}^1$ and $\mathcal{A}^2$ as
\begin{align}
    & \mathcal{A}^1 = \left\{ \nu \in \mathcal{P}(\mathbb{B}_R^{\mathbf{x}_i}) ; \mathcal{W}^2_2(\nu, \nu_t^{i, M, R}) \geq \varepsilon''(1 - a_0) - 2(|\mathcal{E}^\mathbf{x} - \mathcal{E}^\mathbf{y}|)R^2 \exp(-\alpha_0 R^2) \right\} \subset \mathcal{P}(\mathbb{B}_R^{\mathbf{x}_i}),
    \\ & \mathcal{A}^2 = \left\{ \nu \in \mathcal{P}(\mathbb{B}_R^{\mathbf{x}_j}) ; \mathcal{W}^2_2(\nu, \nu_t^{j, N, R}) \geq \varepsilon'' a_0 - 2(|\mathcal{E}^\mathbf{x} - \mathcal{E}^\mathbf{y}|)R^2 \exp(-\alpha_0 R^2) - 4 \alpha_1 \exp(\alpha_2) \right\} \subset \mathcal{P}(\mathbb{B}_R^{\mathbf{x}_j}),
    \\ & \mathcal{A}^1_{\delta'} = \left\{ \nu \in \mathcal{P}(\mathbb{B}_R^{\mathbf{x}_i}) ; \mathcal{W}^2_2(\mathcal{A}^1, \nu) \leq \delta' \right\},  \quad \mathcal{A}^2_{\delta'} = \left\{ \nu \in \mathcal{P}(\mathbb{B}_R^{\mathbf{x}_j}) ; \mathcal{W}^2_2(\mathcal{A}^2, \nu) \leq \delta' \right\},
\end{align}
where $\{\mathcal{A}_{\delta'}^a\}_{a = 1, 2}$ stands for the $\delta'$-thickening of $\mathcal{A}_{\delta'}^a$. We cover the subspace $\mathcal{A}$ with Wasserstein balls of radius $\delta'/2$ in $\mathcal{W}_2$ metric. As the probability measure $\nu_t^{i, N}$ also satisfies Talagrand's inequality with the same constant as $\nu_t^{j, M}$, we take infimum on $\mathcal{A}^{a}_{\delta'}$ to derive
\begin{equation}\label{eq:appendix_entropy_bound_1}
\begin{split}
    \tilde{\mathcal{H}}(\nu | \nu_t^{i, M, R}) & \geq \frac{\kappa(t, \theta)}{2}\mathcal{W}_2^2(\nu, \nu_t^{i, M, R}) - \alpha_3 R^2 \exp(- \alpha_0 R^2)
    \\ & \geq \frac{\kappa(t, \theta)}{2}\left(\left[\varepsilon''(1 - a_0) - 2(|\mathcal{E}^\mathbf{x} - \mathcal{E}^\mathbf{y}|)R^2 \exp(-\alpha_0 R^2)-\delta'\right] \vee 0\right)^2 - \alpha_3 R^2 \exp (-\alpha_0 R^2).
\end{split}
\end{equation}

To get a last line, we first show that there exist constants $c_2, c_3$ depending on $c_0, c_1$ such that the following inequality holds for the arbitrary $c_0, c_1 \in \mathbb{R}$:
\begin{equation}\label{eq:appendix_basic_ineq}
        (c_0 x + c_1 y)^2 \geq 0 ~\longleftrightarrow~ (x - y)^2 \geq (1 - c_2)x^2 - c_3 y^2.  
\end{equation}

Following with above relation with setting $\delta' = \alpha_3 \varepsilon''$
\begin{multline}
    \kappa(t, \theta)(1 - a_4)(1 - a_0 - \alpha_3)^2(\varepsilon'')^2/2 - \kappa(t, \theta)a_5 R^4 \exp(- 2\alpha_0 R^2) \\ \leq \frac{\kappa(t, \theta)}{2}\left(\left[\varepsilon''(1 - a_0) - 2(|\mathcal{E}^\mathbf{x} - \mathcal{E}^\mathbf{y}|)R^2 \exp(-\alpha_0 R^2)-\delta'\right] \vee 0\right)^2,
\end{multline}
Assuming $\ln(1/R^2)/R \leq \alpha_0$, and rescaling numerical terms, we have
\begin{equation}
        \tilde{\mathcal{H}}(\nu | \nu_t^{i, M, R}) \geq  \kappa(t, \theta)a_0(\varepsilon')^2 - \kappa(t, \theta)\alpha_3 R^4 \exp(- 2 \alpha_0 R^2).
\end{equation}
Since $\nu_t^{j, N}$ enjoys an identical constant for Talagrand's inequality compared to $\nu_t^{i, N}$, the lower-bound of $\tilde{\mathcal{H}}(\nu | \nu_t^{j, R})$ for the subset $\mathcal{A}^2$ can be obtained:
\begin{equation}
    \tilde{\mathcal{H}}(\nu | \nu_t^{i, M, R}) \geq \frac{\kappa(t, \theta)}{2}\left(\left[\varepsilon''a_0 - 2(|\mathcal{E}^\mathbf{x} - \mathcal{E}^\mathbf{y}|)R^2 \exp(-\alpha_0 R^2) - 4\alpha_1 \exp(\alpha_2) -\delta'\right] \vee 0\right)^2 - \alpha_3 R^2 \exp (-\alpha_0 R^2).
\end{equation}
As similar to above, we apply the inequality in Eq.~\ref{eq:appendix_basic_ineq} twice to get constants  $a_5, a_6$ such that following relation holds:
\begin{multline}
    \kappa(t, \theta)(1-a_5)(a_0 - \alpha_3)^2(\varepsilon'')^2/2 - \kappa(t, \theta) a_6 \left(R^4 \exp(-2 \alpha_0 R^2) +  \exp 2 \alpha_2 + R^2\exp(-\alpha_0 R^2 + \alpha_2)\right) 
    \\ \leq \frac{\kappa(t, \theta)}{2}\left(\left[\varepsilon''a_0 - 2(|\mathcal{E}^\mathbf{x} - \mathcal{E}^\mathbf{y}|)R^2 \exp(-\alpha_0 R^2) - 4\alpha_1 \exp(\alpha_2) -\delta'\right] \vee 0\right)^2.
\end{multline}
For some $\alpha_3'$ and $\alpha_2'$, we rescale numerical constants in the inequality to have:
\begin{equation}\label{eq:appendix_entropy_final_1}
    \tilde{\mathcal{H}}(\nu | \nu_t^{i, M, R}) \geq \kappa(t, \theta)a_1 \varepsilon^2 - \kappa(t, \theta) \alpha_3' R^4 \exp(-2 \alpha_0 R^2) - \alpha_2'.
\end{equation}

Following by the Theorem A.1.~\cite{bolley_2010}, the metric entropy for the subset $\mathcal{A}^1$ can be bounded for some numerical constants $b_0$
\begin{equation}\label{eq:appendix_metric_entropy_1}
    \mathrm{M}(\mathcal{A}^1, \delta') \leq \mathrm{M}(\mathcal{P}_2(\mathbb{B}_R^{\mathbf{x}_i}), \delta') \leq \left(\frac{b_0 R}{\delta'}\right)^{2 \left( b_0 \frac{2 R}{\delta'}\right)^d} = \left(\frac{b_0 R \mathrm{C}_B }{\alpha_3 \varepsilon}\right)^{2 \left( b_0 \frac{2 R \mathrm{C}_B}{\alpha_3 \varepsilon}\right)^d} \sim \mathcal{O}(\varepsilon^{-\epsilon^{-d}}),
\end{equation}
where we set the radius of Wasserstein ball $\delta' = \alpha_3 \varepsilon''$. By collecting Eq.~\ref{eq:appendix_entropy_final_1} and Eq.~\ref{eq:appendix_metric_entropy_1}, we have
\begin{align}\label{eq:appendix_w_2_1_}
    & \mathbb{P}\left[\mathcal{W}_2^2(\hat{\nu}_t^{M, N, R}, \nu_t^{j, N, R}) \geq \varepsilon'' a_0 - 2(|\mathcal{E}^\mathbf{x} - \mathcal{E}^\mathbf{y}|)R^2 \exp(-\alpha_0 R^2) - 4 \alpha_1 \exp(\alpha_2)\right] \\ & \leq \left(\frac{b_0 R \mathrm{C}_B }{\alpha_3 \varepsilon}\right)^{2 \left( b_0 \frac{2 R \mathrm{C}_B}{\alpha_3 \varepsilon}\right)^d} \exp(-M \kappa(t, \theta)a_1 \epsilon^2 - M \kappa(t, \theta)\alpha_3' R^4 \exp(-2 \alpha_0 R^2) - \alpha_2')
    \\ & \precsim \mathcal{O}(\varepsilon^{-\epsilon^{-d}}) \mathcal{O}(\exp(-M \mathfrak{f}\left[\varepsilon^2\right])).
\end{align}
With a similar calculation as done above, one can obtain
\begin{multline}\label{eq:appendix_w_2_2_}
    \mathbb{P}\left[ \mathcal{W}_2^2(\nu_t^{i, M, R}, \hat{\nu}_t^{M, R}) \geq \varepsilon''(1 - a_0) - 2(|\mathcal{E}^\mathbf{x} - \mathcal{E}^\mathbf{y}|)R^2 \exp(-\alpha_0 R^2)\right] \\ \precsim \mathcal{O}(\varepsilon^{-\epsilon^{-d}}) \mathcal{O}(\exp(-M \mathfrak{f}\left[\varepsilon^2 + R^4 \exp(-R^2)\right])).
\end{multline}

Combining Eq.~\ref{eq:appendix_w_2_1_} and Eq.~\ref{eq:appendix_w_2_2_} with Eq.~\ref{eq:appendix_prob_ineq_1} gives the desired outcome:
\begin{equation}
    \mathbb{P}\left[ \mathrm{C}_B^{\sigma}\mathcal{W}_2^2(\hat{\nu}_t^{ M, N, R}, \hat{\nu}_t^{M, R})|_{\bar{l}} \geq \varepsilon \right] \precsim \mathcal{O}(\varepsilon^{-\epsilon^{-d}}) \cdot \mathcal{O}\left[\exp(-M \mathfrak{f}\varepsilon^2 -M \mathfrak{f}\mathfrak{h}(R))\right].
\end{equation}
Given the fact that the above relation holds for all $t \in [0, T = 1]$, the proof is complete as we take the limitation with $N \rightarrow \infty$.
\end{proof}

\colorbox{rgb:red!10,0.5;green!10,0.5;blue!10,0.5}
{\parbox{0.98\linewidth}{
\begin{manualtheorem}{4.3}{\textit{(Concentration of MF-SM).}} Let $\mathbf{X}_t^{N}$ be a solution to MF-SDE (for dWGFs) for the set of particles. Then, for any $\epsilon \in (0, 1)$, the following is true:
\begin{equation}
    \mathbb{P}\left( \left| \mathbb{E}_t F(\mathbf{X}_t^{N}) - \mathcal{J}_{MF}(N = 1, \theta, \mu_{[0, T]})\right| \geq \varepsilon \right) \leq \exp(-\mathrm{C}_5 \mathfrak{f}(\kappa)^{-2} \left[\varepsilon \sqrt{N} - \mathrm{C}_6 \sqrt{\left(1 + N^{(-q + 4)/2q} \right)}\right]^2),
\end{equation}
where $\mathfrak{f} \coloneqq \mathfrak{f}(\kappa) = \sup_{t \in [0, T]}[c(t, \theta) \vee \kappa(t, \theta)]$, and the log-Sobolev constant of time-inhomogeneous dynamics $c : [0, T] \times \Theta \rightarrow \mathbb{R}$ is defined as
\begin{equation}
    c(t, \theta) = \int_0^t \exp \left( -2 \int_v^t \kappa(u, \theta) du \right) dv, \quad \kappa(t, \theta) = 
\begin{cases}
\displaystyle 
    -\frac{\beta_t}{2} + \frac{\beta_t}{\sigma_{\zeta}^2(t)} & \text{for } \theta = \theta^{*}, \\
\displaystyle 
    -\frac{\beta_t}{2} + \gamma_{\mathrm{A}} + \gamma_{\mathrm{B}}  & \text{for } \theta \neq \theta^{*}.
\end{cases}
\end{equation}
The neural parameter $\theta^{*}$ of score networks ensures vanishing $N$-particle relative entropy $\mathcal{H}_T^{\nu_t^N} |_{\theta = \theta^{*}} = 0$ for all $t \in [0, T]$. In other means, it follows that $\mathbf{s}_{\theta} = \nabla \zeta_{T - t}^N$ almost surely $[\nu_{[0, T]}^N]$.
\end{manualtheorem}
}}

\begin{remark}
 Note that in the main manuscript, we omitted the curvature effect by replacing $\sqrt{\sup_{t \in [0, T]}c(t, \theta)} \hookrightarrow \mathrm{K}(\kappa)$ to only emphasize the connection towards HWI inequality in the estimation of MF-SM. However, the full description specifies the explicit effect of the Bakry-\'Emery curvature condition, showing that the designing factor of VP-SDE $(\eg \beta_0, \beta_1)$ controls convergent behavior of our $N$-particle system towards mean-field limit $\mu_t$.
\end{remark}

\begin{proof}
We provide an analysis of adapting the VP SDE~\cite{score_based_generative_model} to an $N$-particle mean-field system. Through the adoption of VP SDE, the original drift term $f_t^{\otimes N}$ in our denoising WGF is characterized by substituting with the corresponding drift term in MF-VP SDE, $\ie - \nabla\left[{\beta_t \norm{\mathbf{x}^N}^2_E} / 4 \right] =  f_t^{\otimes N}$.  Hence, the vector fields $\nabla V$ of potential $V$ for $N$-particle system can be represented as follows:
\begin{align}\label{eq:appendix_V_N}
    & \nabla V^N(t, \mathbf{x}^N) = - \nabla\left[\frac{\beta_t \norm{\mathbf{x}^N}^2_E}{4} \right] - \beta_t \log \zeta_{T - t}^{\otimes N}(\mathbf{x}^N), \quad \text{for } \theta = \theta^{*}, \\
    & \nabla V^N(t, \mathbf{x}^N, \nu_t^N) = - \nabla\left[\frac{\beta_t \norm{\mathbf{x}^N}^2_E}{4} \right] - \mathrm{A}(t, \mathbf{x}^N) - [\mathrm{B} *_{\mathbb{B}_R} \nu_t^N](\mathbf{x}^N), \quad \text{for } \theta \neq \theta^{*}.
\end{align}

It is noteworthy that $\theta^{*}$ is the parameter profile that can be obtained from perfect score matching where the proposed score networks optimally approximate the score function, $\ie s_{\theta^{*}} = \nabla \log \zeta_{T-t}$. The constant $\beta_t = \beta_{\text{min}} + t(\beta_{\text{max}} - \beta_{\text{min}})$ is defined as a linear function on $t$ for the pre-defined fixed hyperparameters $(\beta_{\text{min}}, \beta_{\text{max}})$. Note that $\beta_t$ is non-decreasing over $t$ and $\sup_{t \in [0, T]}\beta_t =  \beta_T$.

Recall that
\begin{equation}
    \zeta_t \coloneqq \mathcal{N}(\mathbf{m}_{\zeta}(t)\mathbf{Y} ; \sigma^2_{\zeta}(t)\mathbf{I}_d), \quad \nabla \log \zeta_{t}^{\otimes N}(\mathbf{x}^N) = -\frac{1}{\sigma^2_{\zeta}(t)}(\mathbf{x}^N - \mathbf{m}_{\zeta}(t)\mathbf{Y}^N).
\end{equation}
where $\mathbf{Y}^N \sim \zeta_0^{\otimes N}$ stands for the $N$-copies of target data instance and the scalar mean and variance are given as 
\begin{equation}
\mathbf{m}_{\zeta}(t) = e^{-\frac{1}{2}\int_0^t \beta_s ds}, ~~\sigma_{\zeta}^2(t) = 1 - e^{- \int_0^t \beta_s ds}.
\end{equation}
Taking Hessian operator to $V^1$, we have
\begin{equation}\label{eq:appendix_kappa_def}
\kappa(t, \theta) = \mathbf{J}(\nabla V^1) = \nabla^2 V^1 = 
\begin{cases}
\displaystyle 
    -\frac{\beta_t}{2} + \frac{\beta_t}{\sigma_{\zeta}^2(T - t)} & \text{for } \theta = \theta^{*}, \\
\displaystyle 
    -\frac{\beta_t}{2} + \gamma_{\mathrm{A}} + \gamma_{\mathrm{B}}  & \text{for } \theta \neq \theta^{*}.
\end{cases}
\end{equation}
Following by Eq.~\ref{eq:appendix_carre_estimations}, we compute the carr\^e du champ operators as
\begin{align}
    & \boldsymbol{\Gamma}(t)(f,f) = \beta_t\norm{\nabla f}_E^2, \label{eq:appendix_carre_estimations_}
    \\ & \boldsymbol{\Gamma}_2(t)(f) =
    \begin{cases}
         \norm{\nabla^2 f }_F^2 - (-\beta_t/2 + \gamma_{\mathrm{A}} + \gamma_{\mathrm{B}})\norm{\nabla f}_E^2, \quad \text{for } \theta \neq \theta^{*}\\
        \norm{\nabla^2 f }_F^2 - (\beta_t/2 + \beta_t/{\sigma_{\zeta}^2(T - t)}) \norm{\nabla f}_E^2, \quad \text{for } \theta = \theta^{*},
    \end{cases} \label{eq:appendix_carre_estimations_2}
    \\ & \partial_t \boldsymbol{\Gamma}(t)(f) = \partial_t \beta_t \norm{\nabla f}_E^2 \approx \beta_{\text{max}}\norm{\nabla f}_E^2. \label{eq:appendix_carre_estimations_3}
\end{align}
Recall the Bakry-Émery criterion in Eq.~\ref{eq:appendix_bakry_emery_criterion}:
\begin{equation}
    \boldsymbol{\Gamma}_2(t)(f) + \frac{1}{2}\partial_t \boldsymbol{\Gamma}(t)(f) \geq \kappa(t)\boldsymbol{\Gamma}(t)(f).
\end{equation}
Utilizing the estimations from Eq.~\ref{eq:appendix_carre_estimations_} to Eq.~\ref{eq:appendix_carre_estimations_3} gives
\begin{equation}
    \norm{\nabla^2 f }_F^2 + \frac{1}{2}\beta_{\text{max}}\norm{\nabla^2 f }_F^2 \geq \kappa(t, \theta) \norm{\nabla^2 f }_F^2.
\end{equation}

This concludes that $\kappa(t, \theta) = (\beta_t/2 + \gamma_{\mathrm{A}} + \gamma_{\mathrm{B}})$ if $\theta \neq \theta^{*}$ and $\kappa(t) = \beta_t(1/2 + 1/\sigma^2_{\zeta}(t))$ if $\theta = \theta^{*}$. 

Once we determine the curvature estimation for time $t$ and $\theta$, the next step is to derive concentration inequality from $\Phi$-log Sobolev inequality. Let $\mathbf{P}_t^{N, *}$ be the dual semi-group of $\mathbf{P}_t^N$ for the $N$-particle denoising MF-SDEs, which can be represented as
\begin{equation}
    \mathbf{X}_t^N \sim \nu_t^N = \mathbf{P}_t^{N, *} d\zeta_{T-t}^{\otimes N}.
\end{equation}
For the action of dual semigroup onto $\zeta_{T-t}^{\otimes N}$, $\Phi$-log Sobolev inequality in Eq.~\ref{eq:appendix_phi_log_sobolev} can be modified as 
\begin{equation}
    \mathbf{Ent}^{\Phi}_{\mathbf{P}_t^{N, *} d\zeta_{T-t}}(g) \leq c(t)P_{t}\left(\Phi''(g) \Gamma(t)(g) \right), \quad c(t) = \int_0^t \exp \left( -2 \int_v^t \kappa(u) du \right) dv.
\end{equation}
Setting $\Phi(g) = g^2$ and $g = f^2 = \exp(\mathfrak{u}F)$ with the function $F_t \coloneqq \norm{\mathcal{G}_t}_E^2 + \norm{\nabla \mathcal{G}_t}_F^2$ $(F_t : \mathcal{X}^N \rightarrow \mathbb{R})$ that haves Lipschitz constant $\mathbf{Lip}(F)$, we obtain

\begin{equation}
    \mathbf{Ent}^{\Phi}_{\nu_t}(g) \leq 2 c(t) \mathbb{E}_{\nu_t^N}[\Gamma(t)(g)] \leq 2 \sup_t [\beta_t c(t)]\mathbb{E}_{\nu_t^N}[\norm{\nabla g}_E^2].
\end{equation}
By definition of $\Gamma$, we have 
\begin{equation}
    \Gamma(t)(g) = \sum_j \sqrt{\beta_t} \sum_{i} \sqrt{\beta_t}\partial_i g \partial_j g = \beta_t \norm{\nabla g}_E^2,
\end{equation}
Replacing $g = f^2$ gives
\begin{equation}
    \mathbf{Ent}^{\Phi}_{\nu_t^N}(f^2) \leq 2 \sup_t [\beta_t c(t)]\mathbb{E}_{\nu_t^N}[\norm{\nabla f^2}_E^2]
\end{equation}
To estimate the right-hand side, we show that
\begin{equation}\label{eq:appendix_gradient_bound}
    \mathbb{E}_{\nu_t^N}[\norm{\nabla f^2}_E^2] = \mathbb{E}_{\nu_t^N} \left[ \frac{u}{4}\norm{\nabla F}_E^2 e^{u F}\right] \leq \frac{u^2}{4}\mathbf{Lip}^2(F)\mathbb{E}_{\nu_t^N}[f^2].
\end{equation}

On the other hand, the $\Phi$-entropy with respect to measure $\nu_t$ can be directly calculated as
\begin{equation}
\begin{split}
    \mathbf{Ent}^{\Phi}_{\nu_t^N}(f^2) & = \mathfrak{u} F \mathbb{E}_{\nu_t^N}[f^2] - \mathbb{E}_{\nu_t^N}[f^2] \log \mathbb{E}_{\nu_t^N}[f^2] \leq \sup_t [\beta_t c(t)]\frac{\mathfrak{u}^2}{2}\mathbf{Lip}^2(F)\mathbb{E}_{\nu_t^N}[f^2],    
\end{split}
\end{equation}
where the right-hand side is induced from Eq.~\ref{eq:appendix_gradient_bound}. Now, we consider log-expectation to extract the expectation of $F$ in the summation.
\begin{equation}\label{eq:appendix_1_over_u}
    \frac{1}{\mathfrak{u}}\log \mathbb{E}_{\nu_t^N}[f^2] = \mathbb{E}_{\nu_t^N}[F] + \int_0^{\mathfrak{u}} \frac{\partial}{\partial \mathfrak{u}}\left(\frac{1}{\mathfrak{u}}\mathbb{E}_{\nu_t^N}[f^2]\right)d\mathfrak{u} \leq \mathbb{E}_{\nu_t^N}[F] + \frac{\mathfrak{u} \sup_t [\beta_t c(t)]\mathbf{Lip}^2(F)}{2}.
\end{equation}
The inequality comes from the fact that
\begin{equation}
    \frac{\partial}{\partial \mathfrak{u}}\left(\frac{1}{\mathfrak{u}}\mathbb{E}_{\nu_t^N}[f^2]\right) \leq \frac{\sup_t [\beta_t c(t)] \mathbf{Lip}^2(F)}{2} \leq \frac{\beta_T \sup_t [c(t)] \mathbf{Lip}^2(F)}{2}.
\end{equation}
We multiply $\mathfrak{u}$ and subsequently take exponential on both sides of Eq.~\ref{eq:appendix_1_over_u}, and the exponential inequality follows.
\begin{equation}
    \mathbb{E}_{\nu_t^N}[\exp(\mathfrak{u} F)] \leq \exp(\mathfrak{u} \mathbb{E}_{\nu_t^N}[F] + \beta_T \sup_t [c(t)] \mathbf{Lip}^2(F)/2).
\end{equation}
As a direct application of Chebyshev’s inequality, we see that
\begin{equation}
\begin{split}
    \mathbb{P}\left( |F(\mathbf{X}_t^{N}) - \mathbb{E}_{\nu_t^N} F(\mathbf{X}_t^{N}) | \geq \varepsilon \right) & \leq 2 \exp(-\mathfrak{u} \varepsilon + \beta_T \sup_t [c(t)] \mathbf{Lip}^2(F)\varepsilon^2/2). 
\end{split}
\end{equation}
By selecting an optimal variable $\mathfrak{u}$, we finally have
\begin{equation}
\begin{split}
    \mathbb{P}\left( |F(\mathbf{X}_t^{N}) - \mathbb{E}_{\nu_t^N}F(\mathbf{X}_t^{N}) | \geq \varepsilon \right) & \leq 2 \exp(\frac{- \varepsilon^2}{2 \beta_T \sup_t c(t) \mathbf{Lip}^2(F)} ). 
\end{split}
\end{equation}
Given that the particles are exchangeable by the result of Proposition~\ref{prop:exchangebility}, one can demonstrate that with a probability of at least $1 - \varepsilon$, we have
\begin{equation}
    |F(\mathbf{X}_t^{N}) - \mathbb{E}_{\nu_t^N}F(\mathbf{X}_t^{N}) | \geq \sqrt{2 \beta_T \sup_t c(t) \mathrm{L}^2 \log(2/\varepsilon)}, 
\end{equation}
for any $1 \leq j \leq N$ and $F \in \mathbf{Lip}(\mathrm{L}, \mathcal{X}^N)$. Let us decompose $F$ into reducible components as $F(\mathbf{X}_t^N) = (1/N)\sum_i^N \bar{F}(\mathbf{X}_t^{i, N})$. Since one can see that $L = (1/\sqrt{N})\mathbf{Lip}(\bar{F})$, exchangeability of particles gives
\begin{equation}
\begin{split}
    \mathbb{P}\left( \left| F(\mathbf{X}_t^{N}) - \frac{1}{N}\mathbb{E}_{\nu_t^{j, N}}\sum_i^N \bar{F}(\mathbf{X}_t^{i, N}) \right| \geq \varepsilon \right) & \leq 2 \exp(\frac{- \varepsilon^2 N }{2 \beta_T \sup_t c(t) \mathbf{Lip}^2(\bar{F})} ), \quad \forall~ j \leq N.
\end{split}
\end{equation}

Note that the reducibility of score networks assures that ${F}(\mathbf{X}_t^{N}) \coloneqq {F}(\mathbf{X}_t^{N}, \nu_t^N) = \norm{\mathcal{G}_t(\mathbf{X}_t^{N}, \nu_t^N)}_E^2 + \norm{\boldsymbol{J}\mathcal{G}_t(\mathbf{X}_t^{N}, \nu_t^N)}_F^2$ and $\bar{F}(\mathbf{X}_t^{i, N}) \coloneqq \bar{F}(\mathbf{X}_t^{i, N}, \hat{\nu}_t^N) = \norm{\mathcal{G}_t(\mathbf{X}_t^{i, N}, \hat{\nu}_t^N)}_E^2 + \norm{\boldsymbol{J}\mathcal{G}_t(\mathbf{X}_t^{i, N}, \hat{\nu}_t^N)}_F^2$ with relation $F(\mathbf{X}_t^N) = (1/N)\sum_i^N \bar{F}(\mathbf{X}_t^{1, N})$. 

Given the definition of canonical projection $\pi_N^i(\mathbf{x}^N) = \mathbf{x}_i$, we define an empirical measure as $\hat{\nu}_t^N(d\mathbf{x}) \coloneqq \frac{1}{N}\sum_i^N \delta_{\pi_N^i \mathbf{X}_t^{N}}$. Then, the triangle inequality naturally gives the following results:
\begin{equation}\label{eq:appendix_bound_wass}
\begin{split}
    \left| \mathbb{E}_{\hat{\nu}_t^N}\bar{F}(\cdot, \hat{\nu}_t^N) - \mathbb{E}_{\mu_t}\bar{F}(\cdot, \mu_t)\right| & \leq \left| \mathbb{E}_{\hat{\nu}_t^N}\bar{F}(\cdot, \hat{\nu}_t^N) - \mathbb{E}_{\mu_t}\bar{F}(\cdot, \hat{\nu}_t^N)\right| + \left| \mathbb{E}_{\mu_t}\bar{F}(\cdot, \hat{\nu}_t^N) - \mathbb{E}_{\mu_t}\bar{F}(\cdot, \mu_t)\right| 
    \\ & \leq \mathbf{Lip}(\bar{F})\mathcal{W}_2(\hat{\nu}_t^N, \mu_t) + 4d(\gamma_{\mathrm{B}}')^2 \mathcal{W}_2^2(\hat{\nu}_t^N, \mu_t)
    \\ & \leq \mathrm{C}''\left( 4d(\gamma_{\mathrm{B}}')^2 + \mathbf{Lip}(\bar{F}) \right)\sqrt{\left(\frac{1}{N^{1/2}} + \frac{1}{N^{(q - 2)/q}} \right)},
\end{split}
\end{equation}
where the second inequality is induced from the fact that
\begin{equation}
\begin{split}
| \langle \bar{F}|_{\hat{\nu}_t^N}, \hat{\nu}_t^N \rangle - \langle \bar{F}|_{\hat{\nu}_t^N}, \mu_t \rangle| & \leq \mathbf{Lip}(\bar{F})\sup_{\bar{F}/\mathbf{Lip}} \left(\frac{1}{\mathbf{Lip}(\bar{F})}\right) |\langle \bar{F}, \hat{\nu}_t^N \rangle - \langle \bar{F}, \mu_t \rangle| 
\\ & \leq \mathbf{Lip}(\bar{F}) \mathcal{W}_1(\hat{\nu}_t^N, \mu_t) \leq \mathbf{Lip}(\bar{F})\mathcal{W}_2^2(\hat{\nu}_t^N, \mu_t),    
\end{split}
\end{equation}
and one can calculate the bounded Jacobian of score networks as
\begin{equation}
    \norm{\boldsymbol{J}_x\left[\mathbf{s}_{\theta}(t, \bar{\mathbf{X}}_t, \mu_t) - \mathbf{s}_{\theta}(t, \bar{\mathbf{X}}_t, \hat{\nu}_t^N) \right]}_F^2 \leq 2\norm{\gamma_{\mathrm{B}}' \mathbf{I}_{d}}_F^2 = 2d(\gamma_{\mathrm{B}}')^2,   \quad \mathbf{\bar{X}}_t \sim \mu_t,
\end{equation}
The asymptotic upper-bound in the last line of Eq.~\ref{eq:appendix_bound_wass} can be derived from the result explored in Theorem 1~\cite{fournier2015rate} associated with numerical constant $\mathrm{C}''$. By combining the results, we finally have
\begin{equation}
\begin{split}
    \mathbb{P}\left( \left| \mathbb{E}_t F(\mathbf{X}_t^{N}) - \mathbb{E}_{t, \mu_t}F(\bar{\mathbf{X}}_t) \right| \geq \varepsilon \right) & \leq 2 \exp(\frac{- \left[\varepsilon \sqrt{N} - \mathrm{C}''\left( 4d(\gamma_{\mathrm{B}}')^2 + \mathrm{L} \right)\sqrt{\left(1 + N^{(-q + 4)/2q} \right)}\right]^2 }{2 \beta_T \sup_t c(t) \mathrm{L}^2} ).
\end{split}
\end{equation}
Since the expectation of $F$ with respect to measure $\mu_t$ can be represented as squared $W^{1,2}$-Sobolev norm, $\ie \mathbb{E}_{\nu_t^N}F = \norm{\mathcal{G}_t}_W^2$. By rephrasing the result above with numerical constants $\mathrm{C}_5 = \mathrm{C}''\left( 4d(\gamma_{\mathrm{B}}')^2 + \mathrm{L} \right)$, $\mathrm{C}_6 = 2\beta_T L^2$ and $\mathfrak{f}(\kappa) \coloneqq \sup_{t}[c(t) \vee \kappa(t, \theta)]$, we bring the proof to completion, revealing the concentration property of our mean-field score matching objective.
\end{proof}

\begin{table}[h]
    \centering
    \small
    \setlength{\tabcolsep}{6pt}
    \renewcommand{\arraystretch}{1.15}
    \caption{\textbf{Hyperparameters according to cardinality in data instances.}}
    \begin{tabular}{c|cccc}
    \toprule
    \toprule
         Hyperparameters & \multicolumn{1}{c}{$N = 10^3$}  &  \multicolumn{1}{c}{$N = 10^4$} & \multicolumn{1}{c}{$N = 2 \times 10^4$} & \multicolumn{1}{c}{$N = 10^5$}\\
    \midrule
        Learning Rate & $1.0e^{-3}$ & \multicolumn{3}{c}{$1.0e^{-4}$} \\
        (VP SDE) & \multicolumn{4}{c}{$\sigma_t^2 = \beta_t, ~\beta_t = \beta_{\text{min}} + t(\beta_{\text{max}} - \beta_{\text{min}}), \quad \beta_{\text{max}} = 20.0, \beta_{\text{min}} = 0.1$ } \\
         (Diffusion Steps) $~\mathbb{K}$  & \multicolumn{4}{c}{$\{1, \cdots, 300\}, \quad |\mathbb{K}| = 300$} \\
        (Branching Ratio) $~\mathfrak{b}$ & \multicolumn{4}{c}{$2$} \\
         (Branching Steps) $~\mathbb{K}'$ & \multicolumn{1}{c}{$\{100, 200\}$} & \multicolumn{2}{c}{$\{50, 100, 150, 200 \}$} & \multicolumn{1}{c}{$\{50,100,150,200,250\}$} \\
         (Initial Cardinality) $~\{N_0\}$  & $250$ & $625$ & $1250$ & $3125$ \\
         (Interaction Degree) $~k$ & $10$ &  $3$ & $3$ & $3$\\
    \bottomrule
    \bottomrule
    \end{tabular}
    \vspace{-2mm}
    \label{tab:appendix_hyper_parameters}
\end{table} 

\subsection{Implementation Details, Training and Sampling of MF-CDMs}\label{sec:appendix_implementation}

\textbf{Hyperparameters.} Across all experiments, our MF-CDMs are configured to perform a total of $300$ diffusion steps ($|\mathbb{K}| = 300$) in the denoising path. This includes particle branching at selected sub-steps within the subset $\mathbb{K}' \subset \mathbb{K}$, adhering to a branching ratio of $\mathfrak{b} = 2$. 
The radius $R$ of the convolution is determined by the average distance between each particle and its proximate $k$ interacting particles, calculated at every iteration during the training process. In the inference time, we utilized the radius calculated latest training iteration. Table~\ref{tab:appendix_hyper_parameters} summarizes detailed specifications of hyperparameters.

\begin{figure}[h]
\centering
\subfloat{%
  \includegraphics[clip,width=0.9\columnwidth]{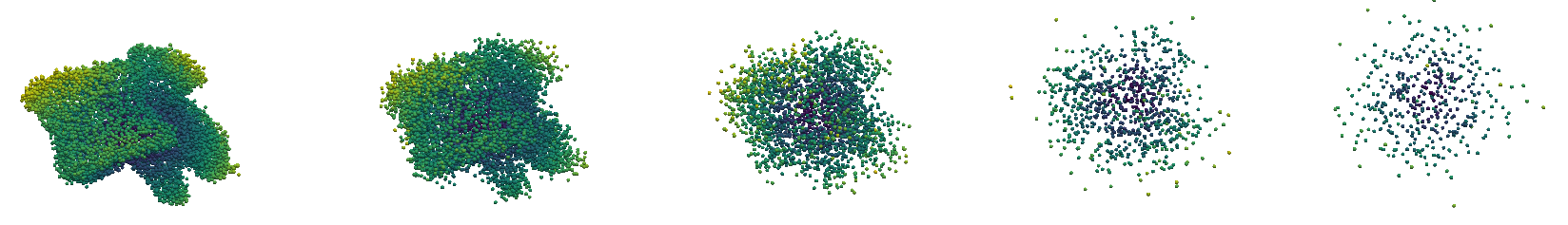}%
}

\subfloat{%
  \includegraphics[clip,width=0.87\columnwidth]{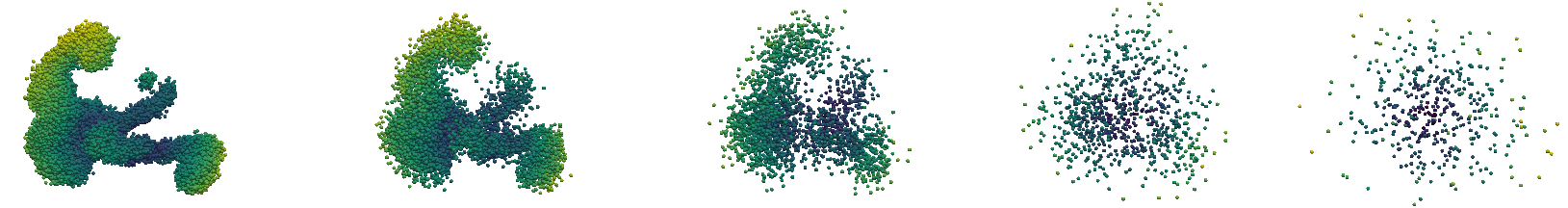}%
}
\caption{\textbf{Additional Qualitative Results on MedShapeNet Dataset}. We display reconstructed 3D shapes \textit{Spine L3 vertebra} and \textit{Colon} in MedShapeNet dataset which comprise $2.0e^{+3}$ points.}
\label{fig:appendix_additional}
\end{figure}

\textbf{Example: Sampling of MF-CDMs on MedShapeNet.} In the experiments targeting a cardinality of $2.0e^{+3}$ on MedShapeNet, we initiate by simulating denoising particle paths starting from a lower cardinality of $N_0 = 1.25 e^{+3}$, proceeding until the first branching steps at $\{50\} \in \mathbb{K}'/\mathbb{K}'$. In the branching step, we apply a point branching function to the simulated particles, which increases to twice the number of particle profiles, $N_{100 \in \mathbb{K}'} = 2.5 e^{+3}$. The following diagram provides an overview of how the branching operation increases cardinality during the denoising process:
\begin{multline}
    \underbrace{N_0 \rightarrow \cdots \rightarrow N_{49 \in \mathbb{K}/\mathbb{K}'}}_{\textbf{Card}:~1.25e^{+3}} \xlongrightarrow[\Phi^{*}]{\text{Branching}} \underbrace{N_{50 \in \mathbb{K}'} \rightarrow \cdots \rightarrow N_{99 \in \mathbb{K}/\mathbb{K}'}}_{\textbf{Card}:~2.5e^{+3}} \xlongrightarrow[\Phi^{*}]{\text{Branching}} \underbrace{N_{100 \in \mathbb{K}'} \rightarrow \cdots \rightarrow N_{149}}_{\textbf{Card}:~5.0e^{+3}}
    \\ \xlongrightarrow[\Phi^{*}]{\text{Branching}} \underbrace{N_{150 \in \mathbb{K}'} \rightarrow \cdots \rightarrow N_{199 \in \mathbb{K}/\mathbb{K}'}}_{\textbf{Card}: 1.0e^{+4}} \xlongrightarrow[\Phi^{*}]{\text{Branching}} \underbrace{N_{200 \in \mathbb{K}'} \rightarrow \cdots \rightarrow N_{299 \in \mathbb{K}/\mathbb{K}'}}_{\textbf{Card}: 2.0e^{+4}}.
\end{multline}
Sec~\ref{sec:appendix_sampling} provides a detailed algorithmic procedure.

\newpage
\textbf{Datasets.} This paper utilizes ShapeNet, a widely recognized dataset comprising a vast collection of 3D object models across multiple categories, and MedShapeNet, a curated collection of medical shape data designed for advanced imaging analysis. 

\begin{enumerate}
    \item \textbf{ShapeNet.}~\cite{chang2015shapenet} We adhered to the standard protocol suggested by~\cite{yang2019pointflow} for preprocessing (\eg random shuffling, normalization) point-sets from 3D shapes, but adjusted the number of points to $10,000$, which is approximately five times larger than the standard setup. All categories were utilized in our experiments.
    \item \textbf{MedShapeNet.}~\cite{li2023medshapenet} This dataset contains nearly $100,000$ medical shapes, including bones, organs, vessels, muscles, etc., as well as surgical instruments. Our data preprocessing pipeline involves randomizing the arrangement of nodes and selecting a subset of $20,000$ points to form a standardized 3D point cloud. Considering the segmentation of each organ shape into smaller and incomplete parts in the dataset, we focused on utilizing only $1,000$ fully aggregated instances within the dataset. We applied uniform normalization and resized each shape to align within a predefined cubic space of $[-1, 1]^3 \subset \mathbb{R}^3$, facilitating comparative and computational analyses.

\end{enumerate}

\textbf{Neural Network Architectures.} In the experiment on a synthetic dataset, we utilized the similar architecture suggested in DPM~\cite{luo2021diffusion} for both functions $\mathrm{A}_{\theta}$ and $\mathrm{B}_{\theta}$. In modeling mean-field interaction, we incorporated a local particle association module, akin to the one used in DCGNN~\cite{dgcnn}. This module dynamically pools particles with close geometric proximity during the inference. All experiments were conducted using a setup of $4$ NVIDIA A100 GPUs.

\subsubsection{Training Mean-field Chaotic Diffusion Models}\label{sec:appendix_simulation}
This section aims to present the algorithmic implementation of mean-field score matching and training procedure with objective $\mathbf{(P3)}$. We train our score networks based on a mean-field score objective, incorporating the Sobolev norm and reducible network structures. The training procedure is comprehensively outlined in the following three steps. 

\begin{steps}
    \item \textbf{Initialization}. Consider an index set $\mathbb{K} = \{0, \cdots, K\}$ for the discrete simulation of SDEs, and its subset $\mathbb{K'}\subseteq \mathbb{K}$ for particle branching steps. This operation is selectively applied to steps $k \in \mathbb{K}'$ out of the entire sequence of diffusion steps, $\mathbb{K}$. For simplicity, let us denote $\zeta_{t} \coloneqq \mathcal{N}(\mathbf{m}_{\zeta}(t), \sigma_{\zeta}^2(t)\mathbf{I}_d)$, where Gaussian parameters are selected from Appendix C~\cite{song2021denoising}. Then, we sample $B$ i.i.d particles having a form of
    \begin{equation}
    \nu_{t_k} \sim \mathbf{Y}_{t_k}^{b} = 
    \begin{cases}
        \mathbf{Y}_{t_k}^{b, N_k} \sim \zeta_{t_k}^{\otimes N_k}, \quad \forall k \in \mathbb{K}', \\  
        \mathbf{Y}_{t_k}^{b, N_{k+1}} \sim (\mathbf{Id}^{\otimes \mathfrak{b} - 1}  \otimes \Psi^{\theta})_{\#}[\zeta_{t_k}^{\otimes N_k}], \quad \forall k \in \mathbb{K} \setminus \mathbb{K}'.
    \end{cases}
    \end{equation}
    Consequently, the cardinality of particles changes with each diffusion step $k$. Specifically, if $k$ belongs to the set for particle branching, $\mathbf{Card}(\nu_{t_k}) = N_{k+1}$; Otherwise, it remains at $\mathbf{Card}(\nu_{t_k}) = N_k$. 
    \item \textbf{Estimation of Sobolev Norm.}
    We first define the discretization of progressively measurable process $\mathcal{G}_t^{\theta}$ with respect to $\mathbf{Y}_{t_k}^{b}$ and its Jacobian as follows:
    \begin{align}
         &\mathcal{G}_{t_k}^{\theta} = \mathbf{s}_{\theta}^{\otimes \mathbf{Card}(\nu_{t_k})}(t_k, \mathbf{Y}_{t_k}^{b}, \nu_{t_k}) - \nabla \log \zeta_{T - t_k}^{\otimes \mathbf{Card}(\nu_{t_k})}(\mathbf{Y}_{t_k}^{b})
        \\& \mathcal{J}\mathcal{G}_{t_k}^{\theta} = \mathcal{J}\mathbf{s}_{\theta}^{\otimes \mathbf{Card}(\nu_{t_k})}(t_k, \mathbf{Y}_{t_k}^{b}, \nu_{t_k}) - \nabla^2 \log \zeta_{T - t_k}^{\otimes \mathbf{Card}(\nu_{t_k})}(\mathbf{Y}_{t_k}^{b}),
    \end{align}
    where each term $\mathrm{A}_{\theta}^{\otimes N_k}$ and $\mathrm{B}_{\theta}$ for score networks $\mathbf{s}_{\theta}^{N_k}$ is estimated by the Table~\ref{alg:appendix_sampling}, Step II. Note that $\mathbf{Card}(\nu_{t_k})$ denotes the cardinality of sampled particles.

    \item \textbf{Update Network Parameters.} For the calculated estimations above, we update the networks by MF-SM with respect to the subdivision of chaotic entropy, $\mathbf{(P3)}$ in Eq.~\ref{eq:P3}:
    \begin{equation}
        \theta ~\longleftarrow~ \theta - \nabla_{\theta}\frac{1}{B |\mathbb{K}|}\sum_b^B \sum_{k \in \mathbb{K}}\left[\frac{1}{\mathfrak{b}^k}\mathbb{E}\left[{\norm{\mathcal{G}_{t_k}^{\theta}}_E^2 + \norm{\mathcal{J}\mathcal{G}_{t_k}^{\theta}}_F^2}\right]\right].
    \end{equation}
\end{steps}

\newpage
\subsubsection{Sampling Scheme for Mean-field Chaos Diffusion Models}\label{sec:appendix_sampling}
To sample the denoising dynamics, this work proposes a modified Euler scheme, adapted for \textit{mean-field interacting particle systems}~\citep{bossy1997stochastic, dos2022simulation}, and approximate the stochastic differential equations in the mean-field limit. The proposed scheme involves a four-step sampling procedure. 

\begin{steps}\label{alg:appendix_sampling}
    \item \textbf{Initialization.} Consider an index set $\mathbb{K} = \{0, \cdots, K\}$ for the discrete simulation of SDEs, and its subset $\mathbb{K'}\subseteq \mathbb{K}$ for particle branching steps. In the initial step $k=0$, the probability measure $\varrho_{t_0}^{N_0}d\mathbf{x}^{N_k}$ is set to $N_0$-product of standard Gaussian density, $\ie \mathcal{N}^{\otimes N_0}(\mathbf{I}_{N_0 d})$. For the steps $k > 0$, we sample i.i.d $B$ particles from the branched probability measure obtained in the previous step: $\{\mathbf{X}_{t_k}^{b, N_k}\}_{b \leq B} \sim \varrho_{t_k}^{N_k}d\mathbf{x}^{N_k}.$
 
    \item \textbf{Estimation of Vector fields.} Given sampled $(N_k d)$-dimensional $B$ vectors in the previous step, we estimate the vector fields in this step. Recall that the vector fields are given as $\nabla V^N(t, \mathbf{x}, \nu_t^N ; \theta) \coloneqq f_t^{\otimes N}(\mathbf{x}) - \sigma_t^2\mathbf{s}_{\theta}(t, \mathbf{x}, \nu_t)$. Given the definition of MF VP-SDE where $(\beta_{\text{max}}, \beta_{\text{min}}) = (20, 0.1)$, we have
    \begin{equation}
        f_t^{\otimes N_k}(\mathbf{X}_t^{b, N_k}) = -\frac{\beta_t}{2}\mathbf{X}_t^{b, N_k}, \quad \beta_t = \beta_{\text{min}} + t(\beta_{\text{max}} - \beta_{\text{min}}), \quad b \leq B.
    \end{equation}
    To estimate $\mathrm{A}_{\theta}$, we adhere to the definition of a reducible architecture explored in Sec~\ref{sec:appendix_ECR}, namely, the concatenation of equi-weighted, identical networks.
    \begin{equation}
        \mathrm{A}_{\theta}^{\otimes N_k}(t_k, \mathbf{X}_{t_k}^{b, N_k}) = \frac{1}{N_k}[\mathrm{A}_{\theta}(t_k, \mathbf{X}_{t_k}^{b, 1, N_k}), \cdots, \mathrm{A}_{\theta}(t_k, \mathbf{X}_{t_k}^{b, N_k, N_k})]^T \in \mathcal{X}^{N_k}.
    \end{equation}
    The mean-field interaction is formally redefined in the following manner: it involves the projection of the probability measure as $\pi_{\#}^{i}\nu_t^{N_k} = \nu_t^{i, N_k} \sim \{\mathbf{X}_{t_{k}}^{b, i, N_k}\}_{b \leq B}$: 
    \begin{equation}\label{eq:appendix_mean_field_desc}
        \quad \left( [\mathrm{B}_{\theta} * \nu_{\mathbb{B}_R}^i](\mathbf{X}_{t_{k}}^{b, N_k})\right)^{\otimes N_k}  = \frac{1}{N_k}\left[ [\mathrm{B} * \pi^1_{\#}\nu_{t_{k}}^{N_k}], \cdots, [\mathrm{B} * {\pi_{\#}}_{t_k}^{N_k}]\right]^T(\mathbf{X}_{t_{k}}^{b, N_k}).
    \end{equation}
    With the finite cut-off radius $R$, we consider Euclidean balls to define truncated convolution:
    \begin{equation}
        \mathbb{B}_R \coloneqq \mathbb{B}_R^{\mathbf{x} = \mathbf{X}_{t_k}^{b, i, N_k}} = \left\{\mathbf{y} ; d_E^{2}(\mathbf{y}, \mathbf{X}_{t_k}^{b, i, N_k}) \leq R \right\}. 
    \end{equation}
    Given definition above, each component in Eq.~\ref{eq:appendix_mean_field_desc} is given by 
    \begin{equation}
    \begin{split}
         [\mathrm{B}_{\theta} * \nu_{\mathbb{B}_R}^i](\mathbf{X}_{t_k}^{b, i, N_k}) & \propto \frac{1}{N_k - 1}\sum_{i \neq j}^{N_k}\int_{\mathbb{B}_R} \mathrm{B}_{\theta}(\mathbf{X}_{t_k}^{b, i, N_k} - \mathbf{X}_{t_k}^{b, j, N_k})\nu_{t_k}^{j, N_k}(d\mathbf{X}_{t_k}^{b, i, N_k}).
    \end{split}
    \end{equation}
    
    \item \textbf{Applying Euler Schemes.} Having collected estimated terms from the previous step, we apply the Euler scheme to have particle simulation of dWGFs accordingly.
    \begin{equation}
        \mathbf{X}_{t_{k+1}}^{b, N_k} = \mathbf{X}_{t_k}^{b, N_k} + \nabla V^N(t, \mathbf{X}_{t_k}^{b, N_k}, \nu_{t_k}^{N_k} ; \theta)\Delta_t + \sqrt{\beta_t} \Delta_t B_{t_k}^{N_k}, \quad b \leq B,
    \end{equation}
    where $\Delta_t B_{t_k}^{N_k} \coloneqq B_{t_k}^{N_k} - B_{t_{k-1}}^{N_k} \sim \mathcal{N}[\Delta_t I_{d {N_k}}]$.

    \item \textbf{Particle Branching.} In the final step, we apply the particle branching operation to enhance the cardinality. This operation is selectively applied to steps $k \in \mathbb{K}'$ out of the entire sequence of diffusion steps, $\mathbb{K}$.
    \begin{equation}
         [\mathbf{X}_{t_{k}}^{ B, \otimes (\mathfrak{b}-1)N_{k}}, ~ \Psi_{\theta}(\mathbf{X}_{t_{k}}^{B, N_{k}})] ~\rightharpoonup~ \mathbf{X}_{t_{k}}^{B, N_{k + 1}}, \quad (\mathbf{Id}^{\otimes \mathfrak{b} - 1}  \otimes \Psi^{\theta}_{N_{k+1}})_{\#}[\varrho_{t_k}^{N_k}]  ~\rightharpoonup~ \varrho_{t_k}^{N_{k+1}} d\mathbf{x}^{N_{k+1}}.
    \end{equation}
    When branching particles, the cardinality grows as $N_{k+1} = \mathfrak{b}N_k$, and the entire sampling scheme is repeated until reaching the final step $k \rightarrow K$.
\end{steps}
\newpage

\end{document}